\documentclass[11pt, a4paper, logo, copyright]{pluralisresearchiclr}
\makeatletter
\renewcommand{\abscontent}{
	\noindent
    \tcolorbox[on line, box align=base, enhanced, colback=prbg, colframe=prframe, arc=4pt, boxrule=0.4pt, left=6pt, right=6pt, top=4pt, bottom=4pt, width=\linewidth, parbox=false]
    {\absfont \theabstract}
	\@ifundefined{@keywords}{}{
		\vskip1em \noindent \keywordsfont \@keywords}
    \endtcolorbox
}
\makeatother
\usepackage{amsmath,amsfonts,bm}

\def\eqref#1{equation~\ref{#1}}
\def\1{\bm{1}}

\def\vtheta{{\bm{\theta}}}

\def\mE{{\bm{E}}}

\def\mG{{\bm{G}}}

\def\mP{{\bm{P}}}
\def\mQ{{\bm{Q}}}

\def\mT{{\bm{T}}}
\def\mU{{\bm{U}}}

\def\mW{{\bm{W}}}
\def\mX{{\bm{X}}}

\def\mZ{{\bm{Z}}}

\DeclareMathAlphabet{\mathsfit}{\encodingdefault}{\sfdefault}{m}{sl}
\SetMathAlphabet{\mathsfit}{bold}{\encodingdefault}{\sfdefault}{bx}{n}

\def\gI{{\mathcal{I}}}

\newcommand{\Ls}{\mathcal{L}}
\newcommand{\R}{\mathbb{R}}

\pdftrailerid{redacted}

\makeatletter
\renewcommand\bibentry[1]{\nocite{#1}{\frenchspacing\@nameuse{BR@r@#1\@extra@b@citeb}}}
\makeatother

\usepackage{kantlipsum, lipsum}
\usepackage{dsfont}
\usepackage{algorithm}
\usepackage{algpseudocode}
\usepackage{adjustbox}
\usepackage{multirow}
\usepackage{nicefrac}
\usepackage{float}


\usepackage{tikz}
\usetikzlibrary{arrows.meta, positioning, shapes.geometric, fit}
\usepackage{graphicx}

\newcolumntype{R}[2]{%
    >{\adjustbox{angle=#1,lap=\width-(#2)}\bgroup}%
    l%
    <{\egroup}%
}

\usepackage[capitalize,noabbrev]{cleveref}
\Crefname{section}{Section}{Sections}
\Crefname{table}{Table}{Tables}
\Crefname{equation}{Equation}{Equations}
\Crefname{figure}{Figure}{Figures}
\Crefname{algorithm}{Algorithm}{Algorithms}
\Crefname{appendix}{Appendix}{Appendices}

\newcommand{\swarm}{SWARM}

\usepackage{pifont}

\usepackage{subcaption}
\usepackage{url}

\graphicspath{{figures/}}

\title{Agora: Collective and Permissionless Internet-Scale Pretraining of Large Language Models}

\keywords{Dashboard: \href{https://agora.pluralis.ai}{agora.pluralis.ai}
\\ Code: \href{https://github.com/PluralisResearch/agora}{github.com/PluralisResearch/agora}
\\ Keywords: Protocol Learning, Distributed Pretraining, Communication-Efficient Distributed Systems}

\reportnumber{}

\renewcommand{\today}{} 

\author{Gil Avraham\textsuperscript{*}, Violetta Shevchenko\textsuperscript{*}, Hadi Mohaghegh Dolatabadi\textsuperscript{*}, Karol Pajak\textsuperscript{*}, James Snewin, Harry Xi, Rodney O'Donnell, Thalaiyasingam Ajanthan, Sameera Ramasinghe, Chamin Hewa Koneputugodage, Shamane Siriwardhana and Alexander Long}
\affil{Pluralis Research \quad \textsuperscript{*}Equal contribution}

\begin{abstract}
Training large language models at the multi-billion to trillion parameter
scale is confined to datacenters, where data-parallel (DP) and model-parallel (MP) techniques presume homogeneous accelerators, high-speed interconnects, and a single orchestrating entity. Frontier model development is thereby concentrated among the few groups able to assemble such clusters. Meanwhile, an enormous pool of compute remains unusable for training: consumer and professional GPUs that are heterogeneous, preemptible, individually owned, and connected only by the internet. We present Agora, a system that makes efficient use of this compute. Agora combines bandwidth-efficient pipeline-parallel model sharding over internet-grade links with multi-party, fault-tolerant collective operations. Each participant holds only one stage of the model, and no single party ever possesses the full weights. We term this setup \textit{Protocol Learning}: it enables collectively trained, collectively owned models, opening a path to open-source frontier training with economic sustainability. This report presents the outcome of a research effort spanning communication-efficient parallelism, asynchronous optimization, and fault-tolerant systems design. It culminates in the first demonstration of its kind: Pluralis-8B, an open, permissionless pretraining run of an 8.6B-parameter model on 500B tokens of FineWeb-Edu. The model was trained over 40 days by 330 contributor nodes, predominantly consumer GPUs on internet connections, joining and leaving throughout. The run sustained $\sim$170k tokens/s and 4.2 tokens per TFLOP of pooled compute, 63\% of the efficiency of a centralized H100 baseline, and converged to within a small margin of a centralized reference run.
\end{abstract}

\begin{document}
\maketitle

\section{Introduction}\label{sec:intro}
Frontier-scale model training is concentrated among a handful of datacenter operators with privileged access to islands of tightly coupled accelerators and dedicated high-performance network fabrics~\cite{grattafiori2024llama, liu2024deepseek}. The reason is technical: the communication primitives that underpin large-scale training presume tightly coupled hardware, with single-digit-microsecond latencies and hundreds of gigabits per second of bisection bandwidth~\cite{shoeybi2019megatron, hu2025demystifying}. A cluster satisfying these assumptions cannot be assembled ad hoc; it must be rented from an operator that has purpose-built one, or constructed outright. What if a frontier language model could be trained not in a datacenter, but by thousands of consumer devices cooperating over the internet, with anyone free to contribute compute and share the result? The pool of such hardware is enormous, and it is precisely the compute the current paradigm cannot touch.

Realizing this alternative requires solving three problems that centralized training never encounters. First, the model no longer fits any single participant, so it must be sharded across devices whose links are orders of magnitude slower than a datacenter fabric, and the activations and gradients crossing shard boundaries on every microbatch would saturate such links; subspace networks~\citep{ramasinghe2025ssn} confine this inter-stage signal, cutting the communicated volume by two
orders of magnitude, making pipeline parallelism a viable option at internet bandwidth. Second, data-parallel replicas must be kept in consensus, but the standard per-step synchronous gradient exchange places a blocking barrier on links with round-trip latencies of tens to hundreds of milliseconds, stalling every participant; sparse, asynchronous parameter averaging~\citep{beton2025sparta, ajanthan2026asyncmesh} exchanges only a small fraction of the slow-moving weights, infrequently and in the background, so weight synchronization overlaps with computation and never blocks the optimizer step. Third, the participants themselves are unreliable: heterogeneous consumer GPUs, individually owned, joining and leaving without warning, so fault tolerance must be a property of the training step itself rather than a recovery mechanism around it~\citep{avraham2025node0, ryabinin2023swarm}. Solving these problems yields more than a training system: because the model is trained model-parallel, no single party ever possesses the complete weights, and Unextractable Protocol Models (UPMs)~\citep{long2025upm} extend this so the weights can never be extracted from a node at all. A trained model's value is thereby retained within the network of contributors that created it, offering a path to economically sustainable open participation. We call this paradigm \textit{Protocol Learning}.

In this report we present Agora, the system that realizes it. A purpose-built system for multi-party, permissionless training of LLMs over geo-distributed, heterogeneous GPUs connected via the internet. The system is elastic: token throughput scales as compute is added, and nodes failing, joining, or leaving are handled gracefully. No single entity owns or orchestrates the cluster; individual accelerators interact with the run only through a defined protocol layer that provides pipeline- and data-parallel training with fault-tolerant collective operations.

Many key changes that depart from traditional centralized LLM training have been made to support training under such conditions. We analyze the resulting system from three angles:

\begin{itemize}
    \item \textbf{Convergence.} How is convergence affected by compute heterogeneity, partial all-reduce participation, and node elasticity?

    \item \textbf{Throughput.} How do node failures, continuous churn, synchronization overhead, and faulty nodes affect overall token
    throughput?

    \item \textbf{Permissionless participation.} How do joining nodes catch up from stale weights without disrupting the run? How is incoming compute placed across pipeline stages for efficient use?
\end{itemize}

We validate Agora and answer these questions empirically, with Pluralis-8B: a publicly open, permissionless pretraining run of an 8.6B-parameter model on 500B tokens of FineWeb-Edu, conducted for 40 days by 330 contributor nodes, predominantly consumer GPUs, alongside a small set of operator nodes. The training loss closely tracks a centralized reference throughout the run despite node failures. Furthermore, ablations at the 1B scale show that training tolerates heterogeneity of up to 10:1 and degrades only when all-reduce participation falls below $\sim$85\%, and fully recovers once failing nodes are pruned. On throughput, the run sustains $\sim$170k tokens/s, 4.2 tokens per TFLOP, 63\% of the efficiency of a centralized H100 baseline, and holds this rate through continuous node churn (669 joins and 607 departures).

The remainder of this report is organized as follows. \Cref{sec:decentralizing} reviews the assumptions of centralized distributed training and states the problem setting. \Cref{sec:agora} describes the Agora system, its training roles, throughput optimization, fault-tolerance mechanisms, and supporting services. \Cref{sec:pluralis-8b} presents the machine-learning changes specific to Pluralis-8B. \Cref{sec:results} reports the public run, contributor dynamics, throughput and efficiency, and convergence. \Cref{sec:ablations} presents ablation studies at 1B scale that probe the system's convergence robustness.

\section{Decentralizing LLM Pre-training}
\label{sec:decentralizing}

\subsection{Distributed Training Today}
\label{sec:distributed-today}

A language model whose parameters, optimizer state, and activations exceed the
memory of a single accelerator must be partitioned across many accelerators that
advance in lockstep under synchronous optimization. The prevailing approach assumes
a single, homogeneous, co-located cluster: a fixed pool of identical accelerators,
provisioned together and connected by a purpose-built high-speed fabric, across which
a global batch is processed and one optimizer step applied at each iteration.

Partitioning is expressed along several largely orthogonal axes. Data parallelism (DP) replicates the model
and splits the batch across replicas, synchronizing weight gradients once per step;
its memory-efficient variants, ZeRO and FSDP~\citep{rajbhandari2020zero, zhao2023pytorch}, shard the
optimizer state, gradients, and parameters across the replica group to remove
redundancy. Tensor parallelism (TP)~\citep{shoeybi2019megatron} partitions the matrix
multiplications within each layer across devices, while pipeline parallelism (PP)~\citep{huang2019gpipe, narayanan2019pipedream} divides the model's layers into
sequential stages, each placed on a different device and fed a stream of microbatches.
Two further axes are added as the workload requires: context parallelism (CP), which
shards the sequence dimension for long-context training, and expert parallelism (EP),
which distributes the experts of a mixture-of-experts model. Production runs compose
these axes to fit a model into aggregate memory while keeping every device occupied.
Llama~3~405B, for example, is trained on 16,384 H100 GPUs under a four-dimensional
scheme combining tensor, pipeline, context, and data parallelism~\citep{grattafiori2024llama},
whereas DeepSeek-V3~\citep{liu2024deepseek} omits tensor parallelism and relies instead on pipeline
parallelism and 64-way expert parallelism. The decomposition
differs across systems, but the principle is the same: the model training is partitioned along
whichever combination of axes balances per-device memory against communication cost
for the available hardware.

That choice is governed less by raw compute than by a steep hierarchy in communication
bandwidth, since each axis carries a substantial collective communication operation: a gradient all-reduce for
DP, all-gather and reduce-scatter for TP, point-to-point activation and gradient
transfers across stage boundaries for PP, and all-to-all exchanges for EP. Within a
node, accelerators are joined by NVLink and NVSwitch on the order of hundreds of
gigabytes per second; across nodes, the fabric falls by close to an order of
magnitude, to the tens of gigabytes per second of InfiniBand or RoCE. The exchanged
volumes are large enough that, without careful design, training becomes bound by communication
rather than computation. The system is therefore co-designed with the network: the
communication-heaviest axis is mapped onto the fastest link (TP confined within a
node, PP spanning the slower inter-node fabric, DP occupying the remainder) and
substantial effort is spent overlapping communication with computation so that the
interconnect does not lie on the critical path. The sensitivity of this mapping to the
hardware is illustrated by DeepSeek-V3, whose reduced intra-node NVLink bandwidth
motivated both the omission of tensor parallelism and a constraint on expert routing
that bounds the number of nodes each token traverses~\citep{liu2024deepseek}. The premise
throughout is the one stated in \Cref{sec:intro}: single-digit-microsecond
latencies and hundreds of gigabits per second of bisection bandwidth between
participating devices.

A second premise is synchrony. Training proceeds as a sequence of global barriers: at
each step every model replica holds identical weights, exchanges gradients, and applies the
same update, so that replicas do not drift apart. This coupling makes the
communication patterns above well-defined, and also constrains fault tolerance, since
a step cannot complete until every participant has, and a single slow or failed device
stalls the run. At the scale of tens of thousands of accelerators such failures are
common rather than exceptional: the Llama~3~405B run recorded 419 unexpected
interruptions over 54 days, roughly one every three hours, the majority attributable
to GPU or memory faults~\citep{grattafiori2024llama}. Centralized systems tolerate this through
mechanisms layered around the synchronous core rather than built into it: frequent
checkpointing, provisioned spare nodes, and automated detection-and-restart machinery
that recover the run after each interruption. These are effective because the cluster
is owned and stable: a faulty accelerator is interchangeable with its replacement, the
topology is fixed and known, and a job can be resumed from its last checkpoint without
any participant departing permanently or without warning.

Centralized large-scale training thus rests on three mutually reinforcing assumptions:
a high-bandwidth, low-latency interconnect of fixed and known topology; a fleet of
homogeneous accelerators; and a stable, operator-owned pool of hardware that fails only
occasionally and recoverably. The barrier to entry described in \Cref{sec:intro}
follows from them, since a cluster satisfying all three must either be rented from an
operator that has built one or constructed outright. The remainder of this section
makes precise what changes when each assumption is removed (internet-grade bandwidth
and latency in place of a datacenter fabric, heterogeneous and unequally connected
accelerators in place of a uniform fleet, and preemptible, unowned nodes that may join
or leave at any time) and why the resulting regime cannot be served by the centralized
stack.

\subsection{Problem Statement and Solution}
\label{sec:problem-solution}

The compute this work aims to use is the compute that the centralized paradigm of
\Cref{sec:distributed-today} excludes: the aggregate of accelerators held by
universities, independent researchers, and individual contributors, substantial in
total but distributed across the globe, owned by no single party, and connected only by
the public internet. Training over such a pool requires relaxing all three assumptions
on which the centralized stack rests. A system targeting this setting must operate under
the following conditions:

\begin{itemize}
  \item \textbf{Internet-grade communication.} Participants are not co-located and do
  not share a datacenter fabric. Inter-node links offer hundreds of megabits per second
  rather than hundreds of gigabits, with round-trip latencies of tens to hundreds of
  milliseconds and no guarantee of stability over time.

  \item \textbf{Preemptibility and churn.} Nodes are not owned by the operator and may
  join, leave, or fail at any time, without warning or coordination. Departure is common
  rather than exceptional, and the system cannot assume a fixed set of participants for
  the duration of a single step.

  \item \textbf{Heterogeneity.} Participating accelerators differ in compute throughput,
  memory capacity, and network quality. No uniform-fleet assumption is available: a
  contributor may hold a consumer GPU on a slow residential link, and the system must
  accommodate it alongside more capable hardware.

  \item \textbf{No central orchestrator on the training path.} With no owned, supervised
  cluster, there is no privileged process holding the full model or driving the run;
  coordination must emerge from the participants rather than from a single controlling
  node.
\end{itemize}

The objective is to train a large language model under these conditions while keeping
aggregate token throughput competitive: to make the contributed compute behave, from
the perspective of the training process, like a single coherent cluster, with the
effects of low bandwidth, variable latency, heterogeneity, and node churn absorbed by
the system rather than passed through to convergence or throughput.

Deploying an existing training framework onto such a pool illustrates why these
requirements call for a different design rather than a reconfiguration of the existing
one. Frameworks such as TorchTitan~\citep{liang2025torchtitan} and Megatron-LM~\citep{shoeybi2019megatron} are built on the
single-program-multiple-data model, executing synchronous collectives (gradient
all-reduce for data parallelism, all-gather and reduce-scatter for sharded variants, and
point-to-point activation transfers for pipeline parallelism) over communicator
libraries such as NCCL that assume a high-bandwidth fabric and a fixed set of ranks.

Consider first the idealized case in which no node fails, so that only the bandwidth
assumption is violated. Data-parallel training synchronizes gradients once per optimizer
step, and that exchange must complete before the step can proceed. Typically, the gradients are of the same order as the parameters themselves (tens of
gigabytes for a billion-parameter-scale model), and a ring all-reduce moves roughly twice that volume across each
participant's link per step. Over a residential uplink of a few hundred megabits per
second, a single exchange takes on the order of tens of minutes, against the sub-second
step times of a datacenter run; the corresponding per-block and per-microbatch figures
are quantified in \Cref{ssec:results_collective_comms}. Pipeline parallelism is similarly
affected, as the activations crossing each stage boundary amount to hundreds of megabits
per microbatch (\Cref{ssec:results_throughput}), so the pipeline becomes communication-bound well
before it is compute-bound. The overlap of communication with computation that hides
these collectives in a datacenter is no longer possible when the link is several orders
of magnitude slower; accelerators idle waiting for data, and utilization drops sharply.
Training does not fail here, but throughput falls to a level that makes the run
impractical.

Introducing node failures makes the run unviable altogether. A static communicator fixes its set of
ranks at initialization and offers no mechanism for a participant to leave or a new one
to join mid-run; when a rank disappears during a collective, the operation stalls until
it times out and then aborts the process group, terminating the job. This is a direct
consequence of the synchronous coupling of \Cref{sec:distributed-today}: a single
departure halts every participant. The only recovery is to restart from the most recent
checkpoint, a cadence mismatched to a setting in which nodes leave every few minutes, so
the run makes little net progress between restarts. Heterogeneity compounds the problem,
since a synchronous, evenly sharded program advances at the pace of its slowest rank, so
a single weak contributor limits the whole group, while a node whose memory cannot hold
its assigned shard cannot participate at all. These are not implementation defects but
consequences of design choices (global synchronization barriers, static process
membership, uniform sharding, and recovery by restart) that are appropriate in the
datacenter and unsuitable for the target setting.

Meeting these requirements calls for a system in which fault tolerance, elasticity, and
communication efficiency are properties of the training step itself rather than
mechanisms layered around a synchronous core. Our solution, detailed in
\Cref{sec:agora}, restricts itself to the two parallelism axes that can be made
viable over the internet (pipeline and data parallelism) and modifies the
communication on each so that it does not block computation. Along the pipeline axis, the
inter-stage signal is compressed by roughly two orders of magnitude, so the activations
and gradients crossing stage boundaries fit within the capacity of a low-bandwidth link
(\Cref{ssec:reparam_ssn}). Along the data-parallel axis, the per-step synchronous gradient
all-reduce is replaced: each replica takes local optimizer steps on its own gradients,
and replicas are kept in consensus by averaging only a small, fixed fraction of their
parameters on an infrequent, asynchronous cadence that overlaps with computation and does
not stall the step (\Cref{ssec:async_sparta}). Orchestration is separated from
computation and placed in lightweight trainer processes that route work through the
available accelerators along the fastest paths, weighting faster nodes more heavily and
skipping failed ones without interrupting the pipeline, so that heterogeneity and churn
are absorbed into the routing rather than allowed to degrade throughput. The system is
consequently structured as a protocol rather than a tightly coupled program, in which
individual accelerators interact with the run only through a defined interface, and the
underlying liquid compute is abstracted away from the training process it sustains. 

\section{Agora}\label{sec:agora}
Producing a system that addresses the problems outlined in \Cref{sec:problem-solution} requires re-framing the entire approach. Agora represents this shift, and is best thought of as a protocol rather than a system. Agora defines multiple node roles, each running a process that exposes a set of interfaces, and a pre-defined protocol prescribing how calls to these interfaces are sequenced to train an LLM. We coin this approach \textit{Protocol Learning}; it adheres to the requirements of the problem statement. Given a set of active nodes, the protocol defines the sequence of interface calls by which a new node joins without disrupting training, and likewise handles a node's departure or failure gracefully. In the following sections, we describe how a large language model can be trained over such a protocol.

\subsection{System Overview}

\begin{figure}[t]
    \centering
    \includegraphics[width=\linewidth]{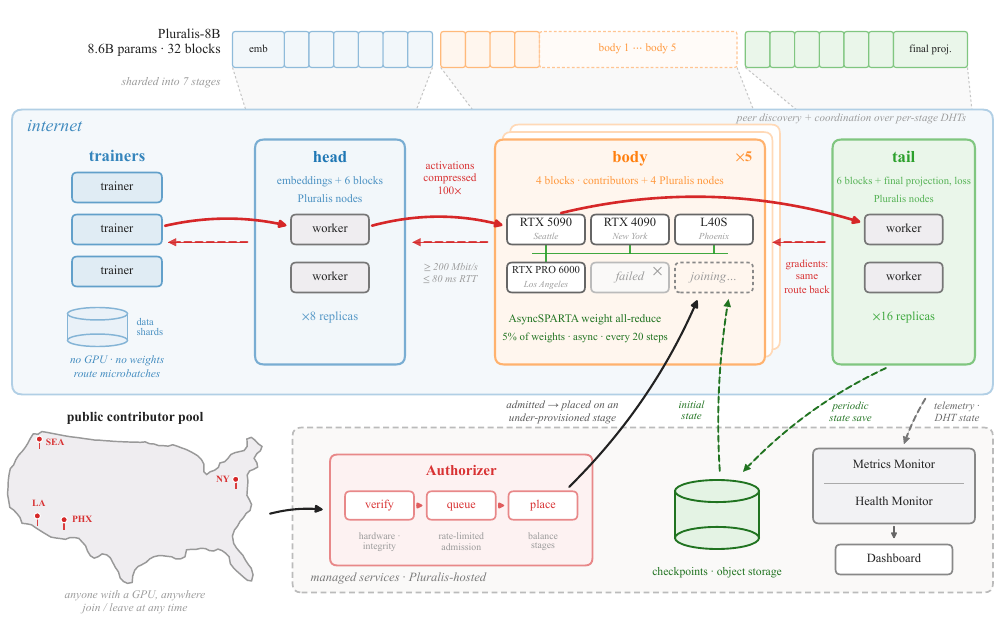}
    \caption{\textbf{Overview of the Agora system, instantiated for the Pluralis-8B run.}
    The model is sharded into seven pipeline stages, each replicated across workers that hold only that stage; trainers route each microbatch through one worker per stage over internet links, and contributors join permissionlessly through the Authorizer.
    Same-stage replicas are kept in consensus by asynchronous sparse weight averaging, while Pluralis-hosted services checkpoint and monitor the run.
    }
    \label{fig:system_overview}
\end{figure}

The complete system diagram of Agora can be seen in \Cref{fig:system_overview}. 

The Agora system can be conceptualized as two parts:
\begin{itemize}
    \item \textbf{Training System.} Provides all the functionality that allows training an LLM in a distributed manner over the internet. The two roles that enable this are coined \textit{worker} and \textit{trainer}.
    \begin{itemize}
        \item A \textit{worker} (\Cref{worker_section}) typically has a GPU card and hosts a shard of weights in a pipeline stage, which is a single or multiple Transformer blocks. A worker is a stateless service and provides three functionalities: it can compute a forward and a backward pass on the layers it hosts when invoked with a batch, and it can coordinate to perform an all-reduce operation with other workers in its stage.
        \item A \textit{trainer} (\Cref{trainer_section}) samples batches from a dataset shard and routes the batches between the stages, invoking a different worker at every stage with a forward pass followed by a backward pass. A trainer has light system requirements and does not require a GPU.
    \end{itemize}

    \item \textbf{Managed Services.} If an individual wishes to train an LLM, the above-mentioned roles are sufficient and provide all the needed functionality. In addition, providing the package containing the trainer and worker roles allows pooling compute from external participants. The need for additional services arises to ease coordination, enable trustless participation, and enhance system visibility.
    \begin{itemize}
        \item The \textit{Authorizer} (\Cref{sssec:authorizer}) is a singleton component and the gatekeeper that admits workers and trainers into an ongoing run. To enable trustless participation, it performs package verification to prevent malicious behavior. To ease operational coordination, it auto-assigns incoming workers to under-resourced pipeline stages. Finally, it manages the \textit{rate} at which incoming workers are admitted into a run, which helps mitigate the stale-weight effect and improves the all-reduce success rate (\Cref{ssec:results_sync_phases}).
        \item \textit{Monitoring} happens on two distinct levels in the Agora system. At the level of individual nodes, system and training dynamics are tracked using the \textit{Metrics Monitor} (\Cref{sssec:metrics}), which provides visibility into each node running a worker or trainer and exposes resource metrics such as GPU utilization and bandwidth usage, as well as pre-clipped gradient values and the all-reduce success rate. At the level of the run as a whole, the \textit{Health Monitor} (\Cref{sssec:health-monitor}) aggregates cross-node activity, such as loss information from trainer nodes, worker node activity (e.g., sync phase, \Cref{ssec:results_sync_phases}), and contributed FLOPs.
        \item The \textit{Dashboard} (\Cref{sssec:dashboard}) provides an external-facing user interface that allows the run operator and external node runners to view training progress.\footnote{A live dashboard is available at \url{https://agora.pluralis.ai}.} It aggregates information from both the Metrics Monitor and the Health Monitor.
    \end{itemize}
\end{itemize}

As shown in \Cref{fig:system_overview}, a rule of thumb that divides the Training System from the Managed Services is that in the former, nodes logically live on the internet, have public IPs, and need to be accessible by other nodes and the Managed Services. The Managed Services can be hosted anywhere (e.g., on a private cloud provider or on the public internet) and are managed by the run operator.

\subsection{Training with Agora}
\label{ssec:training-with-agora}

Agora's training abstraction replaces the tightly synchronized view of distributed training with a protocol in which the model is partitioned into ordered pipeline stages, and each stage is replicated across a changing set of workers. A training step is therefore not executed by one process holding the full model, but by trainers that route microbatches through one worker per stage, in both the forward and backward directions. This scheme is closest to \swarm{} parallelism~\citep{ryabinin2023swarm}, though Agora departs from it substantially on both parallelism axes (\Cref{sec:pluralis-8b}).

Each stage contains one or more consecutive Transformer blocks. The first stage additionally owns the token embeddings, and the final stage owns the language-model head and the loss computation. We refer to the first stage as the \emph{head}, the last stage as the \emph{tail}, and every stage in between as a \emph{body} stage. A worker is a single GPU node holding one replica of exactly one stage.

During the forward pass, a trainer sends a microbatch to an available worker in the head stage and receives the resulting activations, which it forwards to a selected worker in the next stage; this repeats (head → body → tail) until the tail stage computes the loss. The backward pass then unwinds the microbatch along the same route in reverse (tail → body → head): activation gradients are passed back from each stage to the one before it, while the parameter gradients for the blocks of a stage are computed and retained on the worker that owns them.

Workers are stateless. A worker reserves GPU memory only while it is processing a request, and once a microbatch's activations have left the worker, it retains nothing further for that microbatch. Instead, the trainer holds the per-stage activations produced during the forward pass. To run a stage's backward, the trainer replays that stage's input activation together with the incoming activation gradient, and the receiving worker recomputes the stage's forward on the fly to rebuild the local autograd graph before computing the backward. This is the stage-level analog of activation checkpointing: rather than storing intermediate activations for a later backward pass, a worker recomputes them when the backward arrives, here at the granularity of an entire stage.

Agora therefore exposes two axes of parallelism. Across stages, it uses pipeline parallelism whereby different microbatches occupy different stages at the same time. Within a stage, it uses replicated data parallelism such that multiple workers process different microbatches concurrently, so a stage's throughput grows as more compute joins. Unlike conventional data parallelism, workers do not synchronize parameter gradients on every step; instead, the replicas of a stage accumulate gradients locally and, once the stage has collectively seen the target batch size in samples, each worker applies an optimizer step to its own accumulated gradients.\footnote{Depending on the data-parallel method, a synchronization of gradients or weights between same-stage replicas may occur before or after this optimizer step.} Replicas of a stage are kept from drifting apart by periodically averaging their weights, which avoids the per-step gradient synchronization that internet-grade links cannot sustain.

Routing microbatches in Agora is stochastic and load-aware rather than fixed. For each stage, a trainer chooses among the workers currently advertising availability, weighting the choice by observed throughput. Faster workers receive more microbatches, slower ones receive fewer, and failed or unresponsive workers are skipped without stalling the pipeline. Although these mechanics allow the system to tolerate faulty nodes and heterogeneous network conditions, they do not by themselves preserve convergence; the modifications this required, on both the pipeline- and data-parallel axes, are discussed thoroughly in \Cref{sec:pluralis-8b}. 

The resulting training loop preserves the structure of ordinary LLM pre-training (e.g., forward pass, loss computation, backward pass, and optimizer step) while changing where the computation happens and when communication is required. Activations and activation gradients cross stage boundaries; model parameters, optimizer state, and parameter gradients stay inside the worker that owns a stage replica. Periodic same-stage averaging keeps replicas close enough for convergence, while stochastic routing and elastic worker discovery let training continue under heterogeneous bandwidth, variable latency, and node churn.

\subsection{Training Roles}
\label{training_roles}

Training in Agora is carried out by two roles: the worker, which holds model parameters and performs computation, and the trainer, which holds no parameters and orchestrates the pipeline. A worker owns a single pipeline stage and is aware only of that stage, whereas a trainer holds the full pipeline topology but none of the weights. As a result, no participant ever holds the complete model: the parameters are partitioned across workers by stage, while the only nodes with a global view carry no parameters at all. There are two main data-exchange channels, both implemented as RPC over peer-to-peer (P2P) connections: a trainer--worker channel that carries activations and gradients for the forward and backward passes, and a worker--worker channel that performs all-reduce averaging of weights within each stage.

In addition, all of the run's metadata, represented as a key/value dictionary, is stored in a distributed hash table (DHT), with one instance per stage, plus one shared by the trainers. While this run was not decentralized and required specific Pluralis nodes, our general goal is to create a protocol that can function with no central orchestration. Hence, we manage and track the topology, state, and progress via a distributed data store over the same nodes used to perform the computation.

It is worth emphasizing that both worker and trainer roles are designed so a node running either role can be added and removed from a run at any time without affecting ML convergence or the logical operation of the system (i.e., elasticity). One useful way to think about these roles is as hardware resources (CPUs, RAM, GPUs) with different functions in mind that utilize these resources. From this aspect, hardware resources can be dynamically added and removed from a training run and throughput performance behaves accordingly.

\subsubsection{Worker}
\label{worker_section}

\begin{figure}[t]
    \centering
    \includegraphics[width=0.95\linewidth]{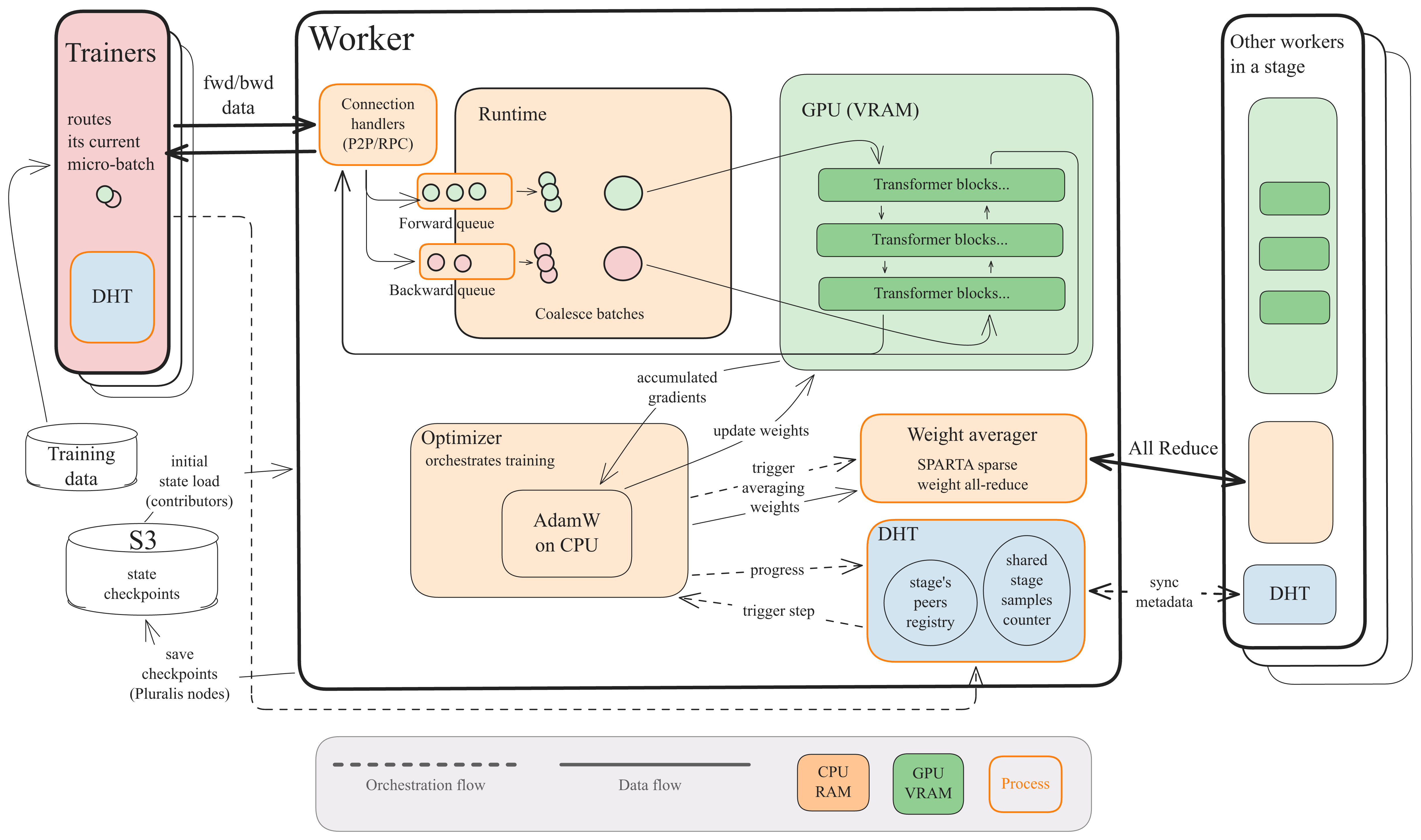}
    \caption{\textbf{Diagram of the worker architecture.} A worker hosts one replica of a single
    pipeline stage. Connection handlers receive forward/backward microbatches from the trainers
    over P2P RPC and enqueue them into forward and backward task queues; the Runtime coalesces
    queued microbatches and runs them through the stage's Transformer blocks on the GPU. The
    optimizer orchestrates training: gradients accumulated on the GPU are moved to host RAM, the
    offloaded AdamW steps on CPU, and the updated weights are copied back to VRAM. On the
    optimizer's trigger, synchronized via DHT, the weight averager all-reduces the stage's weights with the other
    workers in the stage. Contributors load their initial state from
    checkpoints in cloud object storage, saved periodically by Pluralis nodes.}
    \label{fig:worker_diagram}
\end{figure}

A worker holds one replica of a single stage's weights: one or more consecutive Transformer blocks, together with the embedding layer at the head stage or the language-model head at the tail stage. It runs the forward and backward passes for the batches a trainer routes to it. Once the stage's workers have collectively accumulated the target batch size in samples, they each take an optimizer step on their local optimizers with the gradients they accumulated. Because it is aware only of its own stage, a worker operates independently of the rest of the pipeline. Workers periodically average their weights within the stage to prevent drift. The worker architecture is illustrated in \Cref{fig:worker_diagram}. The optimizer is offloaded to the CPU: keeping the master weights and AdamW moments in CPU RAM frees GPU memory for activations and batch coalescing, and places the weights where the background averager can update them without interrupting GPU compute.

\paragraph{State loading and sync phase.}\label{sec:worker_state_load_sync_phase} A worker carries no run-wide state beyond its stage's parameters and optimizer state, which makes it freely replaceable: a failed worker can be substituted by a fresh one that loads the current stage state and joins the stage. Pluralis-owned workers download their state from others directly, whereas contributor nodes get the state from periodically refreshed checkpoints stored in cloud object storage, to reduce the load on Pluralis workers. If the state it downloaded is more than a configured number of steps stale, which is usually the case due to its low transfer speed, the contributor would drag the stage's average backward if it started contributing immediately, so it instead enters a two-phase catch-up mode: \emph{weight sync phase} and \emph{optimizer sync phase}. In the former, it only receives updates to its weights: it is invisible to trainers (no batches are routed to it) while it takes part in state averaging with weight zero, absorbing the averaged weights without mixing in its own stale ones. In the latter phase, it becomes visible to trainers and runs real forward and backward passes to warm its optimizer's momentum and variance on live gradients, but still contributes nothing to the weight averaging and reports no samples toward the step's batch-size target. Once both phases elapse, it resumes full participation. The state-loading and sync-phase flow is depicted in \Cref{alg:join}.

\begin{algorithm}[t]
\caption{\textbf{Worker joining flow.} A joining worker acquires its stage's state, waits for a
window clear of averaging rounds, and starts serving (\Cref{alg:training}); stale state
first passes through the two sync phases (\Cref{sec:worker_state_load_sync_phase}).
Pluralis-8B values: averaging interval $A{=}20$ steps, block window $W{=}2$ steps,
staleness threshold $S{=}3$ steps, $N_1{=}400$, $N_2{=}100$ steps.}
\label{alg:join}
\small
\begin{algorithmic}[1]
\Statex \textbf{modes:} $\textsc{WeightSync} \to \textsc{OptimizerSync} \to \textsc{Normal}$
        \Comment{sync phases, then full participant}
\Statex
\Procedure{JoinRun}{}
    \State connect to the stage's DHT; begin tracking $t_{\mathrm{glob}}$
    \If{first worker of a fresh stage} keep the random init \Comment{seeds the stage}
    \ElsIf{Pluralis worker} \Call{AwaitSafeWindow}{}; download state from a live same-stage peer
    \Else{} download the newest stage checkpoint from cloud storage \Comment{contributor}
    \EndIf
    \State $t_{\mathrm{loc}} \gets$ step stamp of the loaded weights $+$ optimizer state
    \State $t_{\mathrm{glob}} \gets$ current step of the stage
    \State $\mathit{mode} \gets \textsc{WeightSync}$ \textbf{if} $t_{\mathrm{glob}} - t_{\mathrm{loc}} > S$
           \textbf{else} $\textsc{Normal}$ \Comment{checkpoints lag: contributors start stale}
    \State \Call{AwaitSafeWindow}{}; announce on the stage's DHT, tagged $\mathit{mode}$
           \Comment{re-announced every 30\,s}
    \State start connection handlers, runtime \Comment{\Cref{alg:training}}
    \If{$\mathit{mode} = \textsc{WeightSync}$} \Comment{trainers skip this worker: no batches, no gradients}
        \State \textbf{for} $N_1$ steps: averaging weight $0$, report $0$ samples, serve no state
               \Comment{absorb the average only}
        \State $\mathit{mode} \gets \textsc{OptimizerSync}$ \Comment{visible to trainers again}
        \State \textbf{for} $N_2$ steps: process real batches; still weight $0$, $0$ samples
               \Comment{warms the Adam moments}
        \State $\mathit{mode} \gets \textsc{Normal}$ \Comment{full participant}
    \EndIf
\EndProcedure
\Statex
\Procedure{AwaitSafeWindow}{} \Comment{never join mid-round; averaging fires every $A$-th step}
    \While{$(t_{\mathrm{glob}} \bmod A) < W$ \textbf{or} $(t_{\mathrm{glob}} \bmod A) \ge A - W$}
        \State wait; refresh $t_{\mathrm{glob}}$ from the DHT
    \EndWhile
\EndProcedure
\end{algorithmic}
\end{algorithm}

\paragraph{Connection handlers and task queues.} Internally, a worker is a compute loop fed by the trainers. Connection handlers (multiple processes listening for RPC calls over libp2p) receive requests from trainers and place each incoming microbatch onto one of two queues, depending on whether it is a forward- or backward-pass request. Together, the multiple handlers and the queues let a worker receive and accumulate requests from several trainers at once without blocking compute, hiding communication overhead behind computation. Combined with the trainer-side load balancing (\Cref{trainers_load_balancing}), this keeps the pipeline's compute resources efficiently utilized.

\paragraph{Batch coalescing.}\label{batch_coalescing} Because workers are heterogeneous (some have less VRAM than others and can only process small batches) and because the trainer should be able to route each sample to any worker in the stage, trainers dispatch microbatches with only a single sample at a time. The batch-coalescing mechanism on each worker then merges the requests that queue up while it is busy processing the current microbatch, up to a maximum microbatch size set according to the worker's VRAM. Batch coalescing therefore utilizes GPU memory efficiently on heterogeneous nodes. Coalescing can only merge batches from the same queue (either forward or backward); the choice of queue is always based on which one holds the oldest request. Once the computation pass completes, results are split back to their corresponding requests and sent back to the trainers. One downside of this variable microbatch size is that the compiled forward/backward graph must accommodate a dynamic batch dimension rather than specialize to a single fixed shape, forgoing some of the speed-up a static shape would allow.

\paragraph{Forward and backward pass.} The Runtime, the worker's main loop, dequeues the microbatches and dispatches them to the ModuleBackend, which holds the stage's \texttt{nn.Module} and its forward and backward task pools. A forward request runs the stage's layers and returns the activations; a backward request first rebuilds the autograd graph by re-running the forward, recomputing activations rather than retaining them in memory between the two requests. This recomputation trades off compute for VRAM savings, as activations do not need to be cached until the corresponding backward request arrives. \Cref{alg:training} illustrates the worker training loop.

\begin{algorithm}[t]
\caption{\textbf{Worker training loop.} A swarm-wide optimizer step is coordinated
purely through sample counts published on the DHT: parameter gradients
never leave a node; replicas are reconciled by weight averaging
(\Cref{alg:averaging}) instead. Pluralis-8B values: target batch
$B{=}2048$ samples per step, averaging interval $A{=}20$ steps.}
\label{alg:training}
\small
\begin{algorithmic}[1]
\Procedure{ConnectionHandler}{} \Comment{$\times 7$ ingress processes, libp2p RPC}
    \While{alive}
        \State receive a forward / backward request from a trainer
        \State enqueue its microbatch into the forward / backward queue
        \State reply with the microbatch's result once computed
               \Comment{activations / input gradients}
    \EndWhile
\EndProcedure
\Statex
\Procedure{Runtime}{} \Comment{the single compute thread}
    \While{alive}
        \State $q \gets$ the queue holding the oldest pending microbatch
               \Comment{forward and backward share the GPU}
        \State $\mathit{batch} \gets$ coalesce $q$'s queued (FIFO) microbatches, up to the max batch size
        \If{$q$ is the forward queue}
            \State $\mathit{result} \gets$ forward pass of the stage's blocks on $\mathit{batch}$
        \Else
            \State recompute $\mathit{batch}$'s forward pass, then backpropagate its received
                   activation gradients
            \State $\mathit{result} \gets$ input gradients;\quad \Call{OptimizerStep}{$|\mathit{batch}|$}; zero module gradients
        \EndIf
        \State split $\mathit{result}$ per microbatch; hand back to the handlers
    \EndWhile
\EndProcedure
\Statex
\Procedure{OptimizerStep}{$b$} \Comment orchestrates training and worker's state transitions
    \State handle sync-phase transitions; catch $t_{\mathrm{loc}}$ up to $t_{\mathrm{glob}}$
           \Comment{\Cref{alg:join}}
    \State apply a finished averaging round, if one is pending \Comment{\Cref{alg:averaging}}
    \If{$b > 0$ \textbf{and} $\mathit{mode} \neq \textsc{WeightSync}$}
        \State accumulate the stage's parameter gradients \Comment{on GPU}
        \State publish own sample count to the stage's DHT progress key \Comment{$0$ while syncing}
    \EndIf
    \If{stage-wide samples this step $\ge B$} \Comment{summed from the DHT}
        \State $\bar g \gets$ accumulated gradients $/$ samples, moved to the CPU optimizer
               \Comment{gradients VRAM$\to$RAM}
        \State clip $\bar g$; AdamW step on weights on CPU; advance LR schedule; weights CPU$\to$VRAM 
        \Comment{synchronous}
        \State $t_{\mathrm{loc}} \mathrel{+{=}} 1$
        \If{$t_{\mathrm{loc}} \bmod A = 0$ \textbf{and} no averaging round in flight}
            \State \Call{LaunchAveraging}{} \Comment{\Cref{alg:averaging}; returns immediately}
        \EndIf
        \State publish $t_{\mathrm{loc}}$ to the DHT \Comment{the peers' maximum defines $t_{\mathrm{glob}}$}
    \EndIf
\EndProcedure
\end{algorithmic}
\end{algorithm}

\paragraph{Visibility.} The nodes use the distributed hash table (DHT) to announce themselves and become visible to the trainers and other peers in their stage. A dedicated thread continually re-announces the worker under a stage-prefixed identifier, which trainers read to locate workers and which same-stage peers use to find one another. 

\paragraph{Rule-based worker self-termination.} Each worker runs a local log monitor that evaluates its own execution against a declarative set of rules, and terminates itself when a rule signals an unrecoverable condition. Some violations trigger immediate termination: a CUDA failure, invalid values in the gradients, loss of peer-to-peer connectivity, or an inability to fetch the run's global progress. Others are thresholded: a worker whose host RAM usage exceeds a configured limit exits before it degrades, and a worker that fails two of its last three all-reduce rounds terminates itself, since a replica that repeatedly misses averaging drifts from the stage consensus and harms convergence (\Cref{sec:ablations}); successful rounds reset the failure count. Self-termination keeps enforcement local; thus, no central component needs to observe a failing node or force it out, which matters when the operator has no privileged access to a contributor's machine. A terminated worker may rejoin through the normal admission path, re-entering via the sync phases of \Cref{sec:worker_state_load_sync_phase}.

\paragraph{Weight averaging.} Workers within the same stage periodically exchange their state via all-reduce weight averaging (the data-parallel axis) to keep the replicas' weights from drifting apart. This averaging is triggered at a fixed step interval and runs asynchronously (weights are first copied to CPU memory) so it stays off the compute critical path. Once the averaging round is complete, the averaged weights are then copied back to VRAM. Optimizer states are not exchanged to conserve bandwidth.

\begin{figure}[t]
    \centering    \includegraphics[width=0.8\linewidth]
    {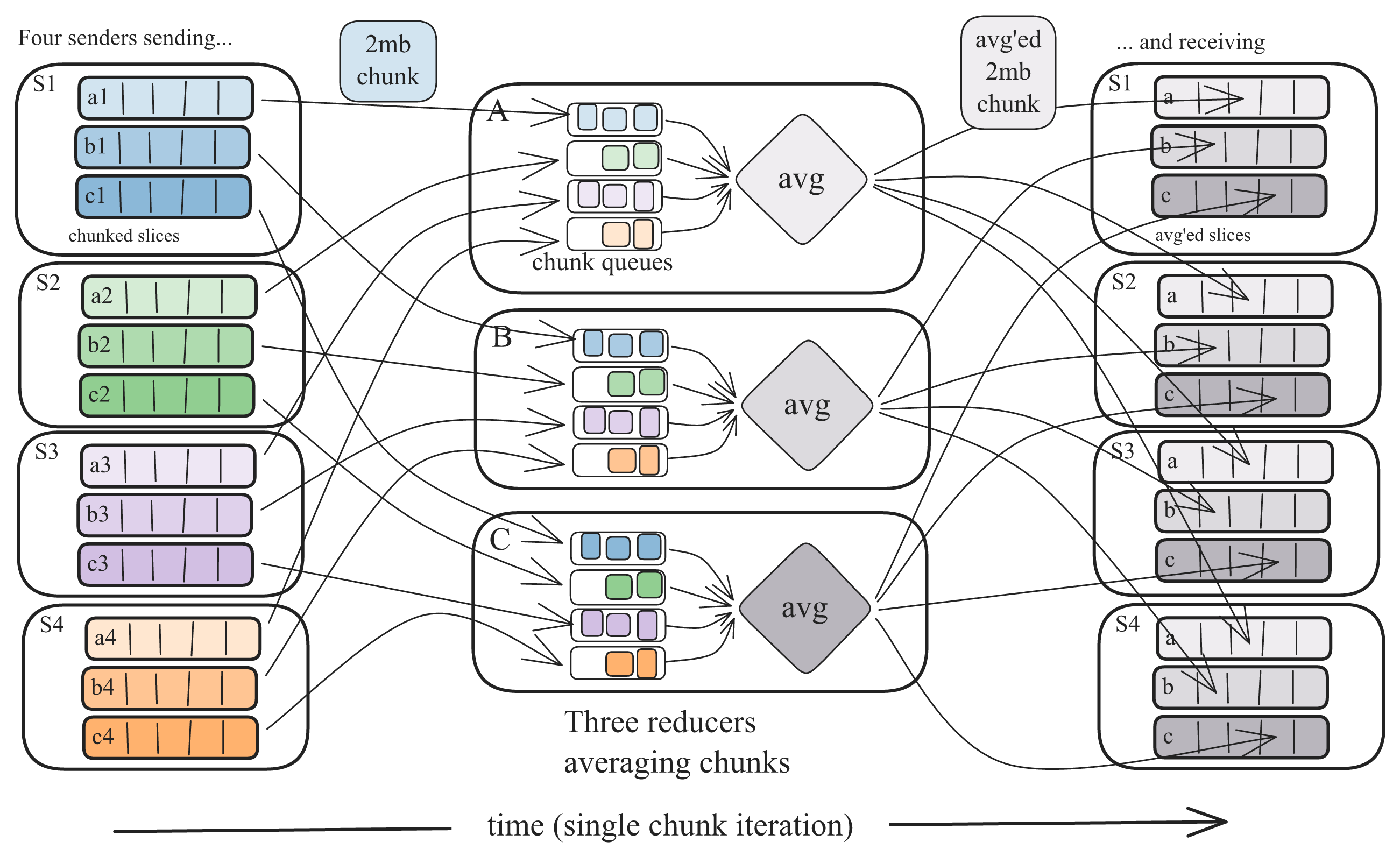}
    \caption{\textbf{Diagram of the all-reduce averaging algorithm.} One chunk iteration of an
    averaging round with four senders (S1--S4) and three reducers (A--C). Each peer's flattened
    parameter vector is partitioned into slices, one per reducer (a, b, c), and every peer streams
    its copy of each slice, cut into small fixed-size chunks (2\,MiB), to the slice's assigned reducer.
    A reducer buffers incoming chunks in per-sender queues, averages each chunk across all senders
    (weighted by their processed samples), and streams the averaged chunks back, so every peer ends
    the round holding the fully averaged slices. Note that Pluralis workers act as senders
    \emph{and} reducers (they also contribute their own weights to the average), a dual
    role omitted from the figure for clarity; contributors act as senders only. For pseudo-code, refer to \Cref{alg:averaging}.}
    \label{fig:averaging_diagram}
\end{figure}

\paragraph{All-reduce algorithm.} The weight averaging uses a modified butterfly all-reduce algorithm, consisting of a single round: reduce-scatter followed by an all-gather (illustrated in \Cref{fig:averaging_diagram}, with pseudo-code in \Cref{alg:averaging}). A round begins with a matchmaking phase: each peer announces itself in the DHT and waits, up to a bounded matchmaking window, for the rest of the stage to register; the resulting group is then fixed and its members are ordered deterministically so that every peer derives the same layout. 
Only Pluralis nodes, which have fast, reliable network connections, are assigned the \emph{reducer} role, receiving data from all peers and performing the reduction (averaging), while contributors act only as \emph{senders}. This prevents an unreliable contributor from disrupting the round: were a reducer to leave mid-average, its slice could not be reduced and every peer would fall back to its own local values for that slice. 
The flattened parameter vector is partitioned into contiguous slices, one per reducer. Every peer streams its copy of each slice to the reducer responsible for it; each reducer computes a weighted average of the slices it receives (weighted by the number of samples each sender processed since the last round) and streams the results back. 
Because the exchanged state can be large (a single stage's weights may be several gigabytes), the data is cut into small, fixed-size chunks (optionally compressed) that are serialized in background threads and streamed over libp2p. Senders and reducers exchange chunks concurrently over the full-duplex connection, so a reducer overlaps computation with communication, accumulating each chunk as it arrives rather than waiting for the whole slice, saving time and memory. Reducers receive chunks into an inbound queue (per sender) to amortize network jitter.
For numerical stability, a reducer returns the \emph{difference} between the averaged and local slice rather than the average itself; each peer concatenates these differences into a full update and applies it to its averaged tensors. The averaging is fault-tolerant: a sender that misses its per-chunk timeout is banned and excluded from the round, and if the averaged slice fails to arrive back from the reducer, the peer falls back to its local values, so a round still finishes with a partially reduced state (\Cref{fig:average_pcnt} and \Cref{fig:reducer_pcnt} in \Cref{sec:results}).

\begin{algorithm}[t]
\caption{\textbf{Asynchronous sparse weight averaging, one worker's view.} Every $A$-th
stage-global step the workers of a stage average a rotating fraction $s$ of their
parameters and keep training while the round runs in the background; averaging rounds
never overlap (timeout is set below the duration of $A$ steps). Pluralis-8B values:
$s{=}5\%$ (consecutive rounds cover every parameter once per $1/s{=}20$ rounds, i.e.,
every 400 steps), matchmaking window 40\,s, 2\,MiB chunks, sender / reducer
timeouts 30\,s / 60\,s.}
\label{alg:averaging}
\small
\begin{algorithmic}[1]
\Procedure{LaunchAveraging}{} \Comment{from a step boundary, \Cref{alg:training}}
    \State $w \gets$ samples since last launch; reset counter
           \Comment{$0$ if $\mathit{mode} \neq \textsc{Normal}$}
    \State submit \Call{AveragingRound}{$w$} to a background thread
\EndProcedure
\Statex
\Procedure{AveragingRound}{$w$} \Comment{background; training continues}
    \State declare on the stage's matchmaking key; collect peers until the window lapses
           \Comment{early exit: whole stage declared}
    \State refresh FP32 $\mathit{buf}$ from CPU weights \Comment{CPU-offloaded optimizer}
    \State $P \gets$ fraction $s$ of parameters, rotated by step;\;
           $\mathit{snap} \gets \operatorname{copy}(\mathit{buf}[P])$ \Comment{same $P$ on every peer}
    \State freeze the group \Comment{deterministic order from each peer's DHT view, no leader}
    \State split $\mathit{buf}[P]$ into one contiguous slice per \emph{reducer};
           stream each slice to its reducer in fixed-size chunks
    \If{reducer of slice $\ell$}
        \State $\mathit{acc} \mathrel{+{=}} w_i\, c_i$ per arriving chunk $c_i$ from sender $i$;
               ban senders past the per-chunk timeout \Comment{round proceeds without them}
        \State all senders accounted: $\mathit{acc} \mathrel{/{=}} \sum_i w_i$;
               return each sender $\mathit{acc} - c_i$ \Comment{$w_i{=}0$: absorbs, adds nothing}
    \EndIf
    \State $\mathit{buf}[P] \mathrel{+{=}}$ received deltas \Comment{$\mathit{buf}[P]$ = group average}
    \State \textbf{on} reducer timeout: keep local values for its slice
           \Comment{a round always completes, possibly partially}
    \State mark round \emph{finished} \Comment{picked up by \Cref{alg:training}}
\EndProcedure
\Statex
\Procedure{ApplyFinishedRound}{} \Comment{from \textsc{OptimizerStep}, $N$ steps after launch}
    \State $\Delta \gets \mathit{buf}[P] - \mathit{snap}$;\;
           master weights$[P] \mathrel{+{=}} \Delta$;\; weights CPU$\to$VRAM
           \Comment{$\Delta$ lands on weights $N$ steps newer, timeout keeps $N{<}3$}
\EndProcedure
\end{algorithmic}
\end{algorithm}

\subsubsection{Trainer}
\label{trainer_section}

A trainer drives what looks like an ordinary training loop: embed the inputs, run the Transformer blocks, compute the loss, call backward --- except that none of these computations run locally. Holding no parameters and needing no GPU, the trainer dispatches every forward and backward over the network to a worker in the relevant stage, threading activations forward along the pipeline (head → body → tail) and activation gradients back (tail → body → head). On the forward pass it sends the batch to a load-balanced worker in the head stage, receives the activations, relays them to a load-balanced worker in the next stage, and so on to the tail, which returns the loss; activations never pass directly between workers: the trainer is the conduit between consecutive stages and retains the activations it threaded at each boundary. On the backward pass it walks the stages in reverse, sending each stage's saved input together with the incoming activation gradient back to the worker that ran that stage's forward. That worker rebuilds its autograd graph by recomputing the forward, returns the gradient with respect to its input (which the trainer forwards to the previous stage), and accumulates the parameter gradients locally. Therefore, a worker does not need to keep any per-request state between the two passes, which makes it easily replaceable.

\paragraph{Data sharding and sequential batch processing.} At startup, a trainer loads its configuration and tokenizer, prepares its shard of the dataset, and connects to each stage's DHT. When every stage advertises at least one available worker, the training can begin. For each batch it walks the stages in order, dispatching the forward chain and then a single backward that unwinds through all stages, with the tail stage computing the loss against the shifted next-token labels. Each trainer is assigned a shard of the dataset that it runs until exhaustion. Because trainers process batches from their shards sequentially (each trainer keeps only one batch in flight), the pipeline is saturated not by any single trainer but by many running concurrently, whose requests pile up at each stage's workers and are merged by batch coalescing (\Cref{batch_coalescing}). Trainers can be added or removed during the training run (\Cref{trainers_saturation}).

\paragraph{Worker discovery.} The trainer does not have a fixed list of workers; it learns each stage's workers from that stage's DHT. For every stage it runs a balancer whose background thread checks the DHT about every 30 seconds to see which workers are currently online. Each worker keeps re-advertising itself in the DHT with a short expiry, so a new worker is picked up at the next check and added, while a worker that goes offline stops updating its entry in the DHT and is dropped. The trainer also skips workers that are in \emph{sync phase} (\Cref{sec:worker_state_load_sync_phase}), and temporarily bans workers that have failed to process a batch.

\paragraph{Load-balanced forward pass and deterministic backward pass.}\label{trainers_load_balancing} Within a stage, the trainer selects a worker by load. The stage's workers are kept in a min-heap keyed by accumulated virtual runtime, and each microbatch's forward is dispatched to the least-loaded worker --- the heap minimum. When a worker is selected, its runtime is incremented up front by the task's estimated cost (roughly the task size divided by the worker's measured throughput, so a slower replica is charged more per request) and it is pushed back into the heap before the call returns; the key is thus a cumulative, speed-weighted account of how much work each replica has been given, and every completed call's measured time updates the throughput estimate used for later charges. Work therefore spreads across a stage's replicas in proportion to their speed, with faster workers drawing more microbatches. The backward pass is not load balanced: it follows the same path the microbatch took during the forward pass. A newly discovered worker does not enter the heap at zero, but is placed level with the most-loaded replica already present. Failures are handled by trainers using retries and temporary peer-banning policies.

\subsection{Throughput Optimization and Scaling}
\label{ssec:throughput_optimization}

Unlike centralized training, where accelerators share high-speed interconnects, Agora must contend with high-latency, low-bandwidth links between participating nodes -- the major obstacle to competitive throughput. Much of the challenge lies in configuring the system, under these communication constraints, so that the contributed compute stays maximally utilized rather than idling while it waits for data to process.

Since Node0~\cite{avraham2025node0}, we have made several throughput improvements, among them a fully asynchronous optimizer that runs weight averaging in the background, as well as improvements in mixed-precision compute and computation-graph compilation, all of which let Agora nodes process batches faster. 

The subsections below examine, in turn, the communication window that governs per-node utilization, the tradeoffs in scaling compute along each axis, the balance between stage types, and the trainer count required to saturate the pipeline. We close with a theoretical analysis of how these constraints evolve with model scale (\Cref{sec:scaling_analysis}), showing that they relax, rather than tighten, as models grow.

\subsubsection{Communication Window}
\label{sec:comms_window}

Communication of the compressed forward-pass activations and backward-pass gradients is overlapped with microbatch forward or backward computation and must complete within the computation window; otherwise GPU utilization and throughput suffer. Processing batches faster lets us target higher token throughput (TPS), but it also shrinks this per-microbatch window and leaves less time in which to hide communication. The objective is to avoid crossing from a compute-bound regime into a communication-bound one, where nodes sit idle waiting on data (\Cref{fig:fig_tps_combined}), and instead to operate just inside the compute-bound regime, with communication fully overlapped but close to saturating its window, so that both TPS and utilization are maximized, and the activations/gradients compression rate is no larger than necessary.

\subsubsection{Scaling Compute Resources}
\label{sec:scaling_compute}

In Agora, total compute capacity (aggregate FLOPs) can be scaled along two axes, each with its own tradeoffs: vertically, by adding more nodes per stage, and horizontally, by adding more stages. 

On the vertical axis, adding nodes also adds all-reduce participants, raising communication overhead (more parallel connections per round) and increasing the risk of communication faults.

On the horizontal axis, splitting a fixed total parameter count across more stages leaves each stage replica holding fewer parameters, so it processes its microbatches faster, which in turn shrinks the communication window for activations and gradients. Note that both communication and computation time grow with the microbatch size, so the microbatch is not an effective lever for changing the ratio between them. We therefore gate contributors on a minimum bandwidth and latency before they may join the run, such that the worst-case per-microbatch communication time (given the model's hidden dimension, sequence length, and the communications compression rate) fits the estimated per-microbatch compute time. How this compute-to-communication ratio evolves as the model itself grows is analyzed in \Cref{sec:scaling_analysis}.

Adding stages also lengthens the batch orchestration path: because each microbatch returns to the trainer before being dispatched to the next stage's worker, the number of trainer--worker round-trips per microbatch grows linearly with the stage count, and with it the accumulated network latency. This can be amortized by using more trainers (see \Cref{trainers_saturation} below); however, more trainers increase the number of distributed hash table (DHT) nodes, and DHT performance and latency degrade non-linearly as that count grows.

Beyond throughput, the choice of stage count has a large impact on state loading and on weight averaging. Dividing the model into fewer but larger stages means a joining node takes longer to download its stage's state, so it begins training with staler weights. Larger stages also degrade weight averaging: the number of parameters to be averaged grows, producing longer all-reduce (AR) rounds with larger payloads and a higher risk of communication failures. Such failures cause partial averaging, which increases weight divergence between replicas and harms convergence.

\subsubsection{Stage Bottlenecks}
\label{sec:stage_bottlenecks}

We divide the pipeline into a head stage, body stages, and a tail stage. Contributors may join only the body stages; the head and tail run on stable cloud compute with high bandwidth. The throughput dynamics between the three are therefore non-trivial, and the allocated compute must be balanced so that none of the three is under-utilized or bottlenecking the others. To this end we monitor four metrics within each of the three stage categories: microbatch data-await time, microbatch compute time, coalesced microbatch size, and GPU utilization. The ratio of the first two reveals whether a stage is communication-bound (await time exceeding compute time), a diagnosis corroborated by low GPU utilization and by coalescing that never accumulates beyond the minimum microbatch size.

Using these metrics, we tune the compute at the head and tail (the stages we operate ourselves) to keep every stage optimally utilized. The tail proves the heaviest of the three: its output projection over the full vocabulary and the cross-entropy loss make it the most compute-intensive stage, and it is therefore the one we provision most heavily.

\subsubsection{Trainers Saturation}
\label{trainers_saturation}

\begin{figure}[t]
  \centering
  \begin{subfigure}[t]{0.27\linewidth}
    \centering
    \includegraphics[width=\linewidth]{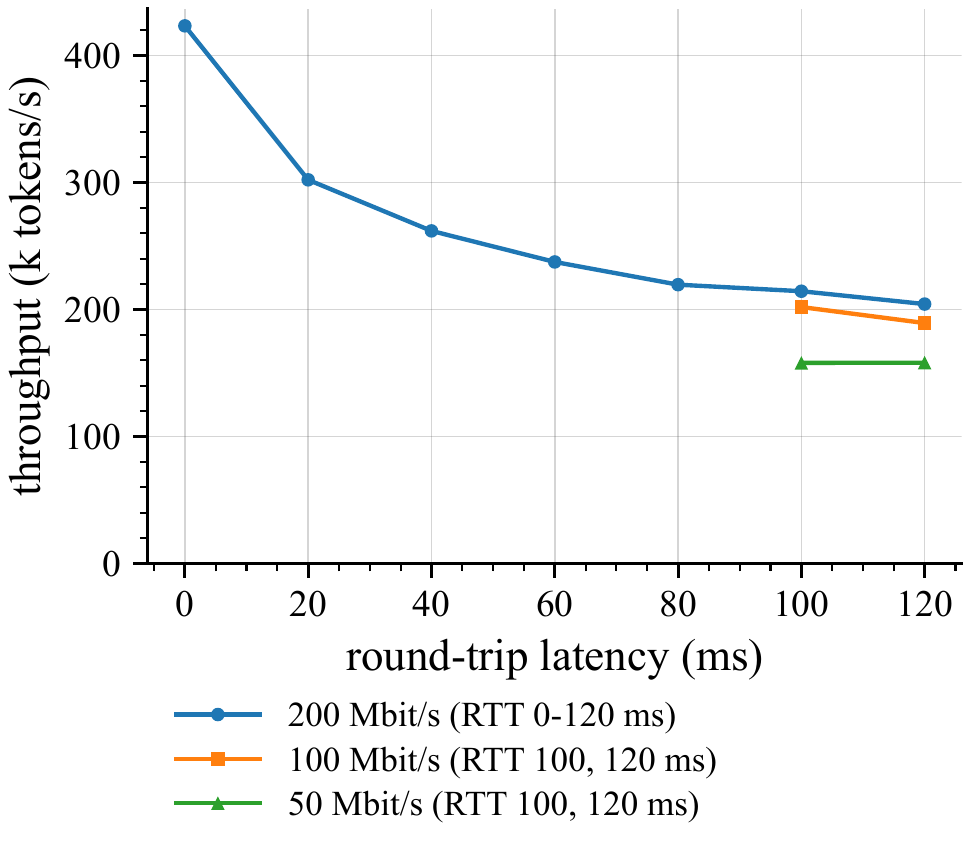}
    \caption{}
    \label{fig:tps_regime}
  \end{subfigure}\hfill
  \begin{subfigure}[t]{0.35\linewidth}
    \centering
    \includegraphics[width=\linewidth]{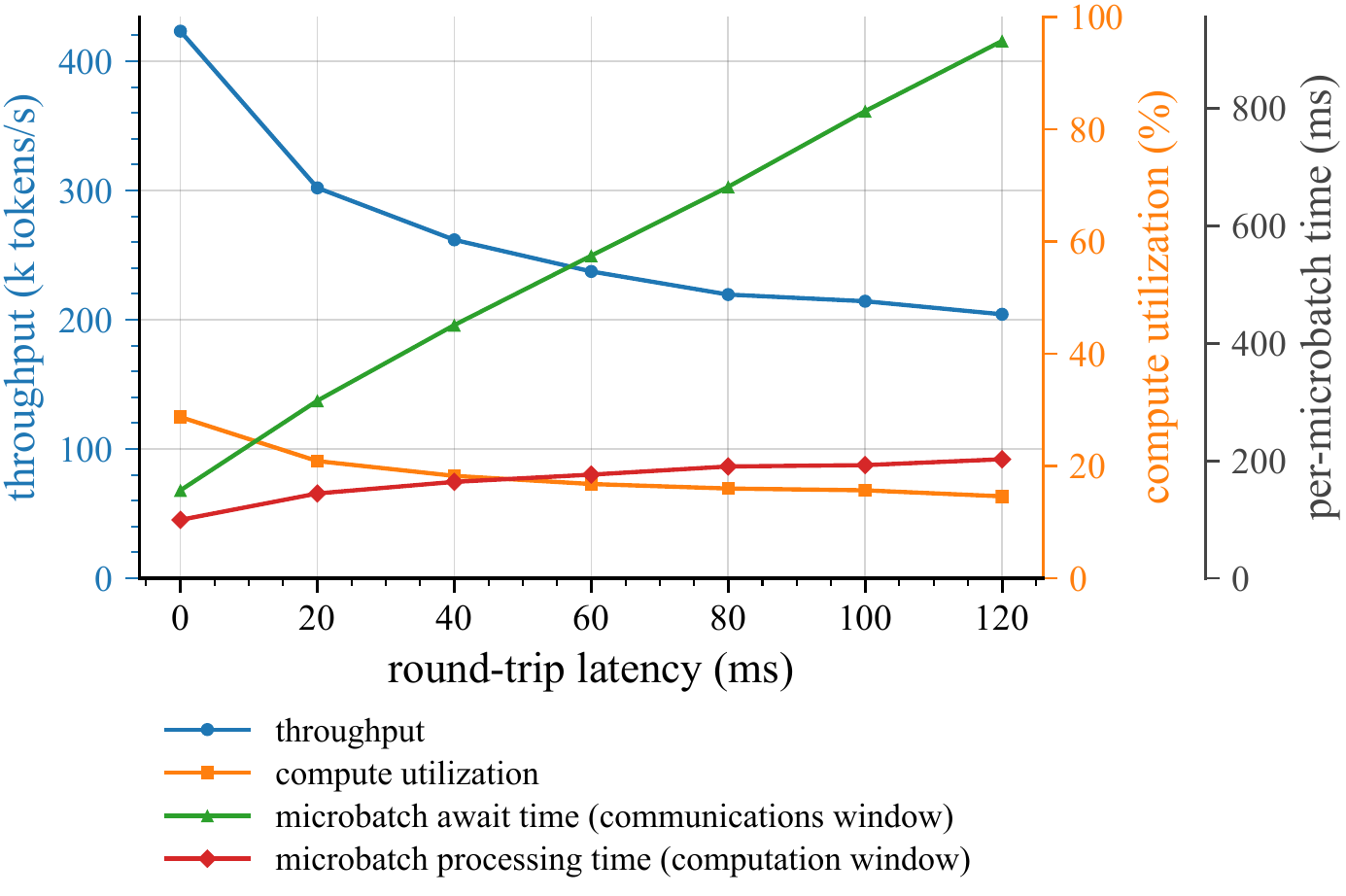}
    \caption{}
    \label{fig:tps_window}
  \end{subfigure}\hfill
  \begin{subfigure}[t]{0.35\linewidth}
    \centering
    \includegraphics[width=\linewidth]{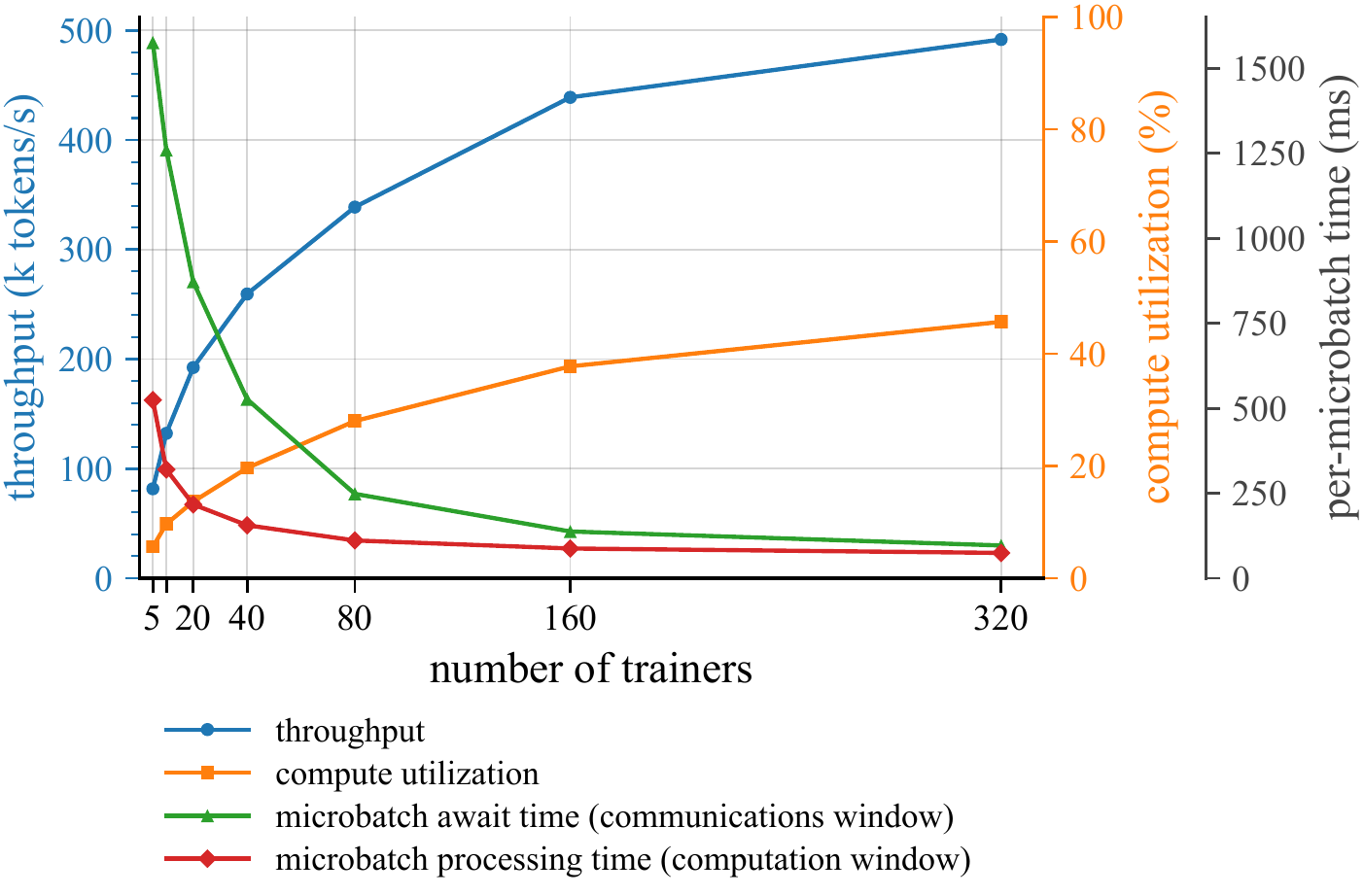}
    \caption{}
    \label{fig:tps_trainers}
  \end{subfigure}
  \caption{\textbf{Impact of network quality and trainer count on throughput.} \textbf{(a)} A pipeline stage operating in a communications-bound regime: 4 Pluralis (high-bandwidth) nodes and 16 contributor nodes with different latency and bandwidth restrictions (simulated). \textbf{(b)} When communication time exceeds the computation window, utilization and throughput suffer. \textbf{(c)} The total compute capacity of the pipeline needs enough trainer instances to reach high utilization (single pipeline stage with 4 Pluralis high-bandwidth nodes and 16 contributor nodes with simulated 100\,ms RTT latency and 200\,Mbit/s bandwidth). As the number of trainers increases, nodes' microbatch await time reduces, transitioning them into compute-bound regime, at which point adding more trainers no longer improves throughput.}
  \label{fig:fig_tps_combined}
\end{figure}

The final lever is the number of trainers. A trainer drives one microbatch through the pipeline at a time, and that microbatch occupies only one stage at any given instant, so a single trainer would leave every other stage idle. Keeping all workers busy therefore requires enough trainers in flight that every stage always has a microbatch to process, which we refer to as saturating the pipeline (\Cref{fig:tps_trainers}). The number needed is hard to predict in advance, particularly under heterogeneous, low-bandwidth participation: when transfers are slow, a larger fraction of each trainer's wall-clock time goes to sending and receiving data rather than to keeping a worker computing, so more trainers are needed to hide that latency and cover the same set of workers. We therefore determine the saturation point empirically, adding trainers until the marginal gain in TPS diminishes.

\subsubsection{Scaling Analysis}
\label{sec:scaling_analysis}

The preceding subsections treat the model as fixed and ask how to 
configure the system around it. We now ask the converse: as the model 
itself grows, do the communication constraints of 
\Cref{sec:comms_window} tighten or relax? A simple 
accounting of compute and traffic per stage shows that they relax, on 
both parallelism axes, so the regime validated at 8B scale becomes 
easier to sustain at larger scale. 

Consider a stage of $L_s$ Transformer blocks with hidden dimension
$d$, processing sequences of length $n$. The stage's parameters are
dominated by its projection matrices, roughly $12 L_s d^2$ values
(\Cref{ssec:async_sparta}), so its forward or backward compute grows
as $\Theta(n L_s d^2)$. The signal crossing a stage boundary, however,
is one vector per token compressed at a fixed rate $c$:
$\Theta(n d / c)$ values. The ratio of per-stage compute to boundary
communication therefore grows as $\Theta(L_s d c)$, linearly in the
hidden dimension and in the stage depth. Equivalently, the bandwidth a
node requires to keep communication hidden inside its computation
window \emph{falls} as $1/(L_s d)$ at fixed accelerator speed:
widening or deepening the model pushes every stage further into the
compute-bound regime that \Cref{sec:comms_window}
targets. Doubling the hidden dimension, for instance, doubles the
boundary traffic but quadruples the per-stage compute, halving the
bandwidth needed per unit of contributed FLOPs. 

This holds the compression rate $c$ fixed as $d$ grows, which is the
pessimistic assumption. In practice, the rank we can use appears to grow
more slowly than the width: the 1B model uses rank 40 at $d = 2048$,
and Pluralis-8B rank $\approx$51 at $d = 5120$, a $1.3\times$
increase in rank against a $2.5\times$ increase in width, so the
effective compression rate was higher at 8B than at 1B. We do not claim
a scaling law from two configurations, and the rank is a hyperparameter
rather than a fixed property of the architecture, but under this trend
the ratio improves faster than the fixed-$c$ estimate of
$\Theta(L_s d c)$. The attention-score computation, which the
projection accounting omits, behaves the same way: it contributes
$\Theta(n^2 d)$ per block against the projections'
$\Theta(n \cdot 12d^2)$, a ratio of $n/12d$, shrinking as the model
widens at fixed sequence length. Should sequence length grow with
scale, attention adds compute that is entirely local to a stage,
crossing no boundary, and so only strengthens the
compute-to-communication asymmetry.

On the data-parallel axis, the sparse averaging payload grows with the
stage parameters, $p \cdot 12 L_s d^2$ per round, 
where $p$ is the
fraction of a stage's parameters exchanged in each averaging round
(\Cref{ssec:async_sparta}). But the window available to the
asynchronous all-reduce grows in the same proportion: a round must
complete within $N$ optimizer steps, and at a fixed batch size and
worker count, the duration of a step is itself set by the stage's
$\Theta(L_s d^2)$ compute. The all-reduce duty cycle, payload over
window, is therefore scale-invariant, and the per-contributor
bandwidth floor it implies does not grow with the model. Since the
exchange is asynchronous and off the compute path, 
data-parallel communication does not
re-enter the critical path at any model size.

The accounting so far is stated in bandwidth, but in the live run,
throughput was bound by latency and GPU class, with bandwidth flat
above the admission floor.
Scale relaxes this constraint too. A stage boundary's round-trip is a
fixed per-microbatch cost, independent of $d$, that must fit inside the
same computation window the bandwidth analysis grows. As per-stage
compute rises with width, a fixed round-trip occupies a smaller
fraction of that window, so latency is easier to hide at larger scale
as well.

These per-stage results compound under the way large models are
actually scaled. Model families grow depth alongside width, keeping
the aspect ratio $d/L$ within a narrow band, so $L \propto d$ and
total training compute per token, $\Theta(L d^2)$, grows as
$\Theta(d^3)$, while the signal crossing any stage boundary grows
only as $\Theta(d)$. Holding the stage count fixed, each stage's depth
likewise grows as $L_s \propto d$, so per-stage compute is
$\Theta(n d^3)$ against $\Theta(n d / c)$ boundary traffic, and the
compute-to-communication ratio improves as
$\Theta(L_s d c) = \Theta(d^2)$, with no growth in trainer
round-trips, DHT traffic, or averaging-group size.
In practice, VRAM bounds how many blocks a stage can hold, so scale is also absorbed by adding stages;
the per-boundary ratio then improves as $\Theta(d)$ rather than $\Theta(d^2)$.
This is the one point at which scale is not purely favorable: adding
stages is what lengthens the orchestration path of
\Cref{sec:scaling_compute}, since each microbatch returns to its
trainer at every boundary, so the trainer round-trips per microbatch
and the DHT membership both grow with the stage count. Deepening fixed
stages avoids this and keeps the $\Theta(d^2)$ rate; adding stages
trades part of the gain back for that coordination cost.
The stage imbalance of \Cref{sec:stage_bottlenecks} fades with
scale in either case: the tail's additional burden, the output
projection over the vocabulary, costs $\Theta(n d V)$ against the
blocks' $\Theta(n L_s d^2)$, so its relative weight shrinks as
$1/(L_s d)$ and the tail's outsized provisioning becomes a small-model
artifact.

This analysis covers the compute and communication volumes that govern
throughput; it does not claim that every component scales freely. The
costs attached to larger stages -- state-loading time for joining
nodes and the absolute all-reduce payload
(\Cref{sec:scaling_compute}) -- grow with stage size, as do
the coordination loads studied in \Cref{ssec:results_collective_comms},
and convergence under sparse averaging constrains the replica count (\Cref{ssec:async_sparta})
independently of model scale. Within those bounds, however, the
direction of scale works in the protocol's favor: the larger the
model, the more compute each transmitted byte requires.

\subsection{Fault-Tolerance and Stability}

\subsubsection{Surrogate Testing Framework}
\label{testing_framework}

\paragraph{CPU testing.} To examine how system stability and fault tolerance behave under different conditions, we employ several techniques to make smaller models representative of larger models along multiple axes. In particular, we replace Transformer blocks with simple linear layers that can also emulate a single pipeline stage independently. This design offers several benefits: tests can run entirely on CPU devices, making it a cost-effective alternative, and the step time can be fully controlled by injecting delays during the forward and backward passes or by varying the target batch size. These surrogate models are then made representative of larger models along the data-parallel communication axis by introducing non-inferenced dummy parameters to increase the all-reduce payload. Additionally, communications in the pipeline-parallel axis are replicated by matching the required sequence length, $\mathrm{seq\_len}$, and hidden dimension of the activations, $\mathrm{hidden\_dim}$.

\subsubsection{Stability Under Bandwidth Constraints}

Bandwidth and latency considerations are crucial for maintaining system throughput (\Cref{fig:fig_tps_combined}) and system stability with first-order effects on activation and all-reduce transfers, as well as second-order effects in all-reduce matchmaking via slow DHT operations and consistency. Hence, it is crucial to determine suitable bandwidth and latency requirements for contributors to balance system stability and accessibility of participation. 

\paragraph{Estimated communication.} We first size the per-node bandwidth floor by considering three types of high-volume traffic a contributor must sustain: forward activations, backward activations, and a periodic all-reduce burst. A contributor in a body stage must communicate $a_{f}$ and $a_{b}$ (batched) activation payloads per second of size $[\mathrm{seq\_len}, \mathrm{hidden\_dim}]$ on average for forward and backward transmission in FP32, along with $\mathrm{AR\_payload}$ parameters in $\mathrm{AR\_dtype}$ (BF16 or FP32) during all-reduce. 

Thus, a contributor may be required to send and receive,
\[
32 \times (a_f + a_b) \times \mathrm{seq\_len} \times \mathrm{hidden\_dim} + \mathrm{AR\_dtype} \times \frac{ \mathrm{AR\_payload}}{\mathrm{AR\_time}}
\]
bits per second in the worst case. Substituting the suggested model parameters for Pluralis-8B, writing in the case of SPARTA (\Cref{ssec:async_sparta}), $\mathrm{AR\_payload} = \mathrm{sparse\_avg} \times \mathrm{num\_params}$, then estimating $a_f = 3$, $a_b = 3$, and $\mathrm{AR\_time} = 10$ from internal testing under ideal conditions, we obtain a required bandwidth of $\sim$113\,Mbit/s under a BF16 all-reduce or $\sim$186\,Mbit/s under an FP32 all-reduce. Indeed, by estimating a worst-case $a_f$, $a_b$, and $\mathrm{AR\_time}$ under optimal compute settings, we have a conservative upper bound on the required bandwidth, noting this estimate does not model network variability, jitter, latency spikes, or transient packet loss, which can dramatically affect stability. However, this worst-case parameterization provides some headroom against these effects.

\paragraph{Stability testing setup.} We perform stability checks on a grid of network settings, varying both bandwidth and latency using the surrogate testing framework following the previous estimate. All nodes are instantiated in the same availability zone (AZ) with network shaping applied to simulated contributor nodes to accurately model latency to our workers. These tests are performed with 4 workers and 16 simulated contributors with communication matching the pipeline-parallel axis (forward, backward activation size) and the data-parallel axis (all-reduce payload) of Pluralis-8B in FP32 and BF16, respectively. Here, we explicitly simulate contributors by ensuring they only contribute parameter chunks and then receive back reduced chunks, without performing the reduce operation.

In particular, we sweep RTT latencies over $\{0, \, 20, \, 40, \, 60, \, 80, \, 100, \, 120\}$\,ms with upload and download bandwidth at 200\,Mbit/s. Additionally, we test lower bandwidths of 50\,Mbit/s and 100\,Mbit/s at 100\,ms RTT latency. Over these settings, we measure the duration of an asynchronous state all-reduce, the all-reduce completeness as the percentage of reduced parameter chunks received back by a worker or contributor, and the percentage of failed all-reduce rounds, measured when completeness for a worker or contributor is less than 100\%.

\begin{figure}[!t]
    \centering
    \begin{subfigure}{0.32\linewidth}
        \centering
        \includegraphics[width=\linewidth]{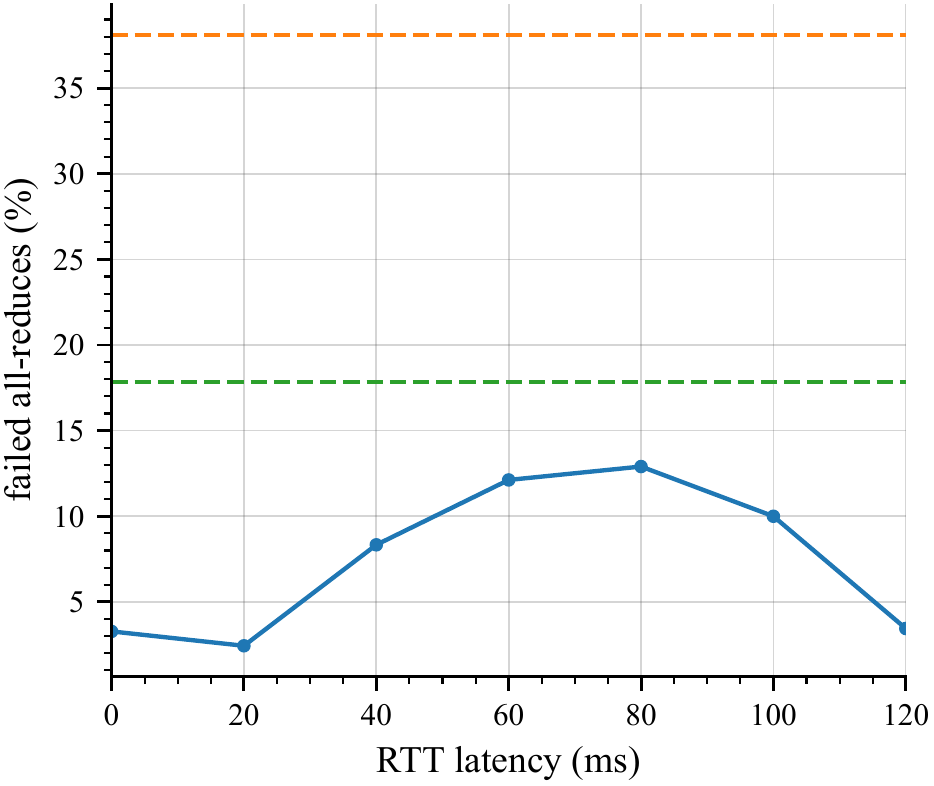}
        \caption{Failed all-reduces.}
        \label{fig:ar_failed}
    \end{subfigure}\hfill
    \begin{subfigure}{0.32\linewidth}
        \centering
        \includegraphics[width=\linewidth]{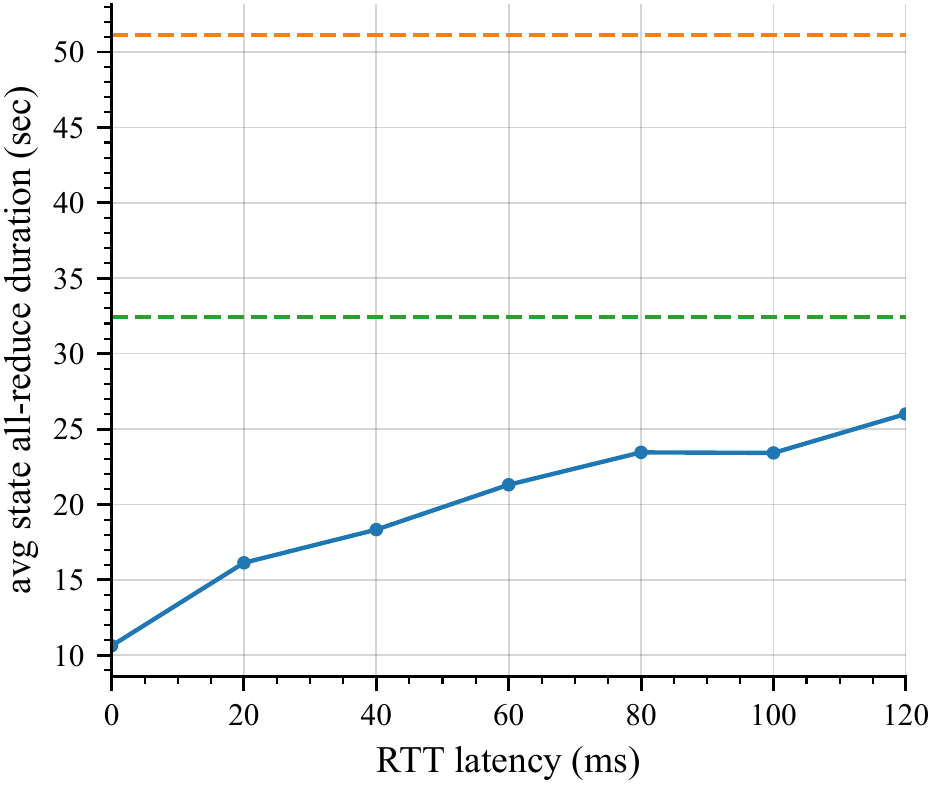}
        \caption{All-reduce duration.}
        \label{fig:ar_duration}
    \end{subfigure}\hfill
    \begin{subfigure}{0.32\linewidth}
        \centering
        \includegraphics[width=\linewidth]{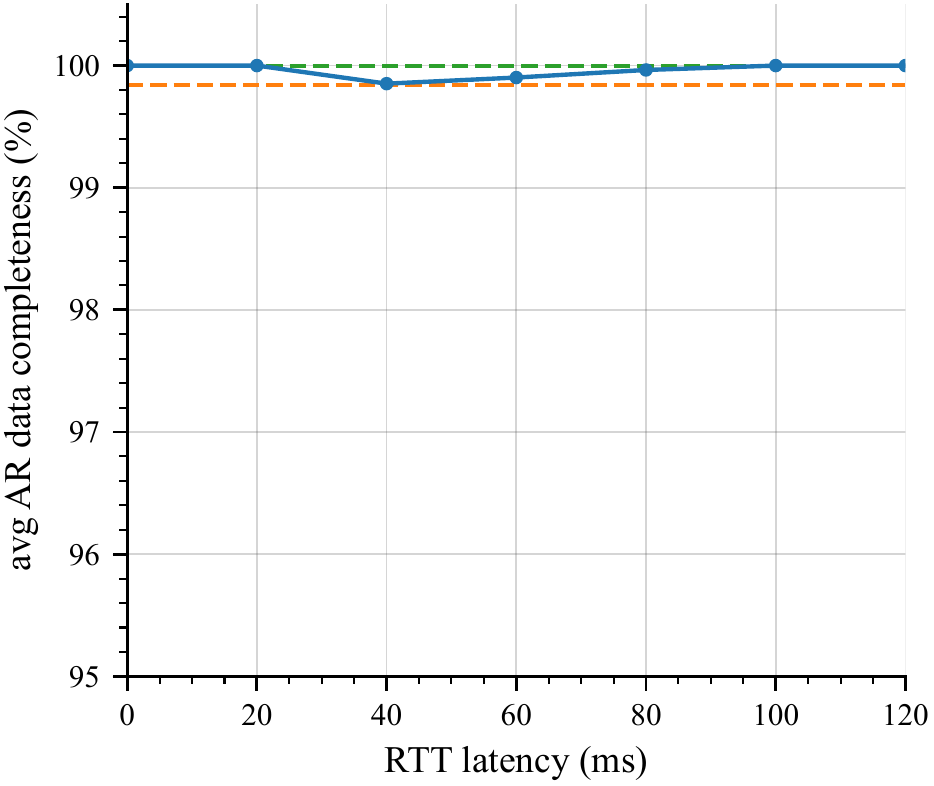}
        \caption{All-reduce completeness.}
        \label{fig:ar_completeness}
    \end{subfigure}\\[3pt]
    \includegraphics[width=0.75\linewidth]{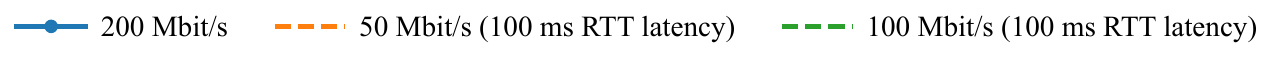}
    \caption{\textbf{All-reduce stability, duration, and completeness under varying network conditions.} \textbf{(a)} Percentage of failed all-reduces as a function of RTT latency. \textbf{(b)} Average all-reduce duration (seconds) over the same range. \textbf{(c)} Average percentage of all-reduce completeness over the same range. The solid curve sweeps RTT latency over 0--120\,ms at 200\,Mbit/s; the dashed horizontal lines mark reference runs at 50\,Mbit/s and 100\,Mbit/s, both at 100\,ms RTT. At 200\,Mbit/s, failure rate remains below 15\% and AR duration below 30\,s across a range of latencies, below lower-bandwidth baselines. All show near-perfect completeness.}
    \label{fig:ar_vs_latency_bandwidth}
\end{figure}

\paragraph{Stability testing results.} Our tests reveal that at 200\,Mbit/s the system remains stable and remains below the system timeouts at a range of latencies, as reflected by a $<$15\% failure rate in all-reduce as shown in \Cref{fig:ar_failed}. In comparison, we find elevated instances of failures on 100\,Mbit/s and 50\,Mbit/s systems, indicating the sensitivity of all-reduce to variable bandwidths. Indeed, without further modeling of network variability and jitter, it is crucial that the system is able to complete the required all-reduce within a practical time frame, as enforced by timeouts, as failure under these idealized conditions guarantees failure for contributors operating in the live setting. This is shown by measuring the asynchronous all-reduce duration in \Cref{fig:ar_duration}, which grows predictably under higher latencies to $>$2.5$\times$ the baseline at 120\,ms RTT, introducing further risk of timeouts under unpredictable networking conditions.

Interestingly, as latency increases, the all-reduce failure percentage peaks and decreases, as a result of lower TPS (\Cref{fig:tps_regime}) which frees up bandwidth for all-reduce from activation communication.

While we find the 200\,Mbit/s bandwidth to be adequately stable under a range of RTT latencies, the failure rate may be too high for practical use in traditional distributed training setups, and would only be exacerbated under the less predictable conditions faced by contributors. However, with Agora's fault-tolerant all-reduce architecture (\Cref{worker_section}), the system may continue with a partial all-reduce, which we see in \Cref{fig:ar_completeness}. Here we see that under all conditions, while the failure rate may be high, the partial all-reduce completeness still remains high at $\sim$100\%, thus tolerable for training.

This means that while failures rise for latencies lower than 80\,ms RTT, the practical all-reduce quality is unchanged; and beyond 80\,ms RTT, any reduction in failures comes at the cost of degraded TPS. The required bandwidth and latency are thus chosen at 200\,Mbit/s and 80\,ms RTT latency given the stability check, further balancing contributor participation, all-reduce time, and system throughput.

\subsubsection{All-Reduce Fault Tolerance}

\paragraph{Failure injection design.} We consider the following model of all-reduce: contributors send parameter chunks to reducers (in our case, Pluralis workers), and reducers average each chunk across all contributors before sending the reduced chunk back. This makes the required network traversals clear: the initial sending of parameter chunks by contributors, and the return of the reduced chunk from the reducer. We target these as failure modes by triggering node preemption during these traversals to monitor the robustness and fault tolerance of the system. Conveniently, the DHT allows a trivial implementation of a preemption signal which can be triggered remotely by an operator attaching a temporary node to the system's DHT and writing the required signal. 

\begin{figure}[!t]
    \centering
    \includegraphics[width=\linewidth]{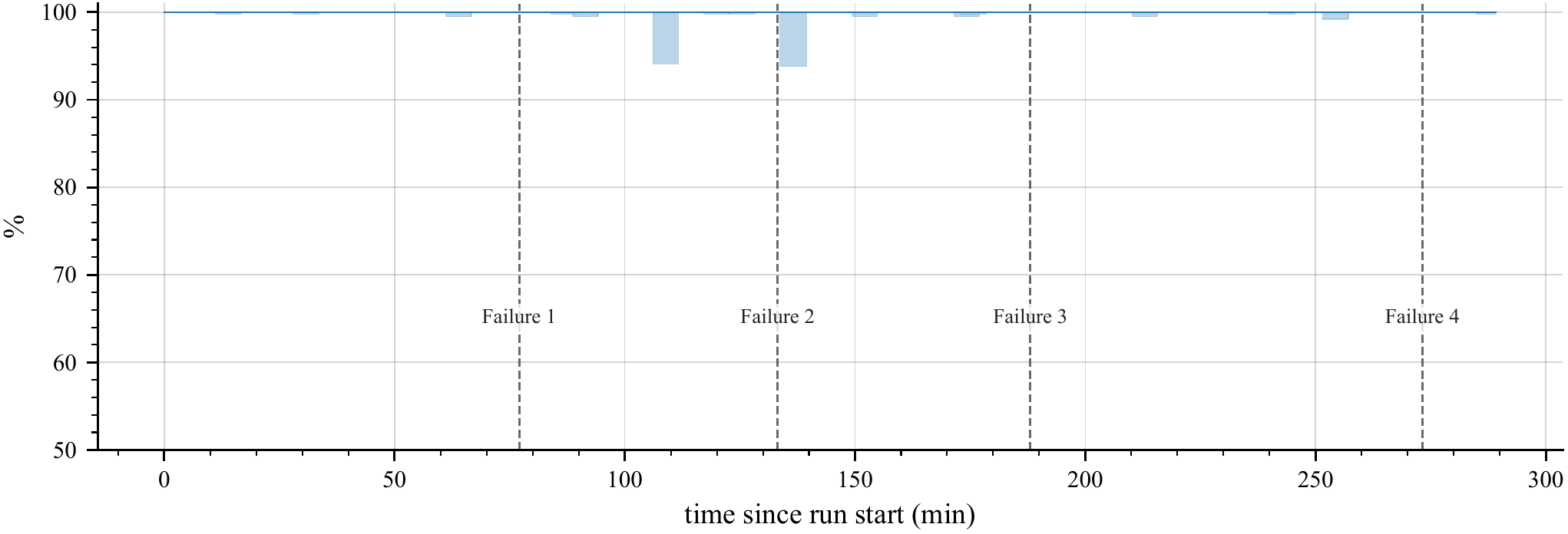}
    \caption{\textbf{All-reduce stability under contributor failure injection.} Contributor failure injection via preemption during all-reduce shows no major effect on system all-reduce stability outside of transient drops in completeness. Overall $>$90\% of the required all-reduce payload is successfully communicated across rounds.}
    \label{fig:ar_average_pcnt_volunteer_fail}
\end{figure}

\paragraph{Contributor fault tolerance.} To test the common case of contributor compute leaving or failing, the above failure injection methodology is applied to a 5-stage instantiation of Agora with 16 workers and 5 simulated contributors with varying network conditions (100--200\,ms RTT latency with 150\,Mbit/s bandwidth) and an all-reduce payload equivalent to that of Pluralis-8B. 

Here, we focus on contributor preemption while sending parameter chunks with the following failure cases:
\begin{itemize}
    \item \textbf{Failure 1:} Preempt 1 contributor from body stage 2 during parameter chunk send
    \item \textbf{Failure 2:} Preempt 3 contributors from body stage 2 during parameter chunk send
    \item \textbf{Failure 3:} Preempt 5 contributors from body stage 2 during parameter chunk send
    \item \textbf{Failure 4:} Preempt 3 contributors from body stages 2 and 3 during parameter chunk send
\end{itemize}
Indeed, as shown in \Cref{fig:ar_average_pcnt_volunteer_fail}, the above failure cases have no noticeable effect on all-reduce completeness when observed from the remaining peers. This is expected, and demonstrates the system works as intended, able to tolerate faults by contributors and, importantly, to isolate the failure. That is, the contributor preemption does not negatively affect the stability or completeness of the all-reduce for any other peers within the same stage, or in different stages, and the subsequent all-reduce rounds are equally stable. 

With $>$90\% successful all-reduce communication in spite of peer failure, strong stability across consecutive rounds, and corroborating convergence results in \Cref{sec:ar-fail-convergence}, we determine a conservative kicking mechanism: a peer is kicked after two AR failures within any three-round window.

\begin{figure}[!t]
    \centering
    \includegraphics[width=\linewidth]{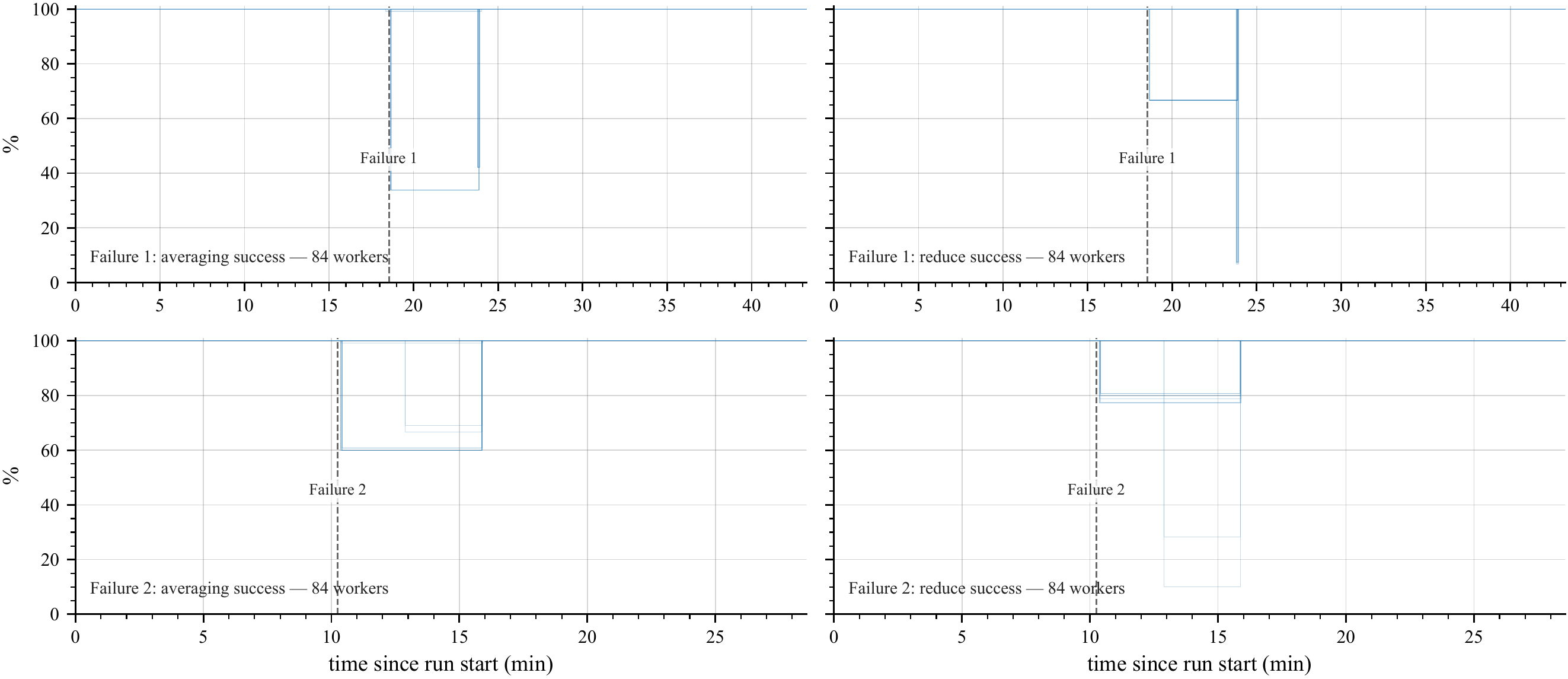}
    \caption{\textbf{All-reduce stability under worker failure injection.} Top: all-reduce completeness and percentage of received parameters by reducer under failure 1. Bottom: all-reduce completeness and percentage of received parameters by reducer under failure 2. Both show only transient instability after failure injection, with perfect subsequent all-reduce rounds.}
    \label{fig:ar_average_reducer_pcnt_worker_fail}
\end{figure}

\paragraph{Worker fault tolerance.} Another important capability of Agora is the ability to run on preemptive compute, rather than only on-demand instances. However, if Pluralis workers are also run on preemptive compute, the system must be able to tolerate failures in nodes that are also assigned the role of reducers. To test these capabilities, we consider a 7-stage instantiation of Agora with 12 workers with an all-reduce payload equivalent to that of Pluralis-8B. 

We focus on the following failure cases, noting that reducers also send back the reduced chunk to contributors:
\begin{itemize}
    \item \textbf{Failure 1:} Preempt 4 workers from body stage 3 during parameter chunk send
    \item \textbf{Failure 2:} Preempt 4 workers from body stage 3 during reduced chunk send
\end{itemize}
As shown in \Cref{fig:ar_average_reducer_pcnt_worker_fail}, failure injection for workers in both cases causes only transient instability of the all-reduce, leading to partial all-reduces but returning to stability in subsequent rounds, demonstrating the system's fault tolerance.

\subsection{System Components}\label{ssec:system-components}

\subsubsection{Authorizer}
\label{sssec:authorizer}

Decentralized training is based on a different trust model than a conventional data center. The machines that perform the computation are not owned or supervised by the operator. They are independent peers that discover and communicate with each other directly over a peer-to-peer network. In an open run, an unknown peer may be underpowered, unreliable, or adversarial. However, training requires that peers behave correctly, maintain the roles assigned to them, and remain mutually reachable. Two requirements are therefore in conflict: the network must remain leaderless during training, but participation cannot be unconditional.

The Authorizer reconciles these requirements. It is a single admission and trust authority with a twofold responsibility: to decide which nodes may participate, and to enable admitted nodes to authenticate one another thereafter without further central involvement. Its functions serve three goals for the run: ensuring that every admitted node is capable of sustaining the role it is given, protecting the run against malicious participants, and keeping the size and per-stage composition of the participant pool compatible with training convergence.

\paragraph{Architecture.}
The Authorizer runs as a single service exposing a small HTTP interface, implemented with the FastAPI framework. Nodes interact with it only when they join the run initially, when polling the queue for admission status, and when refreshing credentials during the run. The service keeps little state of its own and relies on two stores with distinct roles: PostgreSQL for persistent participant records, and Redis for the fast-changing state that coordinates admission.

PostgreSQL holds the persistent data: the record of contributor accounts and the nodes associated with each, together with each node's description, role, its assigned pipeline stage, its recent activity, etc. This is the durable source of truth about who belongs to the run, and it survives restarts of the service.

\begin{figure}[ht]
\centering
  \includegraphics[width=\textwidth]{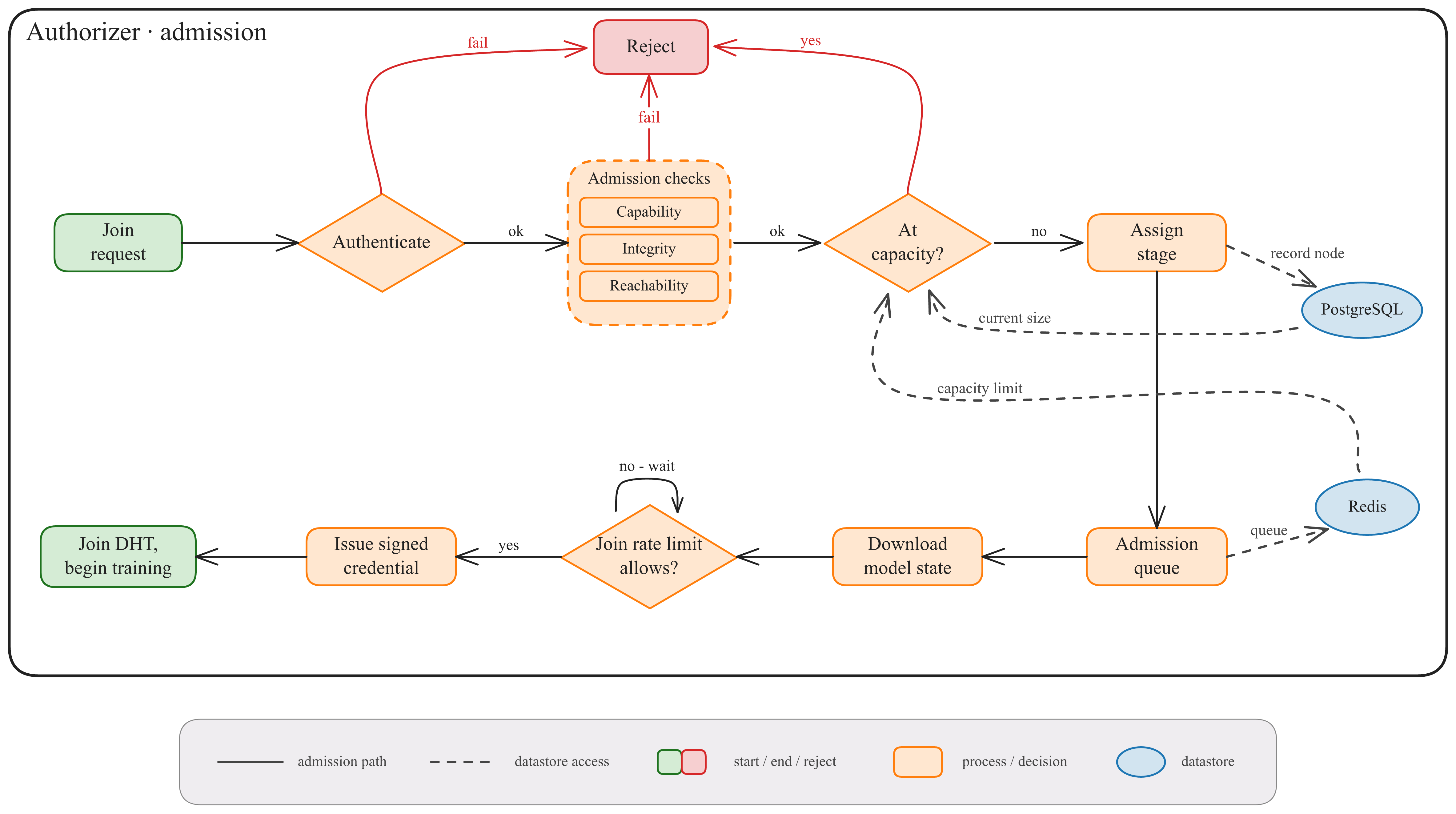}
  \caption{\textbf{Admission flow.} A node's path from a join request to participation in the training network.}
  \label{fig:authorizer-admission}
\end{figure}

Redis holds the dynamic state that changes throughout the run and is read on nearly every request. It stores the admission queue and the operator-tunable parameters that can be changed live without restarting the service. These include the hardware requirements a joining node must meet, the join rate limit, and the capacity limits on the run. Keeping them in Redis lets an operator adjust the run's admission policy while it is in progress. \Cref{fig:authorizer-admission} traces the path that a node follows through admission.

\paragraph{Node admission.}
Any host running the client software may request to join the run, but whether a given node should be admitted depends on the run's specific requirements: some nodes would degrade or endanger the run, so incoming nodes must be filtered at the point of entry. Screening on entry, rather than detecting and removing nodes after they join, confines disruption to outside the training network. The Authorizer performs this filtering through a configurable set of checks, each enabled independently, so that a single system can enforce strict rules for a large public run and relaxed rules for a small internal experiment.

These checks address four distinct failure modes. The first is \emph{capability}: a slow node can drag down the performance of its entire stage, so the Authorizer verifies that a node can do the work assigned to it. It can enforce minimum RAM and VRAM so that a node can hold its assigned shard of the model and process batches, and set bandwidth and latency thresholds so that a node can keep up with its peers over the network; where a run requires a more homogeneous fleet, participation can also be
restricted to an operator-specified set of accelerators. The second is \emph{integrity}: because participants are untrusted, the Authorizer verifies that a joining node runs a legitimate, unmodified build of the client rather than altered code, protecting the run against participants that might otherwise corrupt gradients or poison the shared model state. The third is \emph{reachability}: a node that others cannot reach contributes nothing to a network built on direct exchange, so the Authorizer can confirm that a joining node is able to communicate with the peers it needs to work with. The fourth is \emph{accountability}: every node authenticates before admission, so a participant that misbehaves can be identified and barred, whether by excluding an individual node or an entire identity.

\paragraph{Admission queueing.}
The Authorizer admits contributors gradually rather than all at once. The reason is the cost that each new participant imposes on the rest of the network. A joining node must be inserted into the distributed hash table through which peers discover each other, and every such insertion causes that structure to reorganize. A large number of nodes joining at once would force continual reorganization and could destabilize a run that is otherwise progressing. The Authorizer therefore limits how many contributors may join within a given interval, holding the remainder in an admission queue and releasing them gradually, so that the network grows smoothly rather than in disruptive bursts.

The waiting period is also used productively. A contributor cannot begin training the moment it is admitted, because it must first obtain the current model state, which for a large model is a substantial transfer. The Authorizer overlaps this download with time spent in the queue, so that a contributor retrieves the model state while waiting and is released only once it is ready to begin work.

\paragraph{Stage placement.}
A pipeline-parallel model is divided into ordered stages, and the throughput of the run is limited by its least-provisioned stage. If contributors were free to choose their own stage, frequently chosen stages would accumulate redundant replicas while others remained under-served and became the bottleneck for the entire pipeline. The Authorizer prevents this by retaining ownership of stage assignment. It directs each incoming node to a stage that requires additional capacity, keeping the stages balanced. The stages are not necessarily equal in what they demand of a node: the tail of the pipeline, for example, carries the final projection layer and is therefore more demanding than an interior stage. Where stages differ in this way, the Authorizer can place more capable contributors on the heavier stages, matching each node to a stage it is able to serve.

\paragraph{Peer authentication.}
Controlling admission is not sufficient to protect the run. Because the training network is peer-to-peer, a malicious actor can attempt to connect to honest nodes directly, bypassing the Authorizer. Honest peers must therefore be able to reject an unauthorized party without assistance, and they must do so without consulting the Authorizer on each interaction, since routing every peer-to-peer exchange through a central server would reintroduce the bottleneck and single point of failure that a decentralized design exists to avoid. The mechanism that resolves this is what allows a single root of trust to coexist with a leaderless training loop.

When the Authorizer admits a node, it issues a cryptographically signed credential. The credential binds the node's identity to its public key and an expiry time, and the Authorizer signs it with its private key; the corresponding public key is distributed to every participant. The credential is in this way a portable and unforgeable assertion that a given identity, holding a given key, was admitted to the run.

\begin{figure}[ht]
  \centering
  \includegraphics[width=\textwidth]{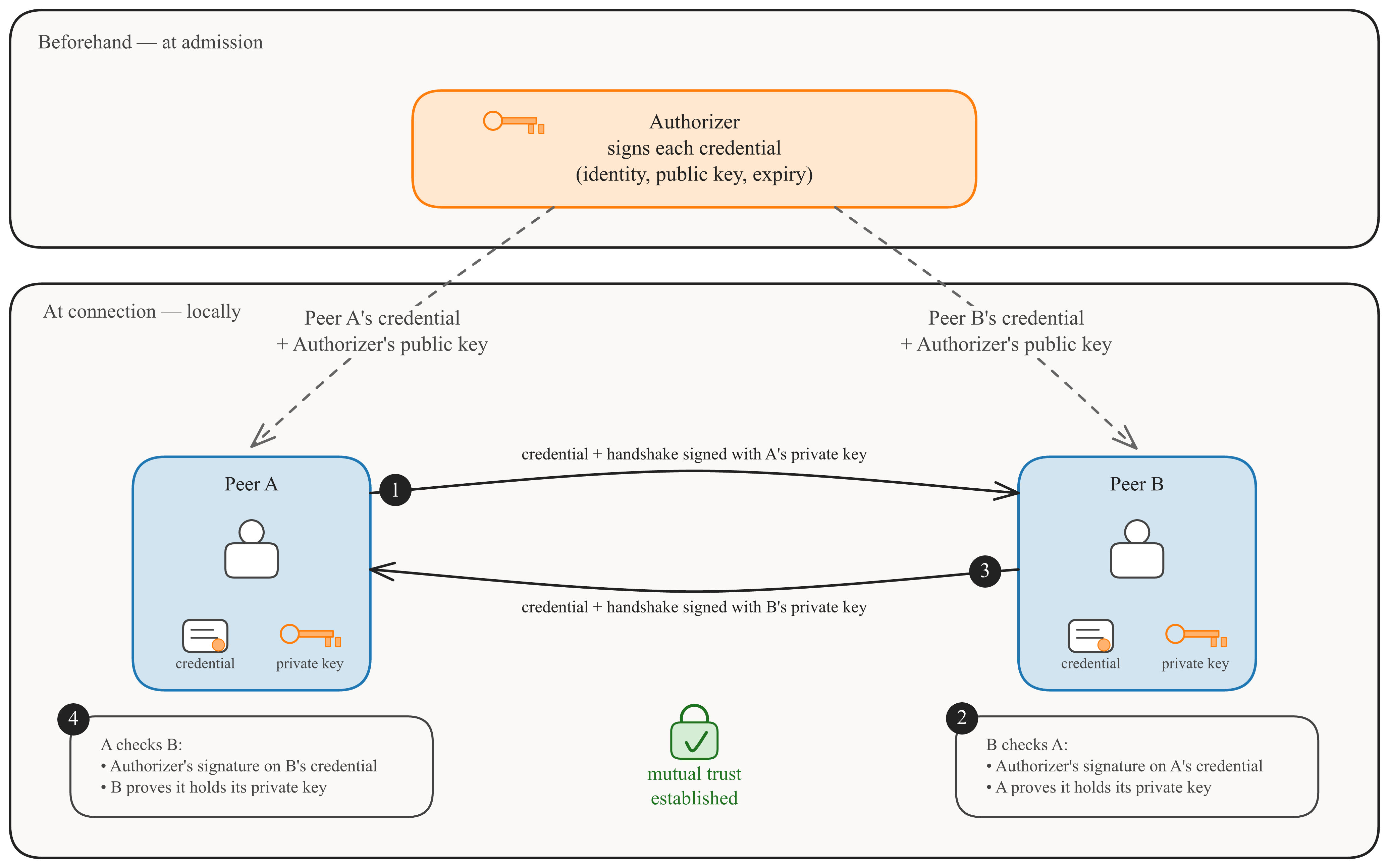}
  \caption{\textbf{Peer authentication.} The Authorizer signs each credential once, and its public key is shared with every peer in advance. Thereafter peers authenticate one another locally.}
  \label{fig:authorizer-handshake}
\end{figure}

Authentication thereafter occurs entirely between peers. When two nodes establish a connection, each presents its credential, and the receiver performs two checks locally. It first verifies the Authorizer's signature on the credential, which establishes that the Authorizer vouched for that identity and its public key. Then, it checks that the connecting party actually holds the private key that corresponds to the public key named in the credential. The party proves this by signing its handshake messages with that private key, and the receiver verifies the signature against the public key it has just authenticated. Because only the legitimate holder could produce a valid signature, a party that merely replays a credential captured from someone else cannot pass this step. The exchange is mutual and incorporates freshness guarantees that prevent the replay of previously observed messages. \Cref{fig:authorizer-handshake} summarizes the exchange.

This design keeps the Authorizer off the training data path: verification is local, so no peer-to-peer exchange waits on it. Credentials are short-lived and must be renewed periodically, which gives the operator a direct means of revocation, since access to a node that has proven malicious, or is otherwise no longer wanted, can be withdrawn simply by declining to renew its credential. An attacker who never obtained a credential, in turn, is rejected by the honest peers themselves, with no central enforcement required.

\subsubsection{Health Monitor}
\label{sssec:health-monitor}

The same property that lets the training network run without a central coordinator also limits what the operator can see. Once the Authorizer admits a node and that node joins the peer-to-peer network, it runs autonomously on hardware the operator neither owns nor can control. There is no privileged channel through which the operator can connect to it, query its state, or confirm that it is doing the work it was admitted to do. The run can still be observed, but only indirectly, through two channels that the nodes themselves produce: the metrics that each node reports to the monitoring pipeline (\Cref{sssec:metrics}), and the DHT through which peers discover and coordinate with each other.

The Health Monitor turns the second of these channels into a view of the run from the inside. It is a utility node that does not perform training; it joins the DHT through the ordinary admission path and reads the coordination state that peers publish in the normal course of operation. Because it observes what the peers themselves observe, it can report which nodes are actually present and contributing, independent of the record of who was admitted. This makes it the operator's primary means of knowing who is still inside the run and who has left.

\begin{figure}[ht]
  \centering
  \includegraphics[width=\textwidth]{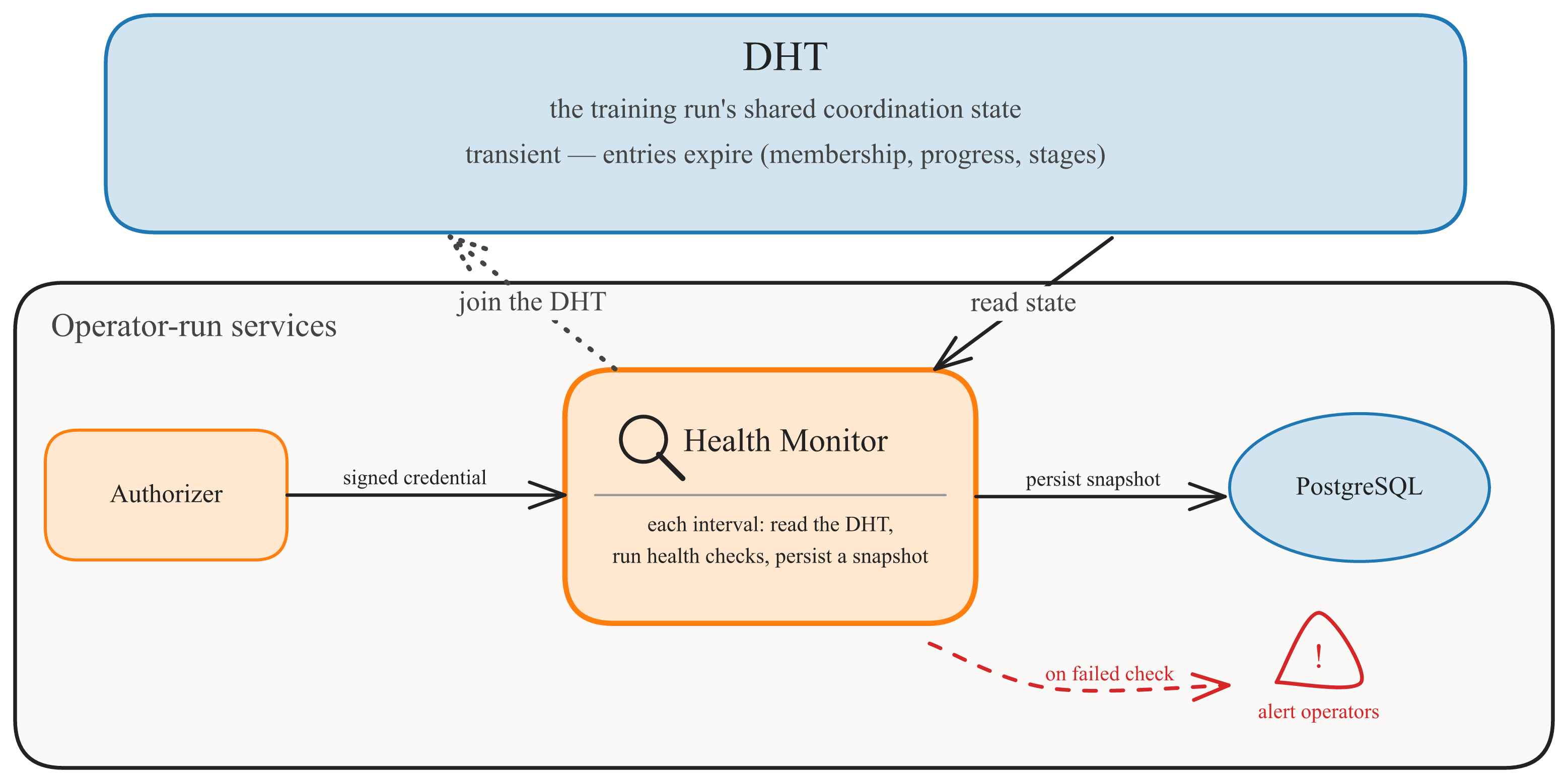}
  \caption{\textbf{Health Monitor architecture.} Each interval, the operator-run Health Monitor reads the run's live state from the DHT, runs health checks, and persists a snapshot to PostgreSQL, raising an alert to the operators when a check fails.}
  \label{fig:health-monitor}
\end{figure}

\paragraph{Architecture.}
The Health Monitor runs as a standalone daemon, separate from the training processes. On a fixed interval, it reads the current state of the DHT, combines it with the reported metrics, and writes an aggregated snapshot to PostgreSQL. The components that consume this information, including the public Dashboard, read from PostgreSQL rather than querying the DHT directly. This gives the run a persistent store of its state, decoupled from the training network, that consumers can query without touching the live DHT. \Cref{fig:health-monitor} shows how the monitor joins the run and where its observations are stored.

\paragraph{Membership tracking.}
The DHT holds only the present-tense state. Each entry a node publishes carries an expiration time; a live node periodically refreshes its entries, and when it stops doing so, whether because it left cleanly or simply failed, those entries lapse and disappear. This self-cleaning behavior is what makes the DHT a reliable signal of current membership: an entry that is still present corresponds to a node that is still refreshing it, and so was alive within the most recent refresh window. The same behavior, however, means the DHT retains no history. A node that has left leaves no trace once its entries expire.

The Health Monitor supplies the history that the DHT does not keep. By snapshotting the network at each interval and persisting what it finds, it builds a durable record of which nodes were present at each point in the run and when each was last seen active. This record is the only place the operator can reconstruct who was inside the swarm over time, distinguish a node that is still working from one that was admitted but has since dropped out, and recover that picture after the live state has expired.

\paragraph{Training progress.}
Trainer nodes periodically report their progress to the DHT, including the current loss and the number of batches each has processed. Because every DHT record carries an expiration time, these reports are transient. On each cycle, the Health Monitor reads the current reports, discards any it has already counted, aggregates the rest across the reporting trainers, and writes the consolidated result to the database. The run thus retains a clean, persistent record of its loss and processed batches that the expiring DHT entries alone would not preserve.

\paragraph{Health checks.}
Because no party directly supervises the running network, the Health Monitor continuously checks that what the network reports matches what the run requires. It reconciles the stage announcements in the DHT against the admission records, confirming that each node is serving the stage it was allocated and that no node is present in the network without having been authorized. It checks that every stage retains enough serving replicas to keep the pipeline from stalling at an under-provisioned stage. Since the metrics live in an open DHT that any admitted peer could write to, it accepts training reports only from the trainer nodes entitled to produce them, verified by signature; a node forging metrics it has no authority to report is flagged and its identity recorded for exclusion, in the same way the Authorizer bars an untrustworthy identity. When any of these checks fails, the Health Monitor can raise an alert to the operators, so that a developing fault is surfaced promptly rather than discovered only after it has affected the run.

\subsubsection{Metrics Monitor}
\label{sssec:metrics}

As established for the Health Monitor (\Cref{sssec:health-monitor}), the run is observable only through the two channels the nodes themselves produce: the DHT they coordinate over, and the metrics they report about their own execution. The Metrics Monitor is concerned with the second, the dense quantitative telemetry each node produces: GPU and RAM utilization, processed batches, utilized bandwidth, gradient statistics, etc. Reporting the full volume of these metrics through the DHT is infeasible. The DHT exists to carry coordination state that the training depends on, and it is built for sparse, slowly changing entries. The metrics are the opposite: dense, high-frequency, and per-node. Pushing them all through the DHT would overwhelm it and degrade the coordination it is responsible for. The run therefore collects metrics through a separate pipeline based on Prometheus~\cite{turnbull2018prometheus}, a system designed for exactly this kind of monitoring: ingesting numerical time series at scale, storing them, and querying and alerting on them.

\begin{figure}[ht]
  \centering
  \includegraphics[width=\textwidth]{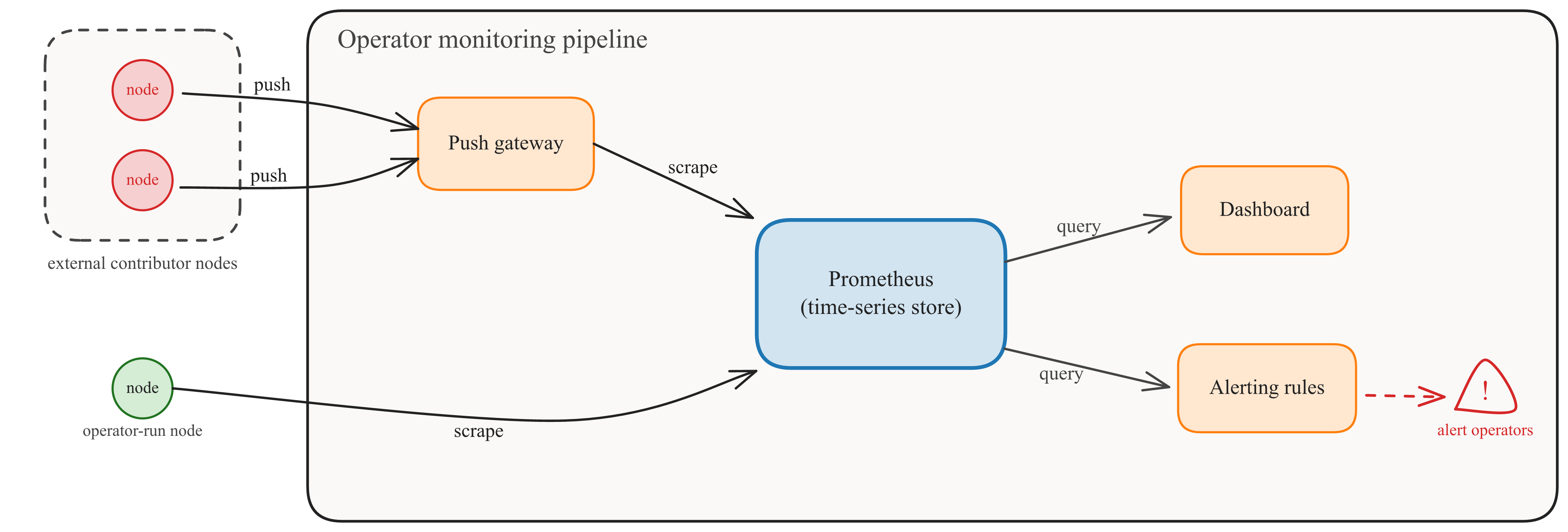}
  \caption{\textbf{Metrics pipeline.} External contributor nodes push their metrics outward to an operator-run gateway, which Prometheus scrapes; operator-run nodes are scraped directly and bypass the gateway. The collected time series are stored in Prometheus and consumed by the public Dashboard and the alerting rules.}
  \label{fig:metrics-pipeline}
\end{figure}

\paragraph{Architecture.}
Prometheus follows a pull model: the monitoring server periodically connects to each target and collects the metrics it exposes at a dedicated endpoint. On the operator's own machines this is straightforward. For an external contributor it is more of an imposition. A contributor already opens a network port for the peer-to-peer training connection; requiring it to open and expose a second endpoint would add another step to the setup and place an extra inbound service on a machine the operator does not run. The pipeline resolves this with a push gateway. Rather than exposing an endpoint to be scraped, a contributor node pushes its metrics outward, on a fixed interval and over an authenticated connection, to a gateway that the operator runs; Prometheus then scrapes the gateway rather than the node. Nodes that the operator runs are scraped directly and bypass the gateway, so the indirection is used only where it is needed. Each node labels its metrics with a per-instance identity recording its role, its pipeline stage, and the instance on which it runs, so that every series is attributable to a specific node and stage and stays distinguishable across restarts. \Cref{fig:metrics-pipeline} shows the resulting paths.

\paragraph{Alerting.}
Collecting metrics centrally also makes it possible to act on them automatically. The pipeline evaluates a set of rules over the incoming series and raises an alert whenever a value crosses a threshold that signals a developing fault, for example, a stage losing peer visibility, training progress stalling, GPU or host memory under pressure, degraded all-reduce or group formation, anomalous network traffic, or ML-signal warnings such as a diverging loss or exploding gradients. Alerts are dispatched to the operators, so that conditions affecting the run are surfaced promptly.

\subsubsection{Dashboard}
\label{sssec:dashboard}

Agora supports collaborative training runs, and such a run is conducted in the open: it publishes its state as it happens, for anyone to follow. The Dashboard is the public window into that state, and it serves two audiences. Contributors want to confirm that the machines they have committed are admitted and working, and to see how those machines are performing. The general audience, following the run, wants to see how it is progressing as a whole. To serve both, the Dashboard presents the training loss and throughput over time; the model's accuracy on standard public benchmarks as training proceeds; the number, status, and geographic distribution of the participating nodes; and live telemetry for any individual node. Contributors can also see their own contribution score and how they rank against other participants.

\paragraph{Architecture.}
The Dashboard pairs a web frontend, built with Next.js, with a backend service built on FastAPI. The backend reads, aggregates, and serves the information about the run that is collected by the other components of the system. Membership, training progress, and per-contributor records come from the PostgreSQL database the Health Monitor maintains (\Cref{sssec:health-monitor}); the per-node telemetry comes from the Prometheus store the Metrics Monitor fills (\Cref{sssec:metrics}). Its responses are cached and served from a content-delivery edge, so that a large public audience places little additional load on the live training infrastructure or on the metrics store it reads from. The Dashboard is also not tied to a single run: it can present more than one at a time, so a completed run remains permanently viewable and can be compared side by side with one still in progress.

\subsection{Orchestrator and Infrastructure, and Testing}

The main strength of the Agora system lies in providing a decentralized Protocol Learning framework that is agnostic to the underlying hardware resources that run it. This enables invariance in where the hardware physically sits, which cloud provider hosts it, which GPU families are used, and how the resources connect to the internet. Taking advantage of this strength, the internal testing of Agora was performed in a cost-effective and flexible manner using preemptive heterogeneous compute that was allocated and managed programmatically. The result is a resource management layer that can use multiple cloud providers with an API to orchestrate the Agora protocol on top of it. For this, we use an internal service for Agora's infrastructure orchestration, handling deploying, operating, and testing distributed training runs. It manages the full lifecycle of a run: provisioning cloud resources, bootstrapping hosts, launching containerized workloads, wiring service discovery, and capturing run metadata and metrics.

The system is built around infrastructure as code using Pulumi. Pulumi provides a declarative resource model, while the service layers imperative orchestration on top of it. This allows for resources to be requested declaratively, while their sequencing, role assignment, and runtime configuration are programmatically controlled. For example, we may need to add workers to specific pipeline stages, assign them unique DP ranks, or add new trainers to an existing run due to shard exhaustion (see \Cref{ssec:results_setup}).

All workloads are containerized and launched across provisioned nodes, supporting all the Agora roles, including seed nodes, workers, trainers, Prometheus, authorizers, health monitors, dashboards, storage, and databases. Its flexible config-driven model allows for lean development runs, staging deployments, production topologies, and targeted experiments.

This service is also used as an experimentation harness, supporting ephemeral environments for short-lived testing, staging environments that mirror production, and production deployments for live training runs. It includes controls for CPU-based workers and private networking to reduce cost for system tests, spot instance support, checkpointing, and rollbacks or resuming. It can also emulate network conditions through traffic shaping, controlling bandwidth and latency on nodes, as well as simulate worker churn, having workers leave and join training (see \Cref{testing_framework}).

Each deployment can be viewed as a resource DAG that has a sequence composed of stacks. A stack is a group of resources that are created, updated, and destroyed together. Stacks expose the runtime information required by later stages of the deployment. This allows for deployments to be brought up incrementally.

\begin{figure}[h]
    \centering
    \includegraphics[width=0.9\linewidth]{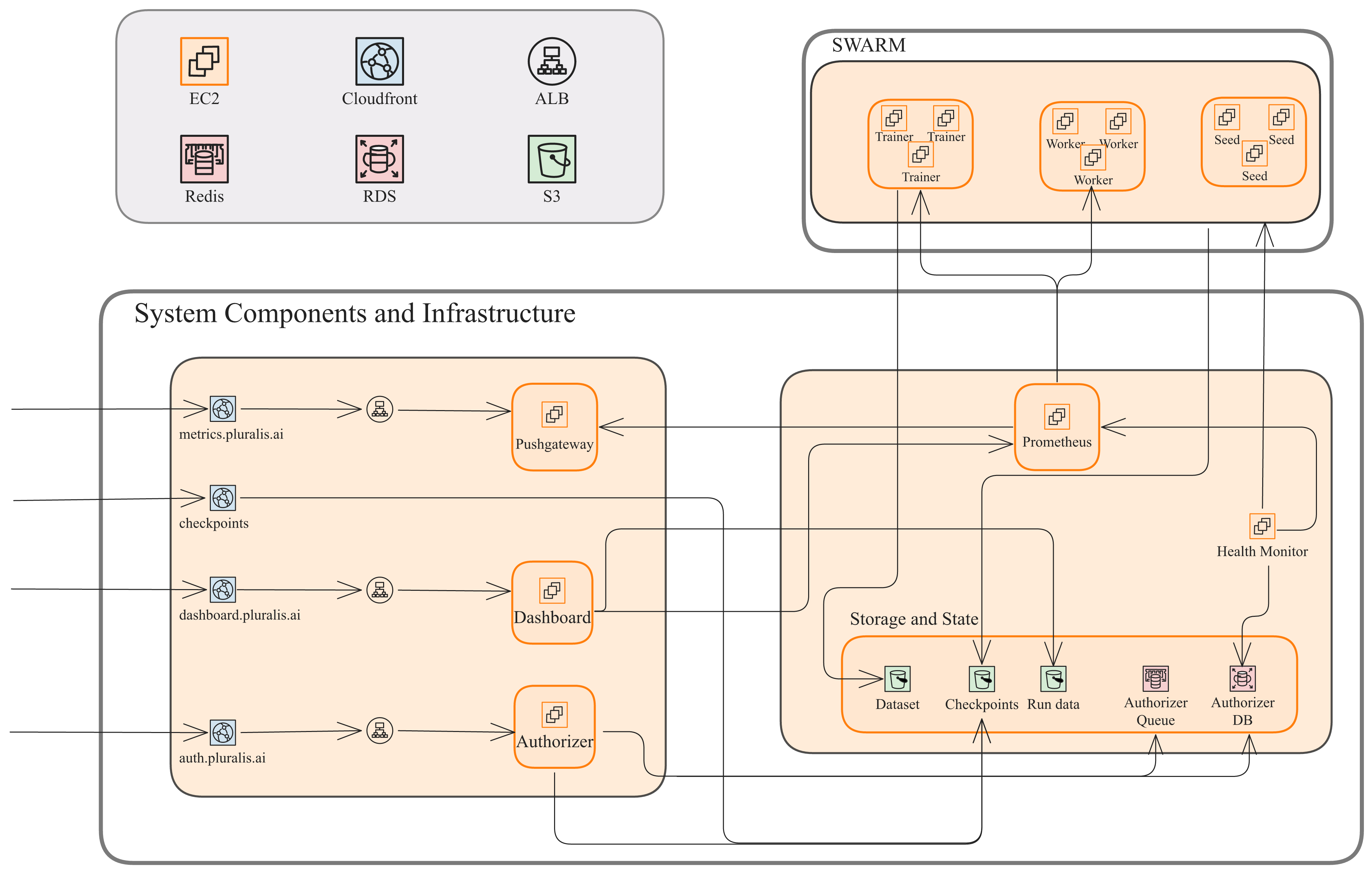}
    \caption{\textbf{Cloud architecture of the Agora system.}
    }
    \label{fig:cloud_arch}
\end{figure}

The cloud architecture is shown in \Cref{fig:cloud_arch}. The deployment flow is split up into two main phases: core and pipeline deployments. The core provisions the infrastructure for a swarm to exist, records the outputs, and exposes metadata for service discovery by later stages, such as seed multiaddresses, Prometheus hosts, database endpoints, cache endpoints, authorization servers, and checkpoint locations. The pipeline phase then consumes these outputs to attach workers and trainers to an existing swarm.

\begin{enumerate}
    \item \textbf{State Stack}: Persistent infrastructure and stateful resources, including databases and caches. These resources back services such as authorization and health monitoring.
    \item \textbf{Observability Stack}: Metrics infrastructure, Prometheus and alerting components.
    \item \textbf{Core Stack}: Core platform services required to bootstrap and operate the swarm, including seed nodes, authorizers, health monitors, push gateways, and networking attachments such as target groups and load balancers for exposed services.
    \item \textbf{Worker and Trainer Stacks}: Workers and trainers doing work within the system. These are horizontally scalable and are typically deployed many times.
\end{enumerate}

\section{Pluralis-8B}
\label{sec:pluralis-8b}
To demonstrate the capabilities of Agora under a realistic pretraining workload, we chose to train an 8B-parameter model, which we call Pluralis-8B. The objective of the run was to demonstrate that the underlying system could sustain a realistic pretraining workload at a meaningful scale. A system of this kind cannot be thoroughly validated through isolated benchmarks or small-scale experiments. Instead, it must be exercised end-to-end under the communication, coordination, and reliability constraints of an actual distributed run. For this purpose, we chose an 8B-parameter model~(see~\Cref{sec:results}) as it is large enough to constitute a realistic modern pretraining workload and to expose system behavior that is difficult to reproduce in small-scale experiments. Models around the 8B-parameter range have been used in several open, full-scale pretraining efforts, including OLMo and Llama~\citep{groeneveld2024olmo,olmo2025olmo,grattafiori2024llama}. At this scale, the model is sufficiently deep and computationally demanding to be partitioned across many pipeline stages, allowing us to thoroughly exercise the distributed system.

This run also introduces significant machine learning changes that make better use of the available compute; we discuss these in detail below. These changes are dictated by the fact that along both the pipeline-parallel and data-parallel axes, workers must communicate with one another over the internet, where bandwidth is far too scarce to exchange uncompressed signals. Each is, at its core, a measure to reduce this communication overhead.

\subsection{Pipeline-Parallel Communication with Reparameterized SSNs}
\label{ssec:reparam_ssn}

Communicating activations and gradients across pipeline stages is expensive. An 8.6B-parameter transformer with embedding dimension $d = 5120$ and a 4096-token sequence produces a $4096 \times 5120$ activation at every stage boundary, about 671\,Mbit per microbatch in FP32. On a 200\,Mbit/s link, the transfer takes several seconds, so the pipeline becomes communication-bound long before it is compute-bound. Subspace networks~(SSNs)~\citep{ramasinghe2025ssn} reduce this cost by constraining the inter-stage signal to a single low-rank subspace shared across all stages, with each stage sending only its low-dimensional coordinates.

This construction relies on the residual structure of a transformer. Let $\mX^{l} \in \R^{n \times d}$ be the hidden states after layer $l$, with one row per token. Each layer adds the outputs of its attention and MLP projections to the residual stream,
\begin{equation}\label{eq:residual_stream}
    \mX^{l} = \mX^{0} + \sum_{i=1}^{l}\big(\mX^{i}_{\mathrm{concat}}\,\mW_{p_1}^{i} + \mX^{i}_{\mathrm{hidden}}\,\mW_{p_2}^{i}\big),
\end{equation}
where $\mW_{p_1}^{i}$ and $\mW_{p_2}^{i}$ are the attention and MLP output projections. If the rows of every such projection lie in a common $k$-dimensional subspace with orthonormal basis $\mU_k \in \R^{d \times k}$, each layer contribution also lies in that subspace. A stage can then subtract the positional embeddings $\mathrm{PE}$ and the fixed token embedding $\mT_{\mathrm{fixed}}$, available to both sender and receiver, and send only the projected coordinates
\begin{equation}\label{eq:ssn_compress}
    \big(\mX^{l} - \mathrm{PE} - \mT_{\mathrm{fixed}}\big)\,\mU_k \;\in\; \R^{n \times k},
\end{equation}
which the receiver reconstructs with $\mU_k^{\top}$. This transmits $k \ll d$ values per token rather than $d$, and the backward pass uses the same compression.

The original SSN enforces this subspace constraint during training by storing each projection as a full $d$-column matrix and projecting it back into the subspace after every optimizer step. Components outside the subspace are therefore removed before they can affect the communicated activations. This observation suggests that the same constraint can be represented more directly, avoiding the additional memory and computation associated with maintaining the projection matrices full-rank. We therefore encode the constraint in the model architecture through parameterization. In particular, a matrix has its rows in the subspace exactly when it factors as $\mZ\mU_k^{\top}$, so we reparameterize each projection through its low-dimensional factor and write
\begin{equation}\label{eq:reparam}
    \mW_{p_1}^{i} = \mZ_{1}^{i}\,\mU_k^{\top}, \qquad \mW_{p_2}^{i} = \mZ_{2}^{i}\,\mU_k^{\top},
\end{equation}
training $\mZ$ directly. The model is still operating under the same function, but each projection now holds $k$ columns instead of $d$ and the separate projection step is removed. The optimizer requires no special handling, since the gradient for $\mZ$ is $(\nabla_{\mW}\Ls)\,\mU_k$ by the chain rule.

The trainable part of the token embedding requires the same treatment. If any component lay outside the subspace, the projection in~\Cref{eq:ssn_compress} would discard that component and the reconstruction would no longer be exact. Thus, we also factor the embedding table as $\mT_{\mathcal{S}} = \mE\,\mU_k^{\top}$. With every projection and the embedding parameterized through $\mU_k$, the residual stream remains in the subspace by construction. The inter-stage signal can then be reconstructed in an entirely lossless manner, and subspace membership no longer requires explicit enforcement. This way, the reparameterized SSN makes the subspace constraint explicit in the model architecture, reducing the total number of trainable parameters by allocating parameters only within the transmitted subspace. As we see in our experiments in~\Cref{sec:ablations}, this reduction does not adversely affect convergence and is accompanied by improved gradient stability during training. Because the factorization is built directly into the weight matrices, the constraint is enforced naturally throughout the forward and backward passes without requiring a separate projection or optimizer-side operation.

\subsection{Data-Parallel Communication with AsyncSPARTA}
\label{ssec:async_sparta}
Reparameterized SSNs compress the activations and activation gradients that cross stage boundaries, which unblocks the pipeline-parallel axis of Pluralis-8B. The data-parallel axis raises a distinct problem where the workers holding replicas of one stage must be kept from drifting apart, and the communication this requires is again far too heavy for an internet-grade link to carry uncompressed. 

The data-parallel communication challenge is twofold: \emph{volume} and \emph{frequency}. A stage holds at least one full Transformer block, whose parameters are dominated by its attention and MLP projections; for a Llama-style block these amount to roughly $12d^{2}$ values, about 310M parameters at $d = 5120$ or approximately 10\,Gbit in FP32. Averaging a single block across replicas therefore can take up to roughly a minute of transfer on a 200\,Mbit/s home link. This volume becomes even more significant when a stage spans multiple blocks. Apart from the volume, the frequency of synchronization is also a problem. Ordinary data parallelism averages once per step, so even a perfectly compressed exchange that must still complete on every step, synchronously, can throttle throughput. Therefore, either quantity (the volume of a single synchronization or how often one must occur) can become the bottleneck.

In our previous run, Node0~\citep{avraham2025node0}, we addressed the volume with PowerSGD~\citep{vogels2019powersgd}, a low-rank gradient compressor that exploits the rapidly decaying spectrum of stochastic gradients. In particular, it approximates each layer's gradient $\mG = \nabla_{\mW}\Ls \in \R^{a \times b}$ by a rank-$r$ product $\mP\mQ^{\top}$, with $\mP \in \R^{a \times r}$, $\mQ \in \R^{b \times r}$, and $r \ll \min(a, b)$. Warm-starting from the previous step's right factor $\mQ$, a single step of subspace iteration suffices to refine the approximation,
\begin{equation}\label{eq:powersgd}
    \mP \;\leftarrow\; \mathrm{Orth}(\mG\mQ), \qquad \mQ \;\leftarrow\; \mG^{\top}\mP,
\end{equation}
after which the replicas exchange only the thin factors $\mP$ and $\mQ$ and reconstruct $\hat{\mG} = \mP\mQ^{\top}$. Because every operation in~\Cref{eq:powersgd} is a matrix product, the compressor is linear and commutes with summation over replicas: $\sum_{j}\mG_{j}\mQ = \big(\sum_{j}\mG_{j}\big)\mQ$. Hence, the factors can be aggregated with a bandwidth-efficient all-reduce rather than a gather~\citep{vogels2019powersgd}.

The low-rank compression used in PowerSGD is biased, meaning that the discarded directions are simply absent from what each replica sends. As such, PowerSGD is usually paired with error feedback~\citep{seide2014onebit,karimireddy2019error}. Each replica keeps a local memory $\mathbf{E}$ of the signals that were dropped so far and folds it back into the gradient before compressing again:
\begin{equation}\label{eq:errorfeedback}
    \mathbf{E}_{t+1} \;=\; \big(\mathbf{E}_{t} + \mG_{t}\big) \;-\; \mathcal{C}\big(\mathbf{E}_{t} + \mG_{t}\big).
\end{equation}
Here, $\mathcal{C}(\cdot)$ is the rank-$r$ compressor of~\Cref{eq:powersgd} and the aggregated message $\mathcal{C}(\mathbf{E}_{t} + \mG_{t})$ is what drives the optimizer step. The residual $\mathbf{E}_{t+1}$ carries the systematically dropped directions forward, so they are eventually transmitted in case their subspace becomes prominent over the course of training.

While PowerSGD addressed the volume of communication in Node0, the frequency became the main bottleneck. A gradient is meaningful only for the step that produced it, and instantaneous gradients vary sharply from one step to the next~\citep{beton2025sparta}. As a result, gradient exchange cannot be deferred, amortized across several steps, or run against stale values. It has to finish, synchronously, before each optimizer step. On the high-latency consumer links Pluralis-8B targeted, it is this per-step blocking synchronization, more than the size of any single exchange, that ultimately could have limited the throughput.

Unlike gradients, weights move slowly. Each optimizer step nudges them by only a small, momentum-smoothed increment, so a replica's weights at consecutive steps are nearly identical. Slow-moving weights tolerate precisely what gradients do not: stale, infrequent, non-blocking exchange. Averaging the full weights would reintroduce the volume problem, but \citet{beton2025sparta} showed that with sparse parameter averaging (SPARTA) this is unnecessary. Concretely, at each round every replica shares only a small random fraction $p$ of its weights (for instance $p = 5\%$) and replaces them with the cross-replica average. Despite sharing such small volume per round, SPARTA helps the replicas to remain in consensus. More importantly, because SPARTA averages slow-moving weights rather than gradients, we can build that average from parameters broadcast on a previous step, overlapping communication with computation at no additional wall-clock cost~\citep{beton2025sparta}.

Pluralis-8B therefore adopts an asynchronous variant of SPARTA~(AsyncSPARTA) on the data-parallel axis~\citep{ajanthan2026asyncmesh}. Each of the $R$ replicas of a stage trains independently and applies an optimizer step to its locally accumulated gradients once the stage has collectively reached its target batch size. Then, on a fixed cadence of every $N$ steps, the replicas average a shared random index set $\gI \subseteq \{1, \dots, |\vtheta|\}$ of their parameters, with $|\gI| = p\,|\vtheta|$:
\begin{equation}\label{eq:sparta}
    \vtheta^{(j)}[\gI] \;\leftarrow\; \frac{1}{R}\sum_{j'=1}^{R} \vtheta^{(j')}[\gI], \qquad j = 1, \dots, R,
\end{equation}
where $\vtheta^{(j)}$ gathers all trainable weights of replica $j$ and $\vtheta^{(j)}[\gI]$ denotes the coordinates selected by $\gI$. The right-hand side may be assembled from parameters that were broadcast a step or more earlier, so the exchange in~\Cref{eq:sparta} never blocks the local optimizer from proceeding while the weight synchronization is happening in the background.

AsyncSPARTA sharply cuts our data-parallel traffic. Returning to the per-block estimate, sharing $p = 5\%$ of a block moves about 16M parameters, roughly 0.5\,Gbit per round in place of the full 10\,Gbit. A smaller $p$ lowers the communicated volume even further, traded against the number of rounds needed to keep replicas in consensus. Because the exchange is sparse and asynchronous, it overlaps with computation and is taken off the critical path~(see~\Cref{worker_section,trainer_section}).

To enable AsyncSPARTA in Agora, we had to make the following two changes.

\subsubsection{Pinning the Backward Pass}
\Cref{ssec:training-with-agora} notes that Agora pins each microbatch's backward pass to the worker that ran its forward, rather than load-balancing it across the stage's replicas. As mentioned, Agora's workers are stateless, so a worker discards a microbatch's activations once it has passed them downstream. When the backward arrives, the receiving worker recomputes the stage's forward to rebuild its autograd graph. That recomputation reproduces the original activations only if it runs under the same weights that produced them. Under synchronous data-parallel methods such as PowerSGD, every replica of a stage holds identical weights, so any of them could serve the backward pass even if they were not the same worker that ran the forward. In contrast, AsyncSPARTA requires the same path to be traversed during the forward and backward passes, since between sparse-averaging rounds each replica advances on its own weights and the replicas of a stage are no longer interchangeable.

To formulate the problem, consider a stage that maps its input activation $\mX$ to an output $\mX' = f(\mX; \vtheta)$. By the chain rule, the stage's parameter gradient factorizes through that output,
\begin{equation}\label{eq:pin_chain}
    \nabla_{\vtheta}\Ls = \Big(\frac{\partial \mX'}{\partial \vtheta}\Big)^{\!\top} \nabla_{\mX'}\Ls.
\end{equation}
The forward produced ${\mX' = f(\mX; \vtheta^{(j)})}$ on replica $j$, and the downstream stage returned the activation gradient $\nabla_{\mX'}\Ls$ for that particular $\mX'$. If a drifted replica $j' \neq j$ ran the backward instead, it would rebuild the graph from $f(\mX; \vtheta^{(j')}) \neq \mX'$ and evaluate the Jacobian in~\Cref{eq:pin_chain} at $\vtheta^{(j')}$. This causes the gradient to then differ from the one replica $j$ would have produced.\footnote{We can show that for a smooth stage, the gradient difference is bounded by an amount controlled by the weight drift $\lVert \vtheta^{(j')} - \vtheta^{(j)} \rVert$.} As we will show in our ablation studies, feeding such inconsistent gradients into the optimizer perturbs each replica's local trajectory and slows convergence. As such, we remove this inconsistency by pinning the backward to replica $j$.

\subsubsection{Preserving Local Progress with Delta Rule Averaging}
AsyncSPARTA introduces a second challenge where it can erase a replica's local progress. Because an averaging round runs in the background while the replica keeps training, the local weights advance between the moment the round starts (when each replica contributes its snapshot) and the moment it completes. Writing the returned average back in place would overwrite the local update made in between. As we show in~\Cref{sec:results}, a single round can span as many as five optimizer steps, so the progress might not be negligible. 

To preserve that progress, we apply the average as a correction rather than a replacement. Let $\vtheta^{(j)}_{\star}[\gI]$ be the snapshot replica $j$ contributed at the start of a round and $\bar{\vtheta}[\gI] = \tfrac{1}{R}\sum_{j'=1}^{R} \vtheta^{(j')}_{\star}[\gI]$ the average returned by~\Cref{eq:sparta}. By the time the round finishes, the replica holds the advanced weights $\vtheta^{(j)}[\gI]$. Therefore, the delta rule adds only the consensus shift to them,
\begin{equation}\label{eq:delta_rule}
    \vtheta^{(j)}[\gI] \;\leftarrow\; \vtheta^{(j)}[\gI] + \big(\bar{\vtheta}[\gI] - \vtheta^{(j)}_{\star}[\gI]\big).
\end{equation}
The local update $\vtheta^{(j)}[\gI] - \vtheta^{(j)}_{\star}[\gI]$ is thus retained while the replica is still pulled toward the average. 
This is the special case of the more general correction introduced in AsyncMesh~\citep{ajanthan2026asyncmesh} where we set $\lambda=1$. \citet{maziane2026loscar} analyze the corresponding delay-corrected sparse averaging scheme and establish convergence guarantees for smooth non-convex objectives under local SGD with communication--computation overlap.

\subsection{Stage-Wise Gradient Clipping}
\label{ssec:stage_clipping}
So far, we have discussed how the preceding methods address communication along both axes of Pluralis-8B. Reparameterized SSNs reduce the activations exchanged between pipeline stages, and AsyncSPARTA makes the data-parallel synchronization sparse and non-blocking. Preserving these gains requires that the remaining optimization steps do not reintroduce a synchronous operation on every step. Gradient clipping is such an operation.

In centralized training, gradient clipping is a global operation where the gradient norm is computed over all of the model's parameters before the optimizer step. 
Under pipeline parallelism, gradient clipping requires a cross-stage reduction of the local squared norms, forcing every stage to wait before any worker can update its parameters.
Although this collective is inexpensive on a datacenter fabric, over Agora it would impose a synchronous barrier on every step, coupling stages and undermining their autonomy. 
Pluralis-8B therefore clips gradients independently at each stage, requiring no cross-stage communication (see~\Cref{alg:training}).

Stage-wise clipping requires setting a threshold for each stage. Applying the usual global threshold of 1 independently at all $S$ stages would no longer approximate global clipping: in the worst case, the combined gradient norm could reach $\sqrt{S}$, and a uniform threshold would ignore systematic differences in gradient scale across the pipeline stages. In particular, the tail stage computes the loss and the output projection over the entire vocabulary and consistently exhibits larger gradient magnitude compared to the head and body stages~(see~\Cref{fig:pre_step_grad_per_stage}). Since the squared global norm of the gradient $\mG$ decomposes as
\begin{equation}
    \lVert \mG \rVert_2^2 = \sum_{s=1}^{S} \lVert \mG_s \rVert_2^2,
\end{equation}
an equal division of a unit squared-norm budget gives a baseline per-stage threshold of $1/\sqrt{S}$. We use this threshold for the head and body stages and double it for the tail to account for its consistently larger observed gradient norm.

\section{Results}\label{sec:results}
\subsection{Setup}
\label{ssec:results_setup}
\paragraph{Pluralis-8B run details.}
We conducted a publicly open, collaborative, permissionless training run of an 8B Pluralis model (\Cref{ssec:model-card}).
We began the run on \textit{14 May 2026} and finished on \textit{23 June 2026}, for a total run duration of \textit{40 days}.
In total, the run processed 500B tokens, with a measured 30.8 ZettaFLOPs ($3.08 \times 10^{22}$) of compute performed over its full duration.
For the first 5,000 training steps, only the worker nodes hosted by us were admitted, with every stage running on 8 Pluralis nodes; the run was then opened to the public.
Thereafter the head and tail stayed on Pluralis nodes alone, at 8 and 16 respectively, while each body stage kept 4 Pluralis nodes, with contributors making up the rest.

\paragraph{Dataset and pre-processing.}
The FineWeb dataset~\cite{penedo2024fineweb} is a 15T token dataset curated from 96 Common Crawl~\cite{commoncrawl2025ccmain13} snapshots via filtering and deduplication strategies.
The pre-training mix our trainer roles used is FineWeb-Edu (1.3T tokens), an education-focused, filtered subsample of the larger FineWeb.
As discussed in \Cref{trainer_section}, each trainer streams one or more shards, and because the run requires a large number of trainers, we needed finer granularity than FineWeb-Edu's original 1,630 shards.
We therefore created a more granular sharding of FineWeb-Edu, composed of 10,080 shards.
Each new shard is a pre-shuffled mix of the original shards, allowing efficient streaming without the pre-buffering that normal streaming-shuffle approaches require.
We allocated 3 shards per trainer, so covering the full 500B tokens required about 1,290 trainers streaming through their shards over the run.

\begin{figure}[h]
    \centering
    \includegraphics[width=0.7\linewidth]{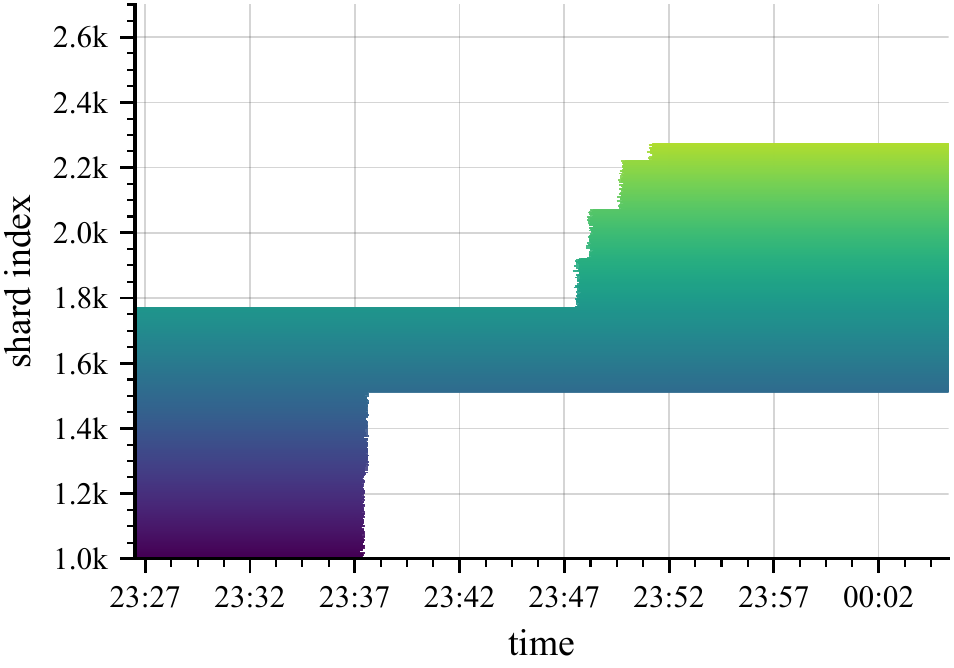}
    \caption{\textbf{The trainer shard indices during a shard rotation.}}
    \label{fig:trainer_shards}
\end{figure}

\paragraph{Dataset shard rotation.}
Trainers exhaust their shards during training, so we rotate their shards dynamically.
The Metrics Monitor (\Cref{sssec:metrics}) tracks each trainer's assigned shard and the number of tokens it has consumed from that shard.
This allows us to detect shard exhaustion and proactively assign new shards, keeping the trainer pool saturated and avoiding prolonged TPS drops.
\Cref{fig:trainer_shards} shows a shard-rotation operation: shards 1,008 to 1,512 are exhausted, and the trainers are stopped and rotated onto shards 1,764 to 2,272.

\subsection{Model Card}
\label{ssec:model-card}

Pluralis-8B is a decoder-only transformer trained over Agora with reparameterized subspace networks (\Cref{ssec:reparam_ssn}) on the pipeline-parallel axis and AsyncSPARTA (\Cref{ssec:async_sparta}) on the data-parallel axis.
The model is partitioned into seven pipeline stages: a head, five body stages, and a tail, for a total of 32 transformer blocks.
\Cref{tab:model-card-8b} summarizes its architecture, optimization recipe, and the public training run.

\begin{table}[h]
\centering
\begin{tabular}{ll}
\toprule
\textbf{Parameter} & \textbf{Value} \\
\midrule
\multicolumn{2}{l}{\textit{Architecture}} \\
Hidden dimension $d$ & 5,120 \\
Number of layers (total) & 32 \\
Layers per stage (head / body / tail) & 6 / 4 / 6 \\
Number of pipeline stages & 7 (1 head, 5 body, 1 tail) \\
Attention heads / KV heads (GQA) & 40 / 8 \\
Head dimension & 128 \\
FFN dim multiplier / rounding & 1.3 / multiple of 1,024 \\
Vocabulary size & 128,256 (Llama-3 tokenizer) \\
RoPE base $\theta$ & 500,000 \\
Max sequence length & 4,096 \\
Normalization & RMSNorm, $\varepsilon = 10^{-5}$, block scale frozen \\
QK-norm + reorder & Enabled \\
Subspace compression rate & $100\times$ (reparameterized SSN) \\
Trainable / dense-equivalent params & $\sim$8.6B / $\sim$12B \\
\midrule
\multicolumn{2}{l}{\textit{Optimization}} \\
Optimizer & AdamW \\
Peak learning rate & $3 \times 10^{-4}$ \\
$(\beta_1, \beta_2)$ / weight decay & $(0.9, 0.95)$ / 0.1 \\
LR schedule & Linear, 1\% floor \\
Warmup / schedule length & 4,000 / 154,900 steps \\
Global (target) batch size & 2,048 sequences ($\approx$8.4M tokens) \\
Data-parallel averaging & AsyncSPARTA \\
Sparse share $p$ / cadence $N$ & 5\% / every 20 steps \\
Delta rule updating & $\lambda = 1$ \\
Gradient clipping (head / body / tail) & 0.3779 / 0.3779 / 0.7558 (per-stage; no global clip) \\
Mixed precision / \texttt{torch.compile} & Enabled / enabled \\
\midrule
\multicolumn{2}{l}{\textit{Data \& training run}} \\
Training corpus & FineWeb-Edu (1.3T-token subset of FineWeb) \\
Sharding & 10,080 pre-shuffled shards, 3 per trainer \\
Tokens processed & 500B \\
Optimizer steps & 60k (LR schedule set to 154,900) \\
Public opening & After 5,000 steps \\
Run duration & 40 days (14 May -- 23 Jun 2026) \\
Compute performed & 30.8 ZettaFLOPs ($3.08 \times 10^{22}$) \\
Final training loss & 2.41 \\
\bottomrule
\end{tabular}
\caption{\textbf{Model and training configuration for Pluralis-8B.} The dense-equivalent
count is the size of the uncompressed model that the subspace factorization
represents.}
\label{tab:model-card-8b}
\end{table}

\paragraph{System configuration.}
Beyond the model and optimization recipe, the Pluralis-8B run is governed by the Agora protocol parameters that coordinate the nodes: the distributed hash table (DHT), the collaborative optimizer that schedules asynchronous averaging, and the sparse state all-reduce that keeps same-stage replicas in consensus.
\Cref{tab:system-config-8b} lists the settings used during the run.

\begin{table}[h]
\centering
\begin{tabular}{ll}
\toprule
\textbf{Parameter} & \textbf{Value} \\
\midrule
\multicolumn{2}{l}{\textit{DHT \& coordination}} \\
Number of DHT instances & 8 (one per stage + one shared by trainers) \\
Connection handlers per worker & 7 \\
Worker re-announce period & 30\,s \\
Stats report interval & 60\,s \\
\midrule
\multicolumn{2}{l}{\textit{Collaborative optimizer}} \\
Target batch size & 2,048 microbatches \\
Auto step time & 3.0\,s \\
Matchmaking time & 40.0\,s \\
Averaging timeout & 150.0\,s \\
Optimizer offload & Enabled \\
Delayed (async) weight averaging & Enabled \\
Sync-phase steps (weight / optimizer) & 400 / 100 \\
Load-state block window & 2 \\
\midrule
\multicolumn{2}{l}{\textit{All-reduce}} \\
Chunk (part) size & 2\,MB \\
Compression & None \\
Sender / reducer timeout & 30\,s / 60\,s \\
\midrule
\multicolumn{2}{l}{\textit{Gradient averaging}} \\
Method & Local accumulation (no cross-replica gradient all-reduce) \\
\midrule
\multicolumn{2}{l}{\textit{Batching \& checkpointing}} \\
Batch size per trainer & 1 \\
Checkpoint interval & Every 500 steps \\
Checkpoint store & Cloud object storage (S3) \\
\bottomrule
\end{tabular}
\caption{\textbf{Agora system and coordination configuration for the Pluralis-8B run.}}
\label{tab:system-config-8b}
\end{table}

\subsection{Contributors and Participation}
\label{ssec:contributors}

\subsubsection{Contributors}
\begin{figure}[tbp]
\centering
\begin{subfigure}[t]{0.49\linewidth}
\centering
\includegraphics[width=\linewidth]{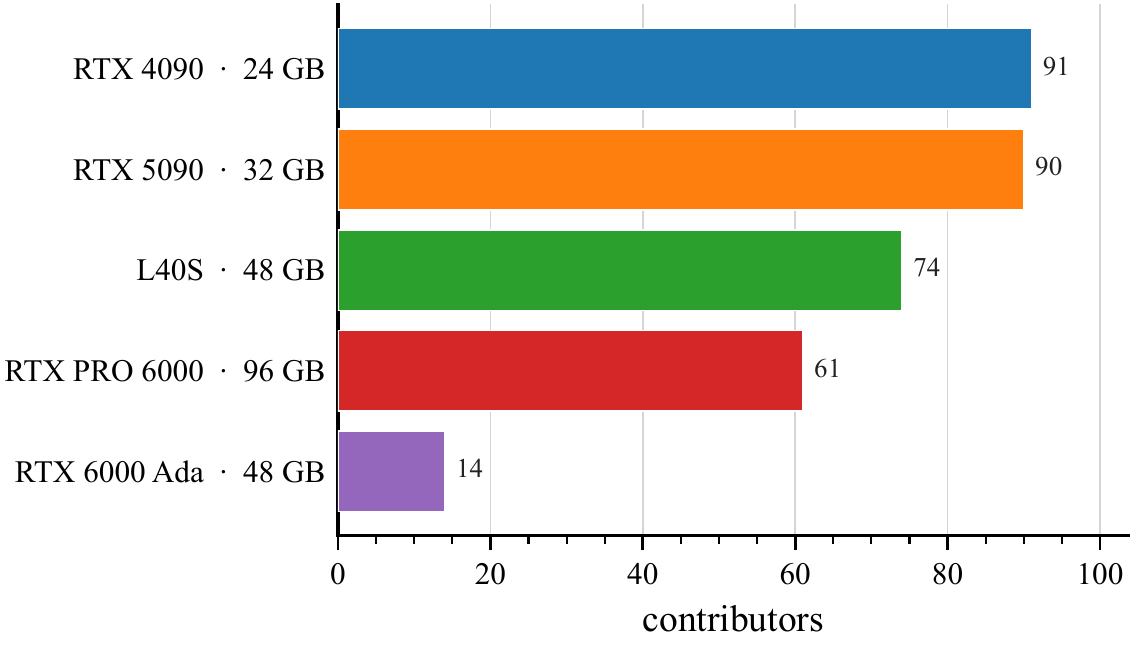}
\caption{GPU models.}
\label{fig:gpu_model_distribution}
\end{subfigure}
\hfill
\begin{subfigure}[t]{0.49\linewidth}
\centering
\includegraphics[width=\linewidth]{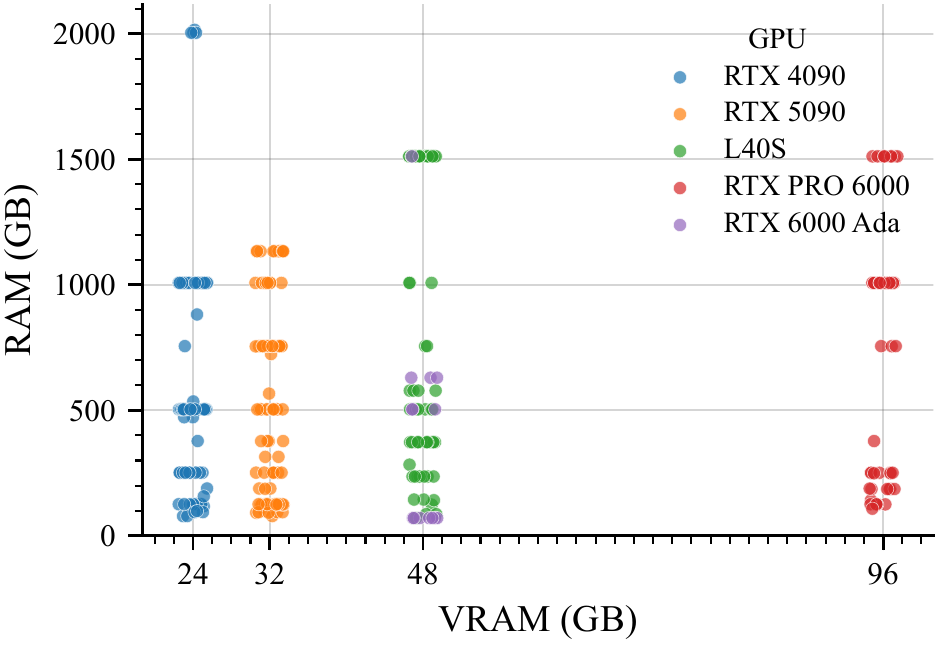}
\caption{Per-node VRAM vs.\ RAM.}
\label{fig:vram_vs_ram}
\end{subfigure}

\begin{subfigure}[t]{0.49\linewidth}
\centering
\includegraphics[width=\linewidth]{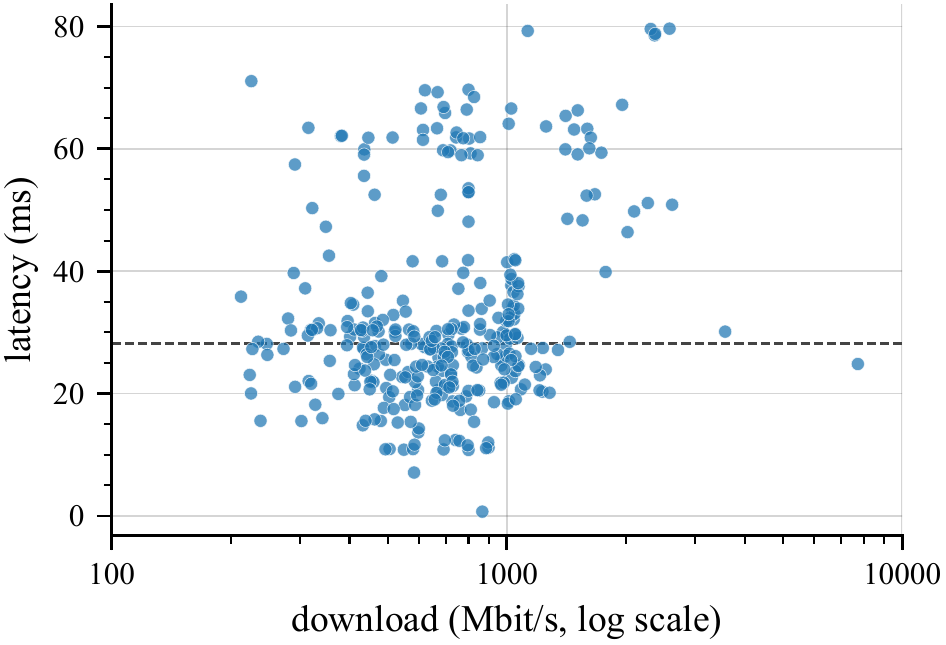}
\caption{Download bandwidth vs.\ latency.}
\label{fig:bandwidth_vs_latency}
\end{subfigure}
\caption{\textbf{Hardware heterogeneity of the 330 contributor nodes.}
\textbf{(a)} Node count by GPU model: participation was restricted to consumer cards plus the L40S, and the RTX~4090 and RTX~5090 together accounted for 55\% of nodes.
\textbf{(b)} Per-node RAM against VRAM: RAM varied largely independently of GPU class, from the 80\,GB admission floor to 2\,TB (median 504\,GB).
\textbf{(c)} Per-node download bandwidth against network latency, with the dashed line marking the median latency of 28\,ms: bandwidth ranged from 213\,Mbit/s to 7.7\,Gbit/s (median 703\,Mbit/s) and latency extended to the 80\,ms admission ceiling.}
\label{fig:hardware_heterogeneity}
\end{figure}

Alongside the nodes that Pluralis operated itself, the run was open to any external contributor who wished to join with their own hardware.
Admission imposed a minimum hardware requirement\footnote{At least 24\,GB of VRAM, 80\,GB of RAM, 200\,Mbit/s of download bandwidth, and network latency no greater than 80\,ms.} and restricted participation to consumer GPU models (we also allowed the L40S, the datacenter equivalent of the RTX~6000~Ada) (\Cref{fig:hardware_heterogeneity}).
Agora has no technical barrier to incorporating more powerful datacenter GPUs; however, we made this choice to
\begin{enumerate*}[label=(\arabic*)]
    \item test the system in the most adversarial setting possible, where all GPUs were small and connected over internet-grade links, and
    \item show that Agora does not depend on datacenter compute.
\end{enumerate*}
The RTX~5090 and RTX~4090 together accounted for 55\% of nodes, the L40S for 22\%, and the RTX~6000~Ada and RTX~PRO~6000 for the remaining 23\%.
RAM varied continuously and largely independently of GPU class (\Cref{fig:vram_vs_ram}), from the 80\,GB admission floor to 2\,TB (median 504\,GB).
Network conditions were similarly broad (\Cref{fig:bandwidth_vs_latency}): download bandwidth ranged from 213\,Mbit/s to 7.7\,Gbit/s (median 703\,Mbit/s) and latency extended to the 80\,ms admission ceiling, both distributions truncated by the entry requirements.

\subsubsection{Participation Dynamics}
\begin{figure}[tbp]
\centering
\begin{subfigure}[t]{0.49\linewidth}
\centering
\includegraphics[width=\linewidth]{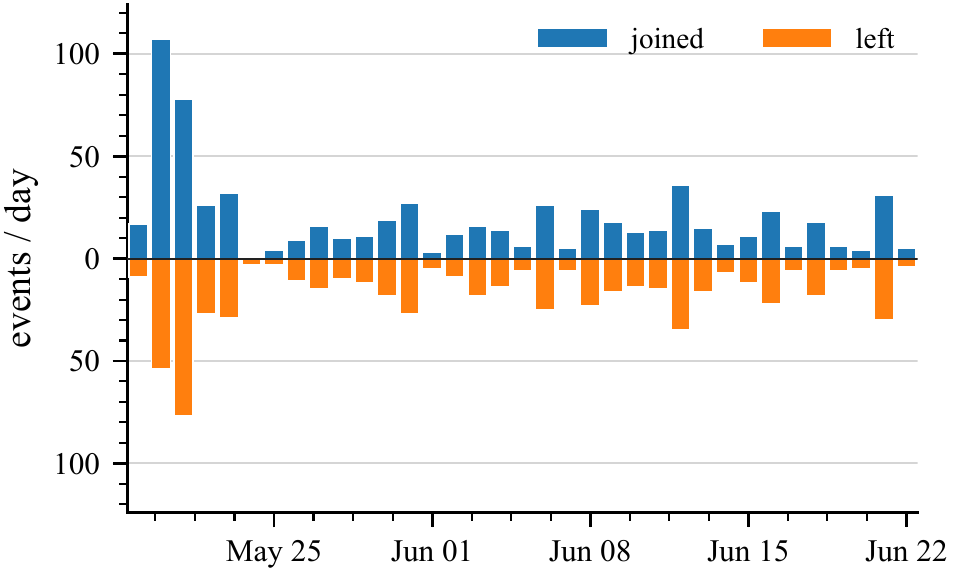}
\caption{Daily joins and drops.}
\label{fig:daily_joins_drops}
\end{subfigure}
\hfill
\begin{subfigure}[t]{0.49\linewidth}
\centering
\includegraphics[width=\linewidth]{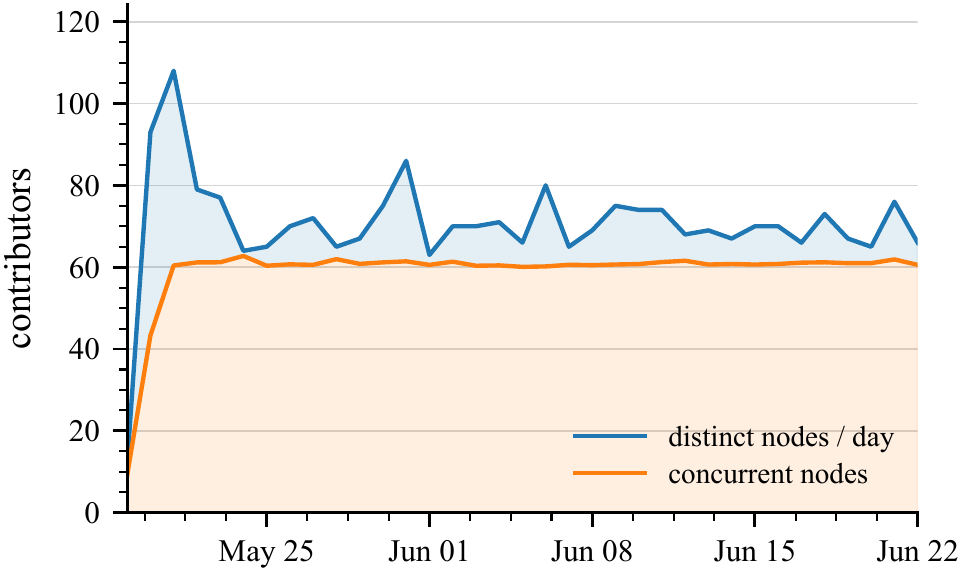}
\caption{Distinct contributors per day vs.\ concurrent fleet.}
\label{fig:distinct_vs_concurrent_fleet}
\end{subfigure}

\begin{subfigure}[t]{0.49\linewidth}
\centering
\includegraphics[width=\linewidth]{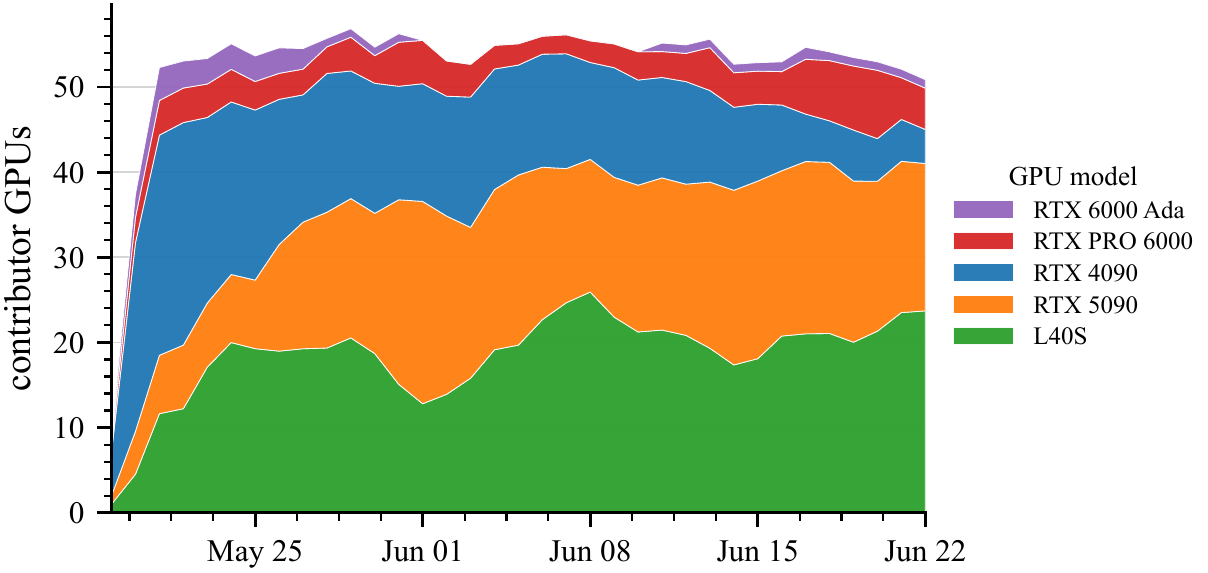}
\caption{Active contributor GPUs over time, by model.}
\label{fig:gpu_composition_over_time}
\end{subfigure}
\hfill
\begin{subfigure}[t]{0.49\linewidth}
\centering
\includegraphics[width=\linewidth]{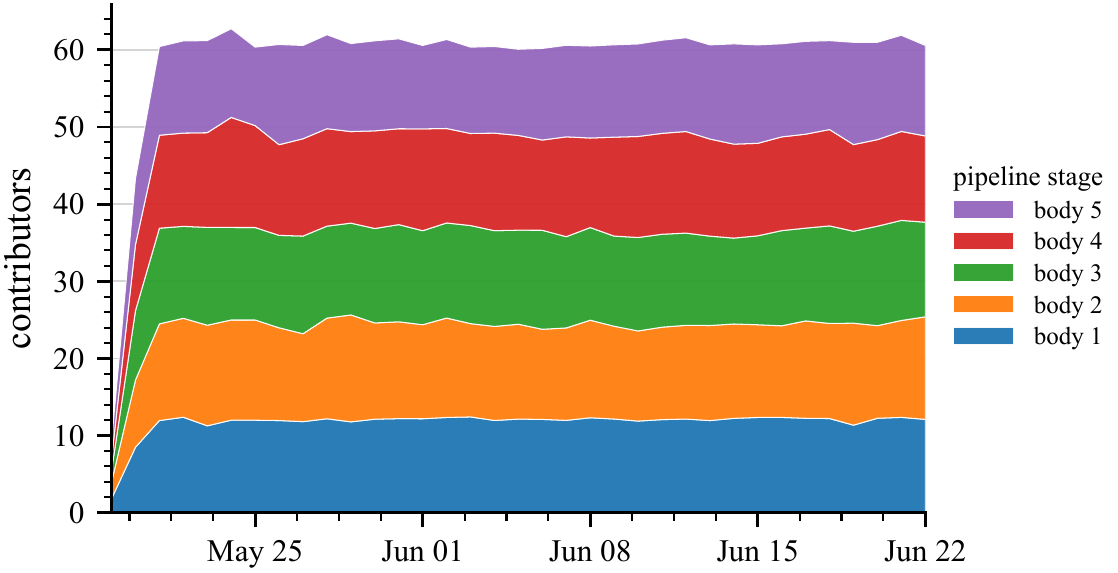}
\caption{Active contributors over time, by pipeline stage.}
\label{fig:stage_composition_over_time}
\end{subfigure}
\caption{\textbf{Contributor fleet over the run.}
\textbf{(a)} Daily join and drop events: 669 joins and 607 drops in total, with an initial wave of 107 joins on 20~May and steady turnover thereafter (a node that dropped and rejoined is counted each time).
\textbf{(b)} Distinct contributors seen per day against the concurrent fleet, which reached the 60-node admission cap within two days.
\textbf{(c)} Live fleet composition by GPU model: the mix shifted from the RTX~4090 toward the RTX~5090, L40S, and RTX~PRO~6000, raising the fleet's total VRAM at an unchanged node count.
\textbf{(d)} Live fleet composition by pipeline stage: each of the five body stages held about a fifth of the nodes throughout, so no stage was starved or overloaded despite the churn.}
\label{fig:fleet_over_run}
\end{figure}

The run admitted at most 60 contributors at a time, a ceiling reached within two days of the first admission and then held through continuous turnover (\Cref{fig:fleet_over_run}).
The concurrent live fleet (\Cref{fig:distinct_vs_concurrent_fleet}) sat just above that cap from 21 May through the remaining five weeks, at about 61 nodes and peaking at 63.\footnote{The slight excess over 60 arose from race conditions during admission, which occasionally let a few extra contributors in at once.}
Roughly 70 distinct contributors appeared on a typical day, and as many as 108 on the busiest day, early in the run.
Rather than the same machines staying connected throughout, membership turned over constantly (\Cref{fig:daily_joins_drops}): 669 nodes joined and 607 left over the run.\footnote{Joins and drops count events, not distinct nodes: a node that dropped and later rejoined the run is counted each time.}
After an initial wave (107 joins on 20 May) when the run was not at capacity, joins settled to a few to a few dozen per day, taking the place of existing nodes that dropped.  
Over the course of the run, the GPU mix shifted toward newer, higher-memory cards (\Cref{fig:gpu_composition_over_time}): the RTX~4090 fell from about 32\% of the fleet in late May to about 9\% by late June, while the RTX~5090 rose from 17\% to 31\%, the L40S from 29\% to 35\% (becoming the single largest model), and the RTX~PRO~6000 from 17\% to 23\%, so the fleet's total live VRAM rose from roughly 2.8 to 3.1\,TB even though the node count was unchanged.
Throughout, the fleet stayed evenly balanced across the five pipeline body stages (\Cref{fig:stage_composition_over_time}), each holding about a fifth of the nodes (19.5\% to 20.4\%) for the entire run, with the largest single-day gap between the busiest and quietest stage at only about three contributors, so no stage was ever materially starved or overloaded even as nodes continually joined and left.
As far as we are aware, this is the first example of a billion-plus parameter scale training run that made use of a dynamic, heterogeneous mix of GPUs. 

\subsubsection{Demand and Admission}
\begin{figure}[tbp]
\centering
\begin{subfigure}[t]{0.49\linewidth}
\centering
\includegraphics[width=\linewidth]{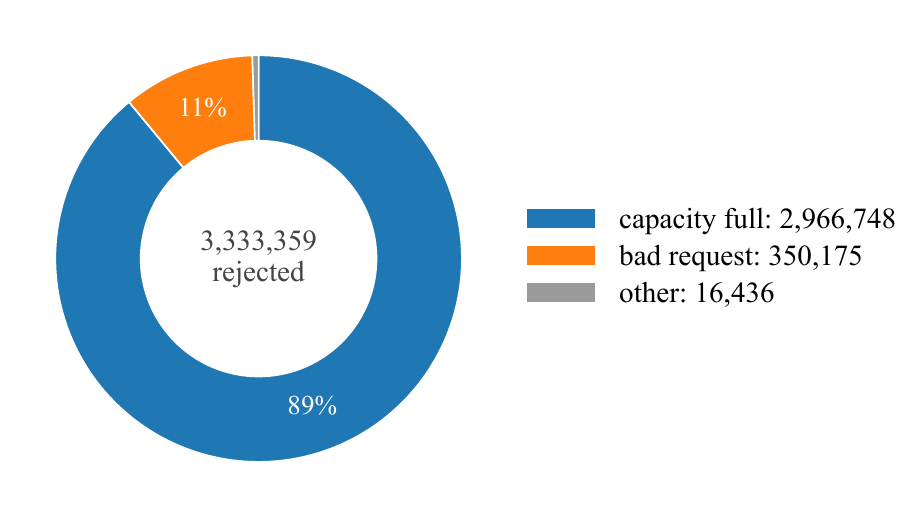}
\caption{Rejection reasons.}
\label{fig:rejection_reasons}
\end{subfigure}
\hfill
\begin{subfigure}[t]{0.49\linewidth}
\centering
\includegraphics[width=\linewidth]{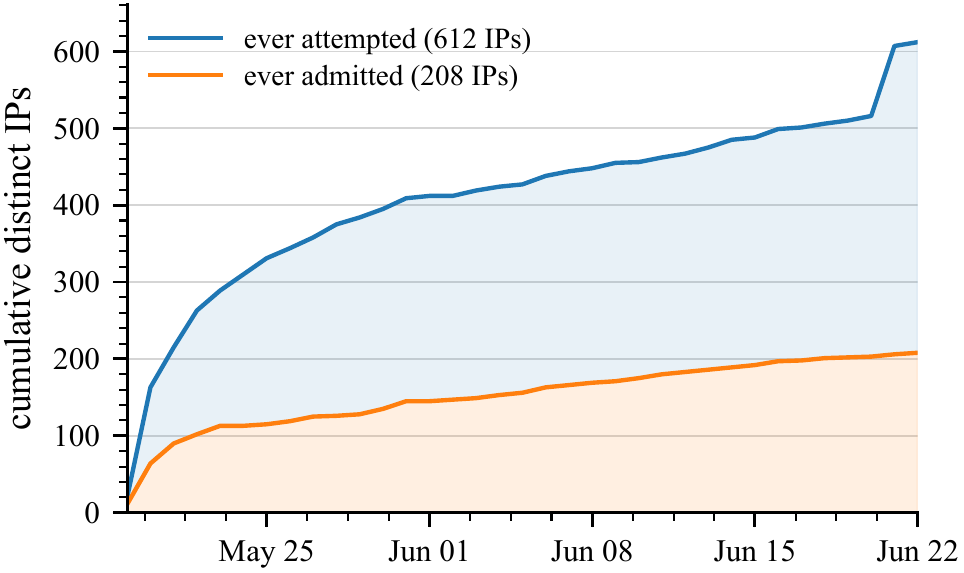}
\caption{Cumulative IPs: attempted vs.\ admitted.}
\label{fig:cumulative_join_funnel}
\end{subfigure}

\medskip

\begin{subfigure}[t]{0.49\linewidth}
\centering
\includegraphics[width=\linewidth]{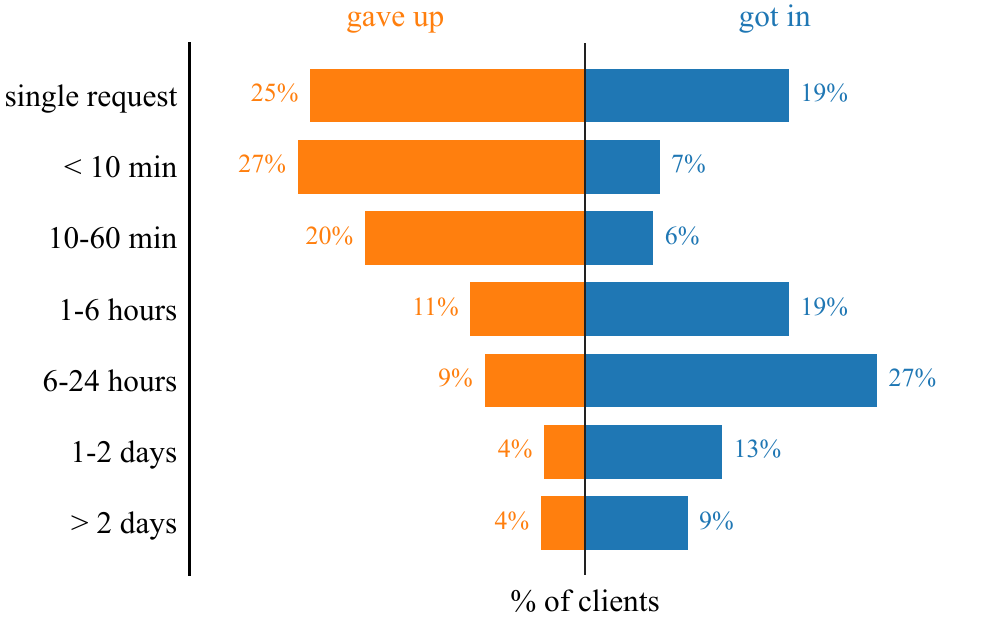}
\caption{How long clients kept trying, by outcome.}
\label{fig:admission_wait_ecdf}
\end{subfigure}
\hfill
\begin{subfigure}[t]{0.49\linewidth}
\centering
\includegraphics[width=\linewidth]{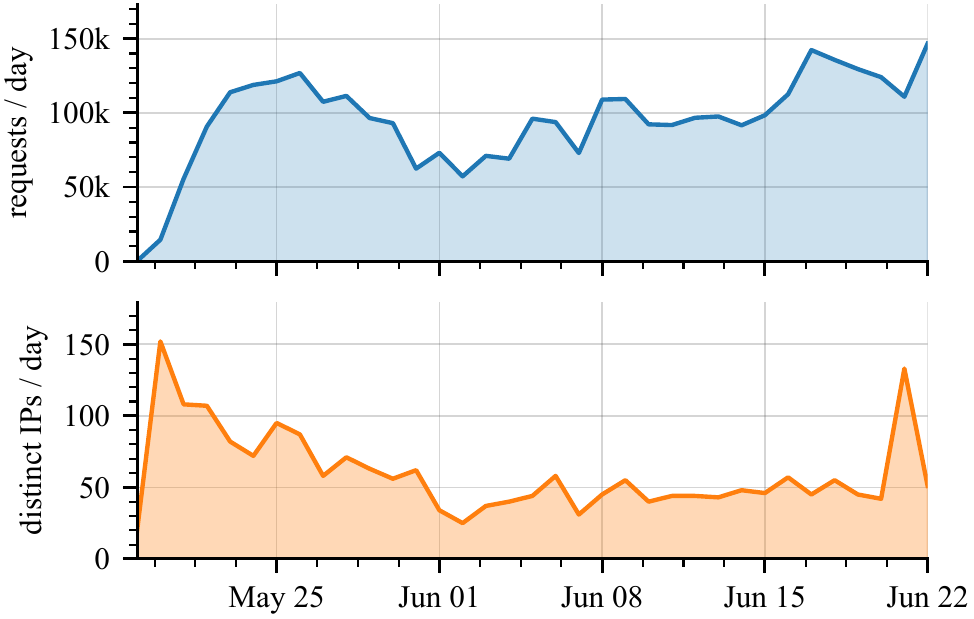}
\caption{Daily requests vs.\ distinct IPs active.}
\label{fig:daily_requests_vs_ips}
\end{subfigure}
\caption{\textbf{Join demand and admission over the run.}
Each client is counted by IP address.
\textbf{(a)} Rejection reasons for the 3.33 million rejected join requests (99.98\% of all requests): 89\% arrived while the run was at capacity, 11\% were malformed, and under 0.5\% failed for other causes.
\textbf{(b)} Cumulative distinct clients that ever attempted to join (612) and that were ever admitted (208): demand kept arriving throughout the run rather than saturating early.
\textbf{(c)} How long clients kept requesting, split by outcome and shown as the share of clients within each outcome: admitted clients typically waited hours (median 5.8 hours), while clients that gave up did so quickly (median 8 minutes).
\textbf{(d)} Join requests and distinct clients active per day: tens of clients generated roughly 100,000 requests daily.}
\label{fig:join_admission}
\end{figure}

Demand to join far outstripped capacity (\Cref{fig:join_admission}).
Over the run, contributors made 3,334,054 join requests from 612 distinct clients (by IP address), and 99.98\% were turned away (\Cref{fig:rejection_reasons}): almost all rejections (89.0\%) were because the run was already at capacity, with a further 10.5\% malformed requests and under 0.5\% from other causes (an outdated or modified client, an unreachable node behind a closed port or firewall, or an invalid access token).
Only 695 requests (0.02\%) succeeded, but that counts requests, not clients: a single client often sent thousands of requests before being admitted.
Per client (\Cref{fig:cumulative_join_funnel}), 208 of the 612 (34\%) were admitted at least once and 404 (66\%) never were.
Demand kept arriving throughout the run rather than saturating early.
Many admitted clients returned to the run more than once: 106 of the 208 were admitted again on later occasions.
When a client did get in, the wait was usually measured in hours (\Cref{fig:admission_wait_ecdf}): a median of 5.8 hours, rising to two days or more for the slowest tenth (up to 11 days), though about one in five got in on the very first request.
Clients that gave up without getting in usually did so quickly, after a median of 8 minutes (a quarter after a single request), though the most persistent kept trying for up to 18 hours.
Only tens of distinct clients were active on a typical day (\Cref{fig:daily_requests_vs_ips}), yet they generated roughly 100,000 requests daily (peaking near 147,000 on 22 June).

\subsection{Throughput and Efficiency} 
\label{ssec:results_throughput}

Unlike a centralized cluster, where identical accelerators on a dedicated interconnect make throughput essentially a property of the hardware, over Agora it is shaped by compute heterogeneity, link bandwidth and latency, and continual node churn, and by how the pooled compute is balanced across the pipeline stages (\Cref{ssec:throughput_optimization}).
Here we report the throughput and efficiency the Pluralis-8B run achieved, and how compute was balanced across stages to reach them.

\begin{figure}[t]
    \centering
    \begin{subfigure}[t]{0.54\linewidth}
        \centering
        \includegraphics[width=\linewidth]{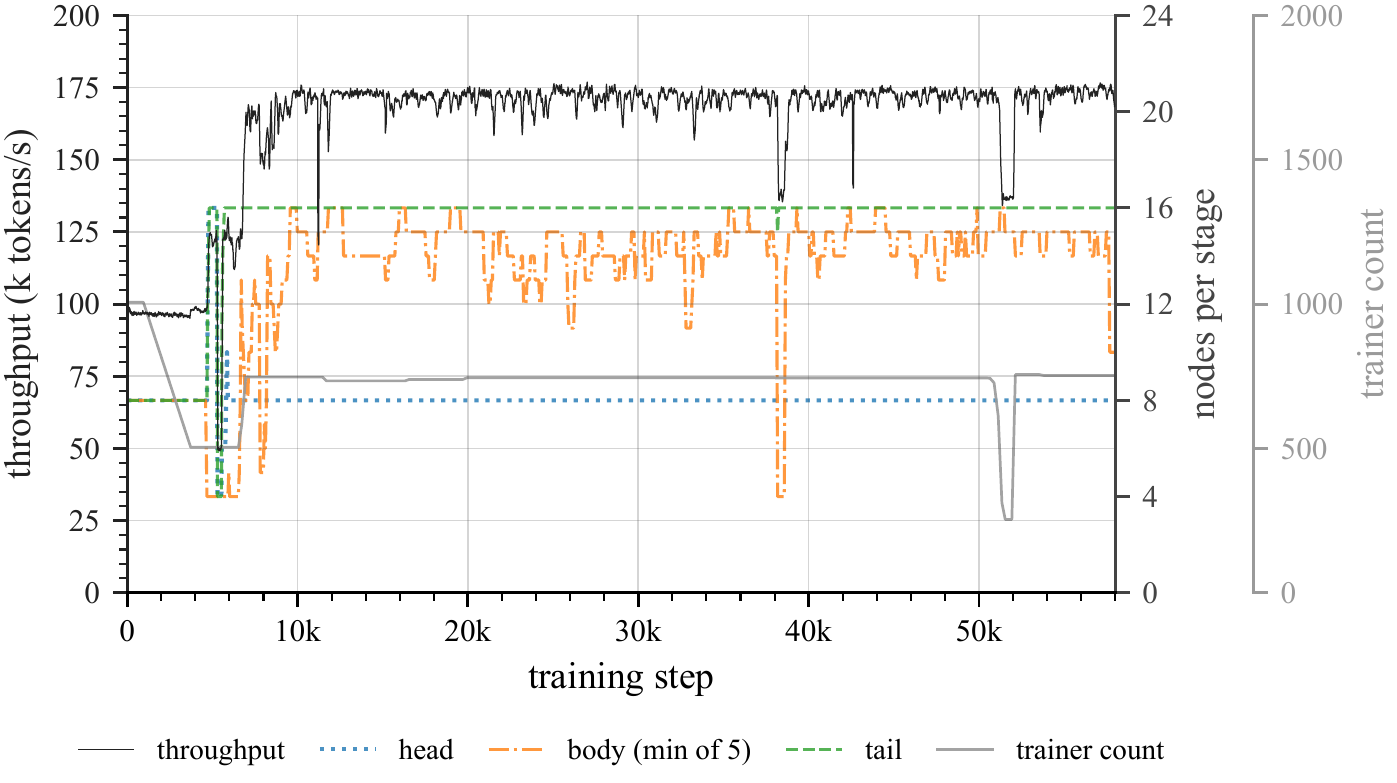}
        \caption{Throughput, node counts, and trainers over the run.}
        \label{fig:live_tps_nodes}
    \end{subfigure}
    \hfill
    \begin{subfigure}[t]{0.45\linewidth}
        \centering
        \includegraphics[width=\linewidth]{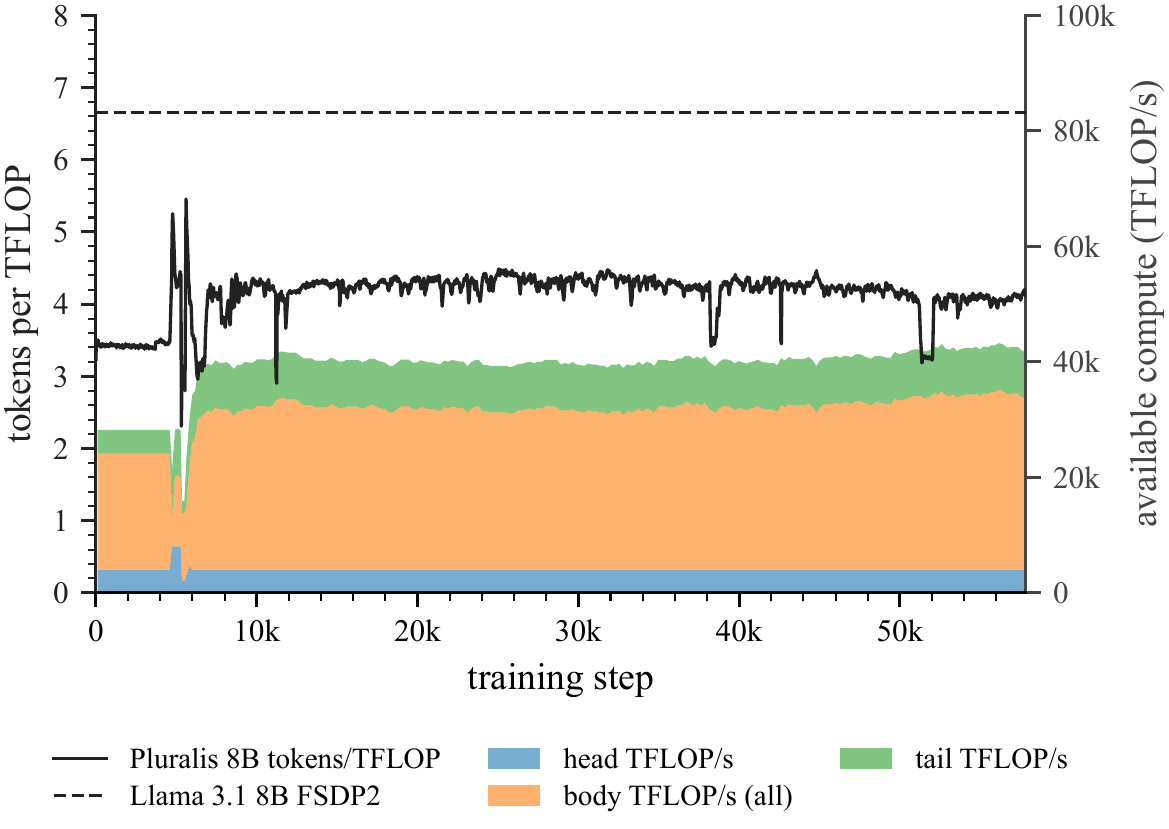}
        \caption{Efficiency vs.\ a centralized baseline.}
        \label{fig:live_tps_eff}
    \end{subfigure}
    \caption{\textbf{Throughput and efficiency.}
    \textbf{(a)} Pipeline throughput, per-stage node counts, and trainer count over the run; for the body stages we plot the minimum node count across them.
    \textbf{(b)} Pipeline efficiency (tokens per pooled TFLOP) vs.\ a centralized baseline (Llama 3.1 8B, H100, FSDP2).}
    \label{fig:fig_live_tps}
\end{figure}

\paragraph{Throughput.}
After an initial ramp, the run settled into a stable regime: pipeline throughput rose to a steady $\sim$170k tokens/s, sustained by 745 active trainers, with 8 workers on the head stage, 16 on the tail, and a median of $\sim$15 across the five body stages (\Cref{fig:live_tps_nodes}; there we plot the size of the smallest, and hence slowest, body stage at each point in the run).
Throughput held at this level aside from two transient drops, around steps 38k and 52k: the first from a large loss of compute in one body stage that bottlenecked the pipeline, the second from a drop in active trainers that starved it of batches.
We define pipeline throughput (tokens/s; TPS) as:
\[
\mathrm{TPS} \;=\; \frac{B_{\mathrm{global}} \cdot L_{\mathrm{seq}}}{t_{\mathrm{optim\_step}}},
\]
where $B_{\mathrm{global}}$ is the global batch size, $L_{\mathrm{seq}}$ the sequence length, and $t_{\mathrm{optim\_step}}$ the wall-clock time (in seconds) of one optimizer step (the time it took to compute gradients for $B_{\mathrm{global}}$ samples).

The measured throughput must be the same in every stage, because all stages process the same samples routed by the trainers, in both the forward and backward passes, so no stage can get ahead of the others.
The entire pipeline therefore always slows to the pace of its slowest stage, as the adjacent stages must wait for it to process its samples. 

\paragraph{Efficiency.}
In Agora we aim to maximize the efficiency of the pipeline measured as the achieved throughput (TPS) per unit of total compute capacity pooled in the run (TFLOP/s):
\[
\eta \;=\; \frac{\mathrm{TPS}}{C_{\mathrm{pool}}},
\]
where $C_{\mathrm{pool}}$ is the total spec-sheet reported dense-BF16 compute pooled across all participating GPUs (in TFLOP/s), so that $\eta$ is expressed in tokens per TFLOP. 

\paragraph{Centralized baseline.}
We benchmark Agora's efficiency against the baseline of a comparable centralized setting, namely the efficiency achieved by training a Llama 3.1 8B model, using TorchTitan with FSDP2, mixed precision, and \texttt{torch.compile}, in a cluster of H100 GPUs~\citep{liang2025torchtitan}.
This is presented in \Cref{fig:live_tps_eff}: Agora achieved a peak efficiency of 5.45 tokens per TFLOP and a steady-state 4.2, about 63\% of the centralized baseline efficiency.
We regard this gap as modest given that Pluralis-8B was trained over heterogeneous, geographically distributed GPUs rather than a tightly coupled cluster.
Notably, the system was also resilient to churn, maintaining its stable throughput while contributor nodes joined and left throughout the training run (\Cref{fig:live_tps_nodes}).

\paragraph{Balancing compute across stages.}
Since the pipeline is paced by its slowest stage, efficiency is maximized when the pooled compute is balanced so that no stage bottlenecks the others; where a stage falls behind, we add compute to it: Pluralis nodes on the head and tail, a higher contributor cap on the bodies (\Cref{ssec:throughput_optimization}).
We read this balance off the per-microbatch communication and compute times, GPU utilization, and coalesced microbatch size (\Cref{sec:comms_window,batch_coalescing}); for the bodies, contributor connection quality makes raw GPU utilization misleading, so we rely on the coalesced microbatch size, which sits near 1 when a stage is starved and near its cap when work is piling up faster than it can process.

\paragraph{Stage dynamics.}
In the run, this balance settled with the tail as the compute-bound bottleneck: it was the heaviest stage (6 transformer layers plus the de-embedding matrix and loss; \Cref{tab:stage_layers}) and held the largest share of the pooled compute, yet still paced the pipeline (\Cref{tab:stage_breakdown}).
We nonetheless did not add tail nodes: the extra replicas would have raised the risk of weight divergence (\Cref{ssec:sparta_fault_tolerence}), and the slower convergence would have outweighed the throughput gain.
The head was the lightest stage, with 8 Pluralis nodes (4,030\,TFLOP/s) at 87.4\% utilization matching the tail's utilization and coalesced-microbatch size.
The body stages ran less utilized than head and tail, and we kept them deliberately over-provisioned so that contributor churn did not dent throughput.

  \begin{table}[t]
    \centering
    \begin{tabular}{lr}
      \toprule
      Stage & Model layers \\
      \midrule
      head   & 6 transformer blocks + embeddings \\
      body-1 & 4 transformer blocks \\
      body-2 & 4 transformer blocks  \\
      body-3 & 4 transformer blocks  \\
      body-4 & 4 transformer blocks  \\
      body-5 & 4 transformer blocks  \\
      tail   & 6 transformer blocks + de-embedding \\
      \bottomrule
    \end{tabular}
    \caption{\textbf{Distribution of model layers across pipeline stages.}}
    \label{tab:stage_layers}
  \end{table}

\begin{table}[t]
\centering
\begin{tabular}{lrrrrrrrr}
  \toprule
        & \multicolumn{2}{c}{microbatch p50 (ms)} & \multicolumn{2}{c}{comms/compute (\%)} & microbatch & GPU util & MFU & TFLOP/s \\
  \cmidrule(lr){2-3} \cmidrule(lr){4-5}
  Stage & comms & compute & p50/p50 & p80/p20 & avg size & (\%) & (\%) & pooled \\
  \midrule
  head   &  19 & 293 &  6 &  11 & 2.9 & 87.4 & 38.9 & 4,030 \\
  body-1 & 138 & 261 & 53 & 168 & 1.1 & 44.8 & 22.4 & 5,568 \\
  body-2 & 150 & 283 & 53 & 189 & 1.1 & 42.0 & 20.9 & 5,711 \\
  body-3 & 140 & 306 & 46 & 172 & 1.1 & 50.4 & 18.6 & 5,632 \\
  body-4 & 149 & 282 & 53 & 169 & 1.1 & 45.7 & 19.2 & 5,402 \\
  body-5 & 128 & 255 & 50 & 153 & 1.1 & 45.9 & 19.9 & 5,345 \\
  tail   &  23 & 501 &  5 &  19 & 2.8 & 80.9 & 36.0 & 8,061 \\
  \midrule
  \textbf{all} & -- & -- & -- & -- & -- & \textbf{54.4} & \textbf{24.2} & \textbf{39,749} \\
  \bottomrule
\end{tabular}
\caption{\textbf{Per-stage communication and compute breakdown.} Measured over the stable 20k--30k-step window, when the run held its peak throughput (171.5k~tokens/s at 4.31 tokens per TFLOP). The comms/compute columns report the ratio of microbatch await time to compute time, at the median (p50/p50) and for fast-compute nodes on slow connections (p80/p20).}
\label{tab:stage_breakdown}
\end{table}

\begin{figure}[t]
    \centering
    \includegraphics[width=0.6\linewidth]{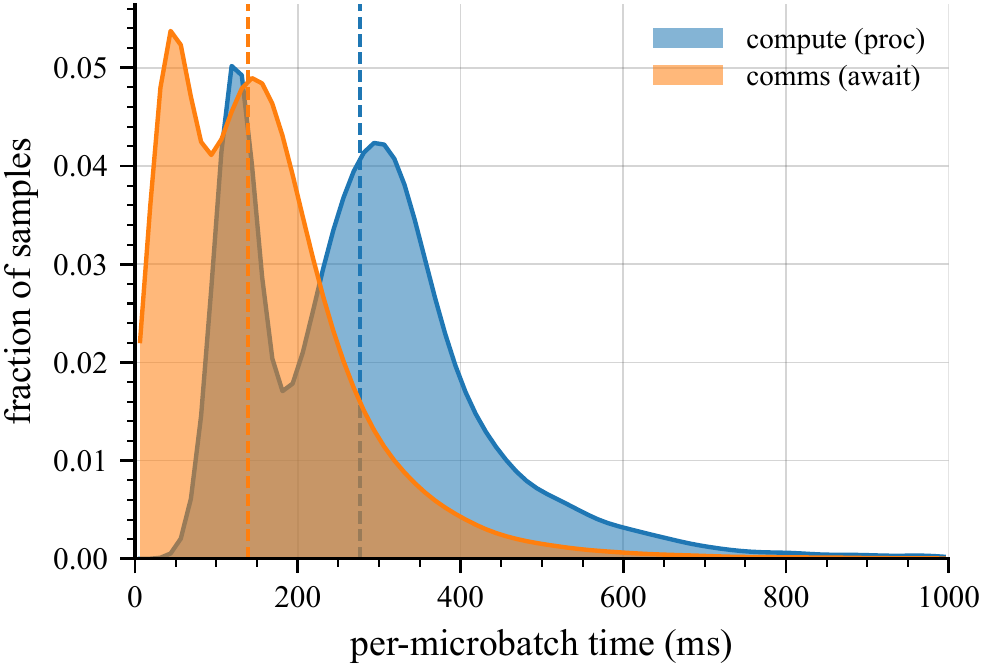}
    \caption{\textbf{Communication vs.\ computation time in body stages.}
    Per-microbatch communication (await) and computation time across all body-stage (contributor) nodes, over the stable 20k--30k-step window.
    In the median, communication was shorter than computation (139\,ms vs.\ 276\,ms), so the body stages ran compute-bound.}
    \label{fig:body_timing_hist}
\end{figure}

\needspace{6\baselineskip}
\subsection{Training Loss}
\begin{figure}[t]
    \centering
    \begin{subfigure}{0.49\linewidth}
        \centering
        \includegraphics[width=\linewidth]{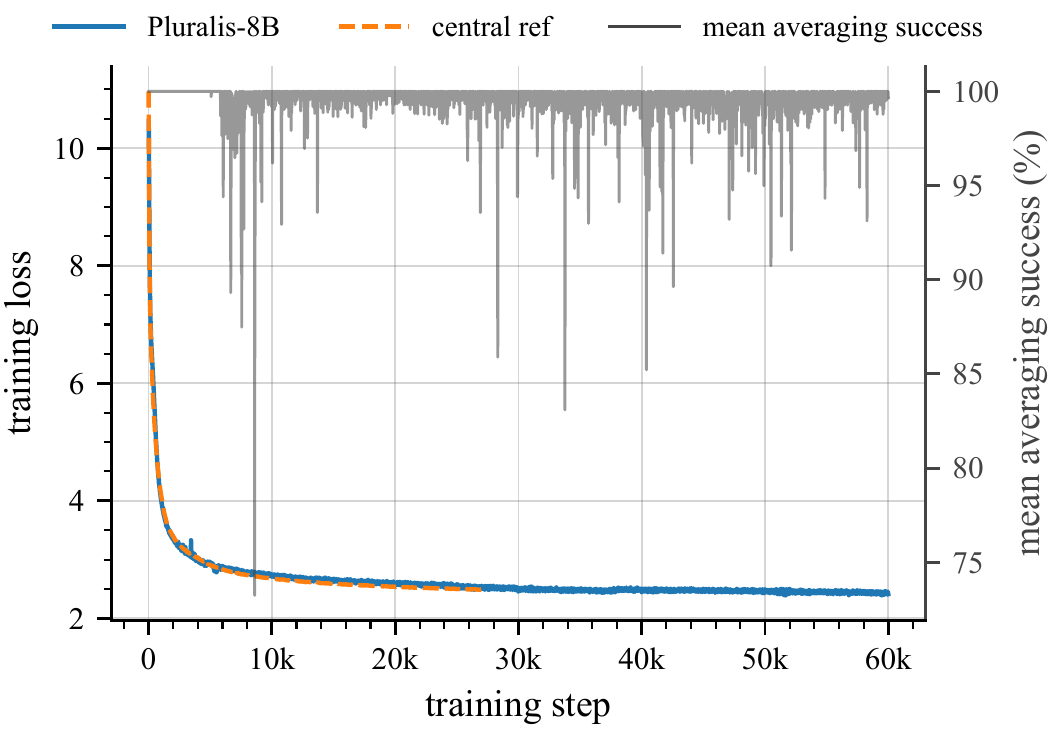}
        \caption{Full run.}
        \label{fig:training_loss_full}
    \end{subfigure}\hfill
    \begin{subfigure}{0.49\linewidth}
        \centering
        \includegraphics[width=\linewidth]{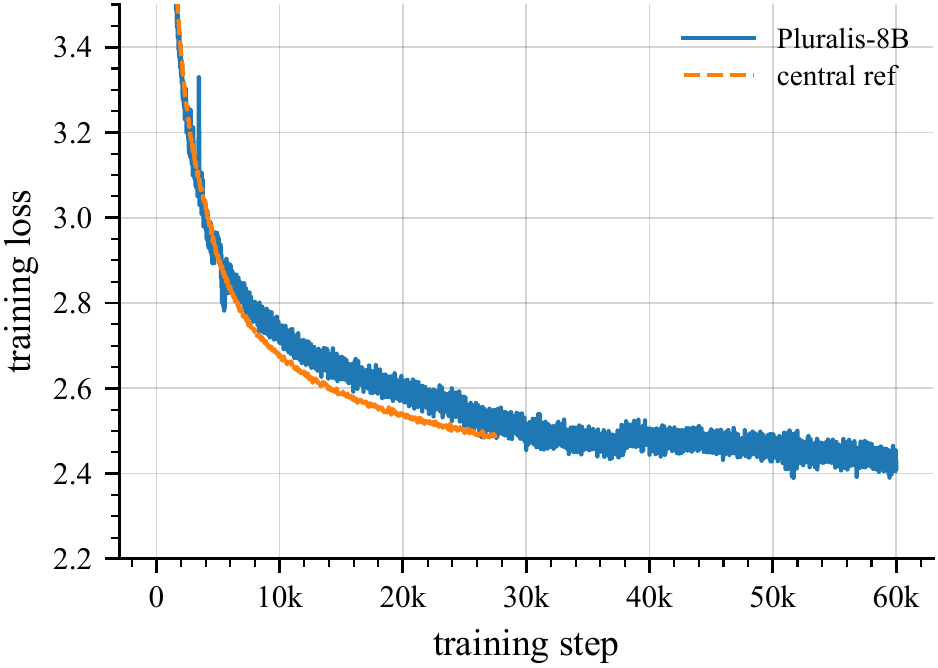}
        \caption{Zoomed y-axis.}
        \label{fig:training_loss_zoom}
    \end{subfigure}
    \caption{\textbf{Pluralis-8B training loss against a central reference curve.}
    \textbf{(a)} The full run, with the average success of workers in averaging their weight shard overlaid above.
    \textbf{(b)} The same loss curves zoomed in on the y-axis. The early jaggedness coincides with the period of largest weight change; the perturbation near step 40k follows a simultaneous exit of many contributor nodes, from which the loss recovered without intervention.}
    \label{fig:training_loss}
\end{figure}

Since the goal of this work is to show that a system trained with Agora matches standard centralized training, rather than to produce a state-of-the-art model, training loss against a centralized reference on identical data is the natural figure of merit. The reference is a TorchTitan~\citep{liang2025torchtitan} run modified to use the same subspace-network reparameterization (\Cref{ssec:reparam_ssn}) and SPARTA averaging (\Cref{ssec:async_sparta}) as Pluralis-8B, but executed on a dedicated homogeneous cluster with synchronous, fully reliable collectives. The reference thus carries the same compression-induced effects on the loss, so any difference between the curves reflects the decentralized execution itself: heterogeneity, partial all-reduce participation, and node churn, rather than the model changes. The training loss of the Pluralis-8B run converged to a value of 2.41 after 60k optimizer steps and 500B tokens.
\Cref{fig:training_loss_full} shows the Pluralis-8B training loss throughout the entire run alongside a central reference curve run for a shorter period of 27k iterations, with the average success rate of workers averaging their weight shards overlaid above. \Cref{fig:training_loss_zoom} shows a zoomed-in view of the y-axis for both training curves. The Pluralis-8B curve is visibly more jagged in the early portion of the run, when the weights change the most and gradients are still high, leaving replicas most sensitive to partial averaging; the jitter smooths out as training progresses and the gradients settle. A brief perturbation is also visible around step 40k, when a large set of contributor nodes exited the run simultaneously. Such mass departures can stem from various causes, such as an operator's decision to withdraw, a compute-provider outage, or a network glitch. Nevertheless, the loss recovered within a few thousand steps without intervention. While the central curve sits slightly lower, the Pluralis-8B curve still converged, despite heterogeneity and workers intermittently failing to complete the full all-reduce.

\begin{figure}[t]
    \centering
    \begin{subfigure}[t]{0.48\linewidth}
        \centering
        \includegraphics[width=\linewidth]{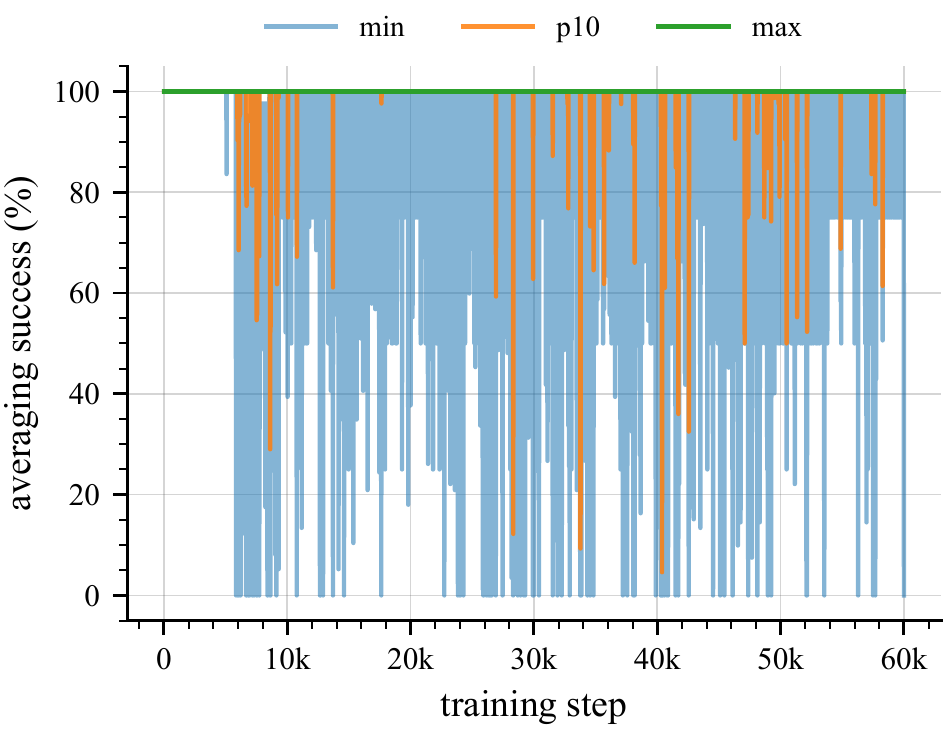}
        \caption{Averaging.}
        \label{fig:average_pcnt}
    \end{subfigure}
    \hfill
    \begin{subfigure}[t]{0.48\linewidth}
        \centering
        \includegraphics[width=\linewidth]{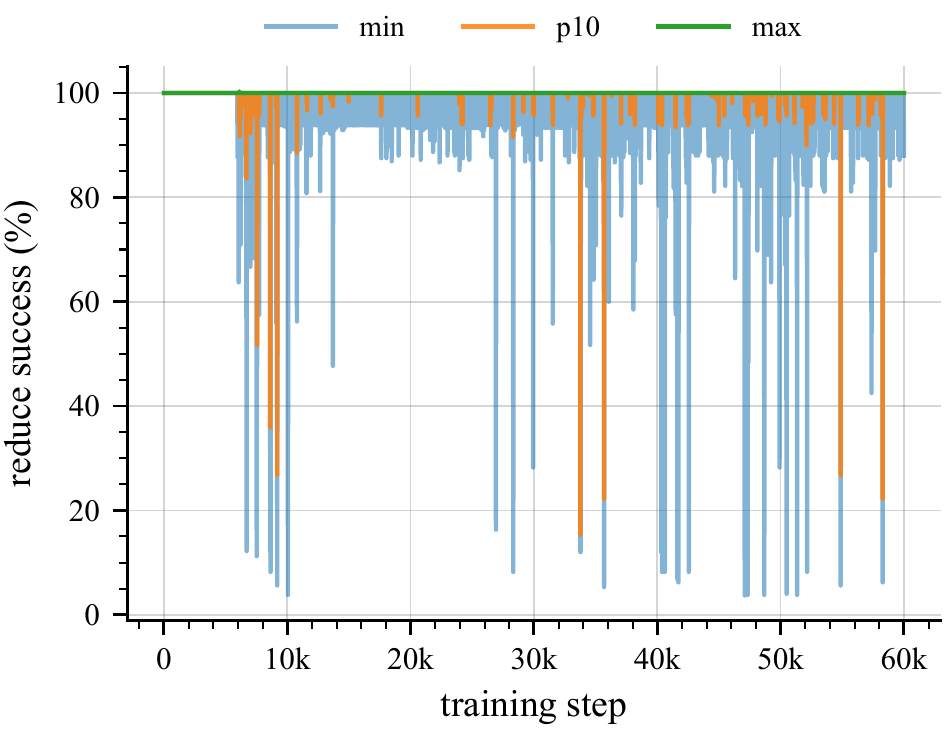}
        \caption{Reduce.}
        \label{fig:reducer_pcnt}
    \end{subfigure}
    \caption{\textbf{The minimum, 10th percentile, and maximum success of workers across the training steps.}
    \textbf{(a)} Averaging their weight shard.
    \textbf{(b)} Performing the reduce operation of the assigned weight partition.}
    \label{fig:pcnt}
\end{figure}

In \Cref{fig:average_pcnt} we see the minimum, 10th percentile, and maximum of the success rate across workers over the iterations.
Looking at the minimum, there is often at least one worker that fails completely to average its weight shard with the others.
Similarly, in \Cref{fig:reducer_pcnt} we observe the minimum, 10th percentile, and maximum success rate of a worker performing the reduce operation.
Failing to perform the reduce has worse consequences: the assigned weight partition then remains unaveraged in SPARTA for all workers, allowing the replicas to drift further apart; we examine this weight divergence in detail in \Cref{ssec:sparta_fault_tolerence}. The frequency of complete reducer failure is low compared to how often a worker fails to average its shard.

\begin{figure}[t]
    \centering
    \begin{subfigure}[t]{0.48\linewidth}
        \centering
        \includegraphics[width=\linewidth]{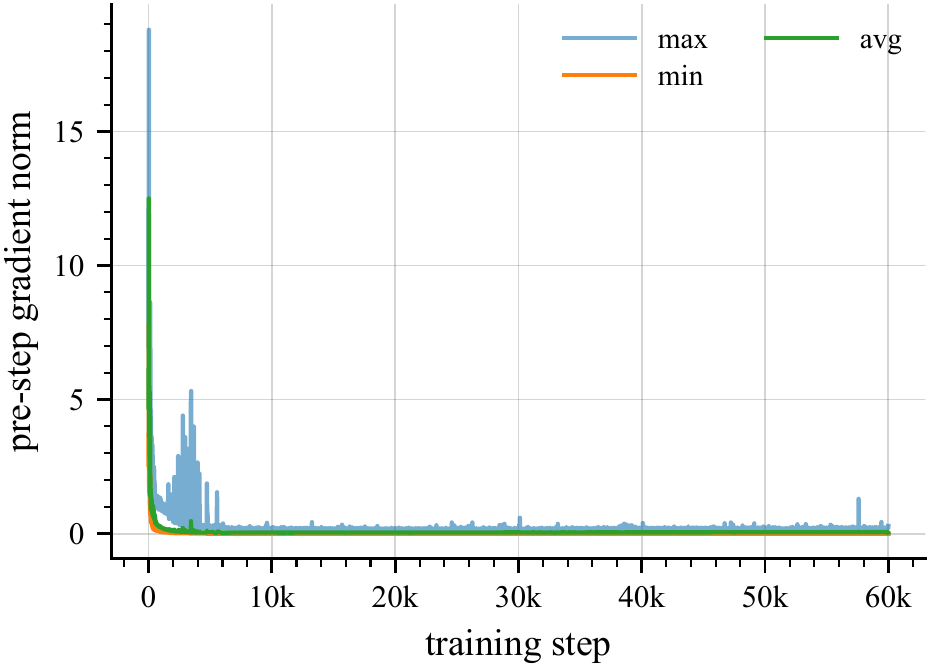}
        \caption{Across all workers.}
        \label{fig:pre_step_grad}
    \end{subfigure}
    \hfill
    \begin{subfigure}[t]{0.48\linewidth}
        \centering
        \includegraphics[width=\linewidth]{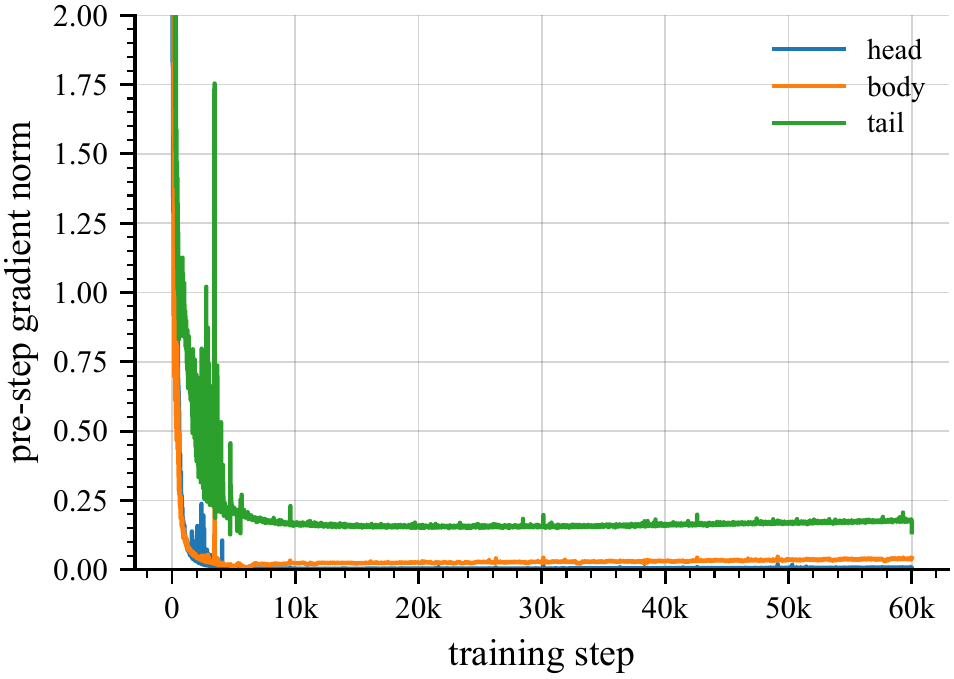}
        \caption{Segmented by stage type.}
        \label{fig:pre_step_grad_per_stage}
    \end{subfigure}
    \caption{\textbf{The gradient norm against the training steps.}
    \textbf{(a)} Across all workers (minimum, average, and maximum).
    \textbf{(b)} The average gradient norm segmented by stage type: \textit{head, body, tail}.}
    \label{fig:pre_step_grad_combined}
\end{figure}

We next observe the gradient norms of the run.
In \Cref{fig:pre_step_grad} we see the minimum, average, and maximum of the norms across all workers.
\Cref{fig:pre_step_grad_per_stage} breaks the average gradient norm down by stage type: \textit{head}, \textit{body}, and \textit{tail}.
This per-stage view reveals a clear difference between the tail's gradient magnitude and that of the head and body stages.
We recall in \Cref{ssec:stage_clipping} that, due to the communication overhead, the Agora system does not perform global gradient clipping but instead applies a per-stage clipping value.

\subsection{Sync Phases}
\label{ssec:results_sync_phases}

Before a contributor worker can fully join the run, it passes through two \emph{sync} phases: \emph{weight sync} followed by \emph{optimizer sync} (\Cref{worker_section}).
These are one-time warm-up phases for a joining worker that bring the stale state it downloaded up to date with the other active workers in its stage.
During both phases the worker receives the averaged weights without yet contributing to them, warming up its weights.
During the second phase it additionally processes batches, warming up its optimizer state.
In this section we analyze whether these phases successfully prevent the negative effects that a stale-weight replica has on convergence, namely gradient and loss spikes.

\begin{figure}[t]
    \centering
    \includegraphics[width=0.9\linewidth]{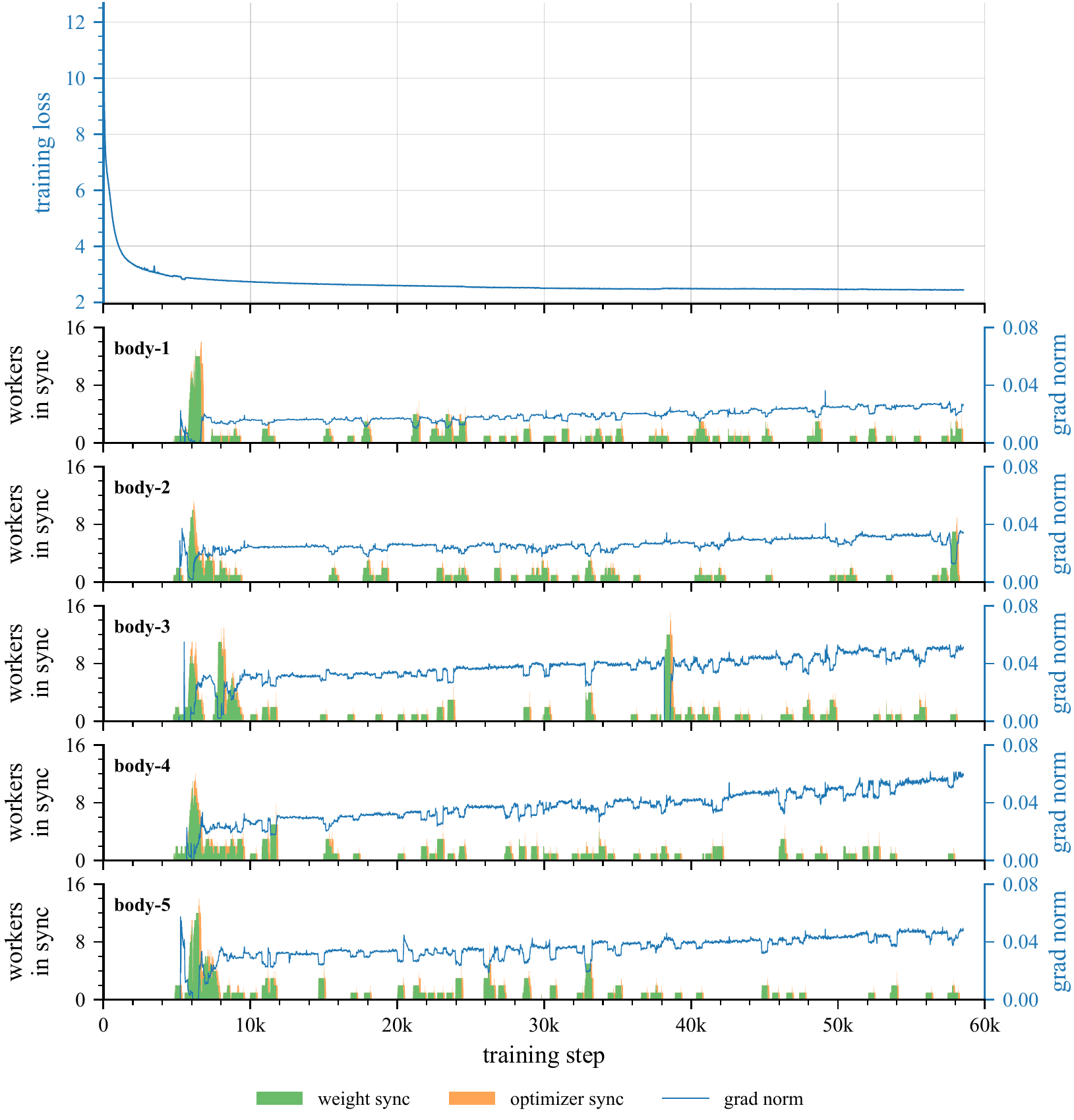}
    \caption{\textbf{Sync phases vs.\ loss and gradient norm.} Global training loss (top) and, per body stage, the number of workers in weight sync and optimizer sync, together with the stage's pre-step gradient norm. Sync activity produces no loss or gradient spikes as workers join.}
    \label{fig:fig_ghost_mode}
\end{figure}

\paragraph{Convergence stability.}
We set the \emph{weight sync} duration to 400 steps and the \emph{optimizer sync} duration to 100 steps.
The former is the number of steps needed for SPARTA to average the full weight tensor: it averages 5\% of the weights per round, one round every 20 steps, so a complete pass takes 20 rounds (400 steps).
We set the \emph{optimizer sync} duration to 100 steps so that a joining worker warms up its AdamW second moment ($\beta_2 = 0.95$, an EMA of time constant $1/(1-\beta_2)=20$ steps) over five time constants, decaying the stale initialization to $\beta_2^{100}\approx0.6\%$.
In \Cref{fig:fig_ghost_mode} we show the loss curve for the whole run, together with the number of workers in a sync phase and the pre-step gradient norm (per body stage).
In practice only the body stages engage the sync phases; the head and tail are operated by stable Pluralis workers, which are fast enough to fetch the fresh state from their peers.
Sync activity was heaviest at start-up and sporadic thereafter.
We observed no significant spikes in loss or gradients after workers exited the sync phases and joined the run, and therefore conclude that the sync-phase mechanism effectively mitigates the stale-weight problem.

\paragraph{Purging slow workers.}
The sync phases play an additional role in maintaining system stability.
Although workers in these phases do not yet contribute to the averaged result, they still participate in the all-reduce rounds, which lets us measure their success rate.
If a worker fails to complete 2 of any 3 consecutive rounds, we remove it from the run before it starts contributing.

\begin{figure}[t]
    \centering
    \includegraphics[width=0.6\linewidth]{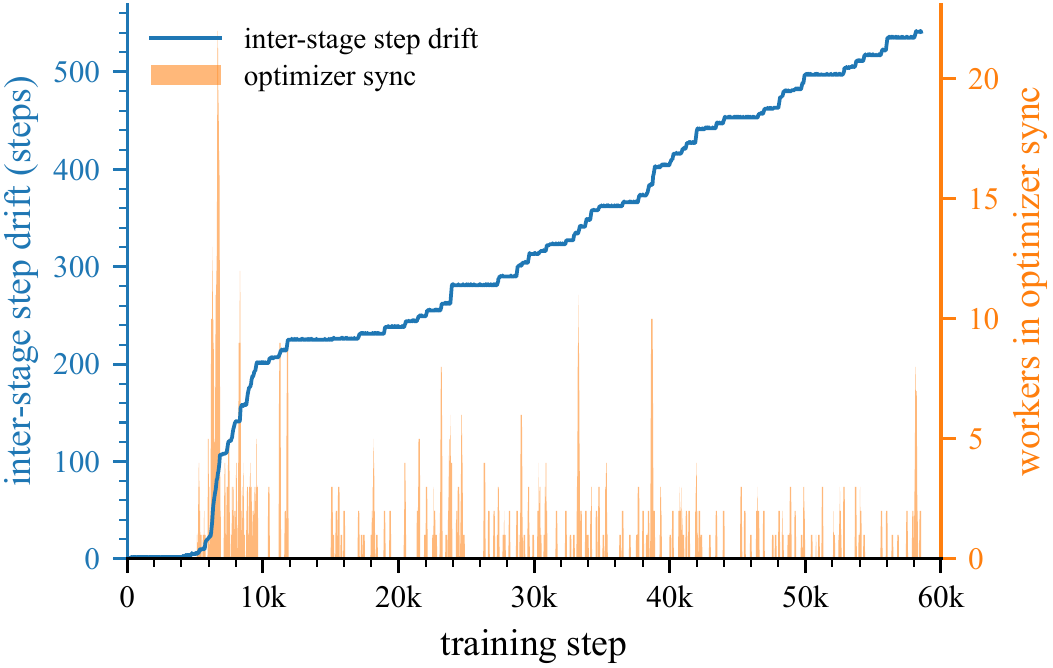}
    \caption{\textbf{Inter-stage step drift under optimizer sync.} The inter-stage step
    drift (max$-$min of the per-stage median step) climbs with each burst of
    optimizer-sync activity (workers in optimizer sync summed over stages), accumulating to
    540 steps (0.9\%) over the run.}
    \label{fig:fig_step_drift}
\end{figure}

\paragraph{Step drift under optimizer sync phase.}
During the \emph{optimizer sync} phase, a joining worker already receives and processes real batches from the trainers, but does not report them to its stage's progress (the stage's collective count of samples processed toward the current global batch).
A stage that hosts a worker in optimizer sync therefore has to process \emph{more} batches than the others to reach the global batch size, because the syncing worker's batches do not count.
Its reported step (incremented each time the global batch size is reached by the stage) consequently advances more slowly than that of stages without a syncing worker, and the reported step across stages drifts apart.
\Cref{fig:fig_step_drift} shows exactly this: each optimizer-sync burst nudges the inter-stage step drift (the gap between the most- and least-advanced stages) upward.
Because the head and tail run only stable Pluralis workers and never sync, the body stages fall progressively behind, so the drift accumulates rather than recovering.
The effect is small in absolute terms: optimizer sync lasts only 100 steps, and a joining worker is lightly loaded throughout it while the trainers' load balancer adapts to it, so over the whole run the drift reached only 540 steps, about 0.9\% of the 60k steps.
This divergence is a matter of step \emph{accounting} rather than of the data: every stage still processes the same samples at the same throughput, so the optimization itself is unaffected.
The only step-dependent quantity affected is the per-worker learning-rate schedule, but a stage lagging by $\sim$0.9\% simply sits $\sim$0.9\% earlier on its schedule, a negligible offset over a smooth 60k-step decay.

\subsection{SPARTA Fault Tolerance and Divergence}
\label{ssec:sparta_fault_tolerence}

As discussed in \Cref{ssec:async_sparta}, AsyncSPARTA keeps $R$ data-parallel replicas of a stage's parameters in consensus by averaging $p = 5\%$ of the parameters each round, where a round occurs every 20 optimizer steps.
The parameters are partitioned into 20 disjoint index sets that tile the parameter vector; we call these \emph{SPARTA segments}, numbered 0--19.
Rounds cycle through the segments in a fixed order, so every individual parameter is averaged exactly once per 20 rounds (every 400 steps).

The all-reduce step is designed to be fault-tolerant and completes even if some replicas fail to send or receive their complete parameter segment in a given round.
We explored the stability of SPARTA under bandwidth constraints and fault injection in \Cref{testing_framework}; here we examine what actually happened during the Pluralis-8B run.

\subsubsection{All-Reduce Fault Tolerance}

\paragraph{Measuring partial all-reduce failure.}
To gauge the health of all-reduce steps, we measure the fraction of parameters successfully gathered from the replicas for averaging, and the fraction of averaged parameters successfully received back.
We call these \emph{reduce success} and \emph{averaging success}, respectively; they are logged by each node during each SPARTA round.

In our SPARTA implementation, we only reduce to nodes hosted by Pluralis, which are typically more stable than contributor nodes.
However, the slices being averaged are gathered from every replica in the data-parallel group, contributors included.
The two percentages therefore describe quite different aspects of the all-reduce scatter-gather process, with different consequences for failure.

Reduce success is the gather side: the fraction of parameters due to be averaged that the reducer actually receives.
This is less than 100\% when a sender fails to deliver its entire slice in time.
The reducer only averages the parameters it receives, so contributions from flaky senders are effectively lost and excluded from the global average.

Averaging success is the scatter side: the fraction of the averaged slices that nodes receive back from the reducer.
This is less than 100\% when a receiver fails to receive the averaged slice in time.
If a node fails to receive the averaged slice, it retains its own values for those parameters and effectively holds some stale weights until the next averaging round.

In the results below, we focus on the body-3 and head stages of the model.
The body-3 stage contains 4 layers (15--18), and its results are very similar to those of the other body stages.
It is served by 4 Pluralis nodes and 12 contributor nodes.
The head stage contains 6 layers (1--6) and is served by 8 Pluralis nodes.
As there are significantly fewer failures on Pluralis nodes than on contributor nodes, we use them as a stable baseline for comparison.

\begin{figure}[tbp]
    \centering
    \begin{subfigure}{0.49\linewidth}\centering
        \includegraphics[width=\linewidth]{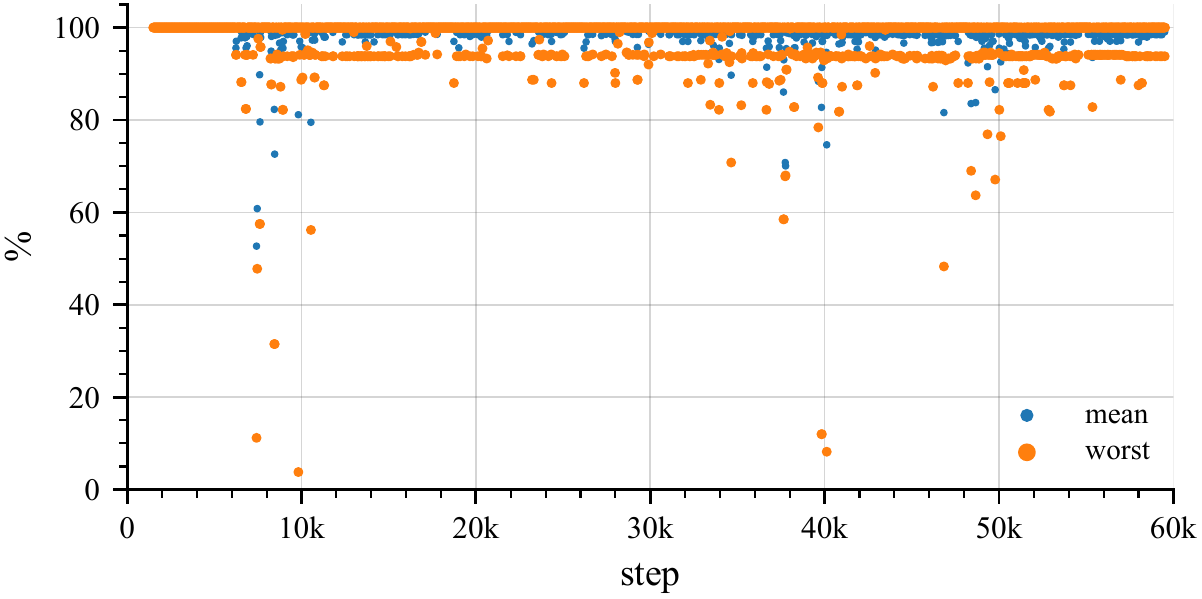}
        \caption{Reduce success (gather side): worst vs.\ mean over the 4 Pluralis reducers, per round.
        The worst value usually sits near 100\%; the band at $\approx$94\% corresponds to exactly 1 of the 16 replicas failing to deliver its slice in time.}
        \label{fig:sparta_reducer}
    \end{subfigure}\hfill
    \begin{subfigure}{0.49\linewidth}\centering
        \includegraphics[width=\linewidth]{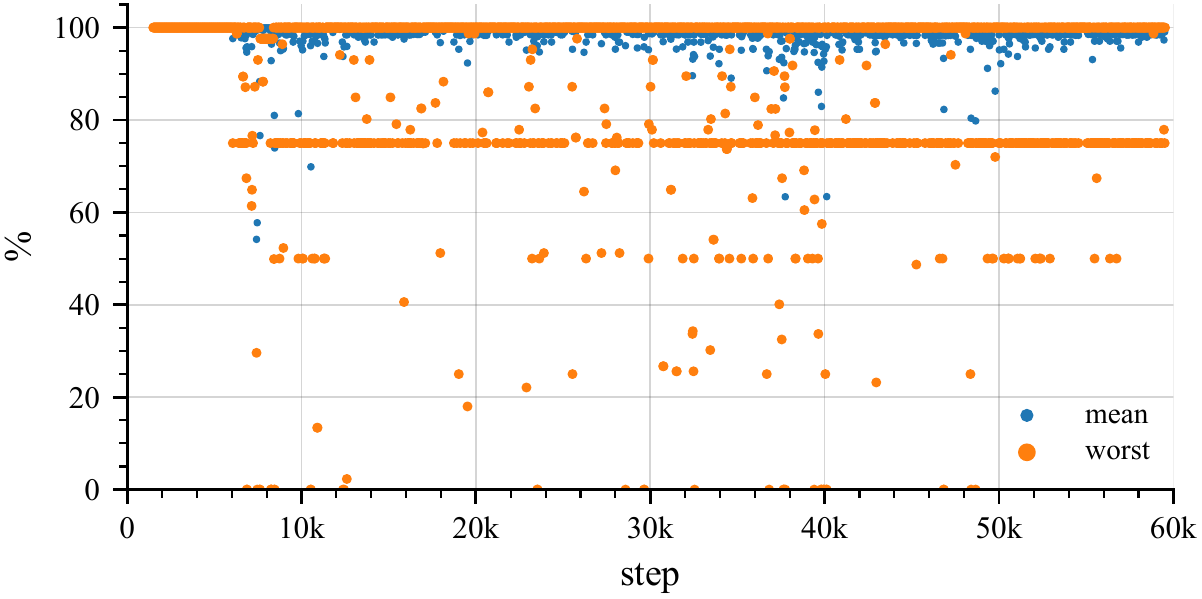}
        \caption{Averaging success (scatter side) over \emph{all} body-3 nodes (contributors $+$ Pluralis), worst vs.\ mean per round.
        The band at exactly 75\% corresponds to receiving the averaged slices from only 3 of the 4 reducers.}
        \label{fig:sparta_average_allnodes}
    \end{subfigure}\\[4pt]
    \begin{subfigure}{0.49\linewidth}\centering
        \includegraphics[width=\linewidth]{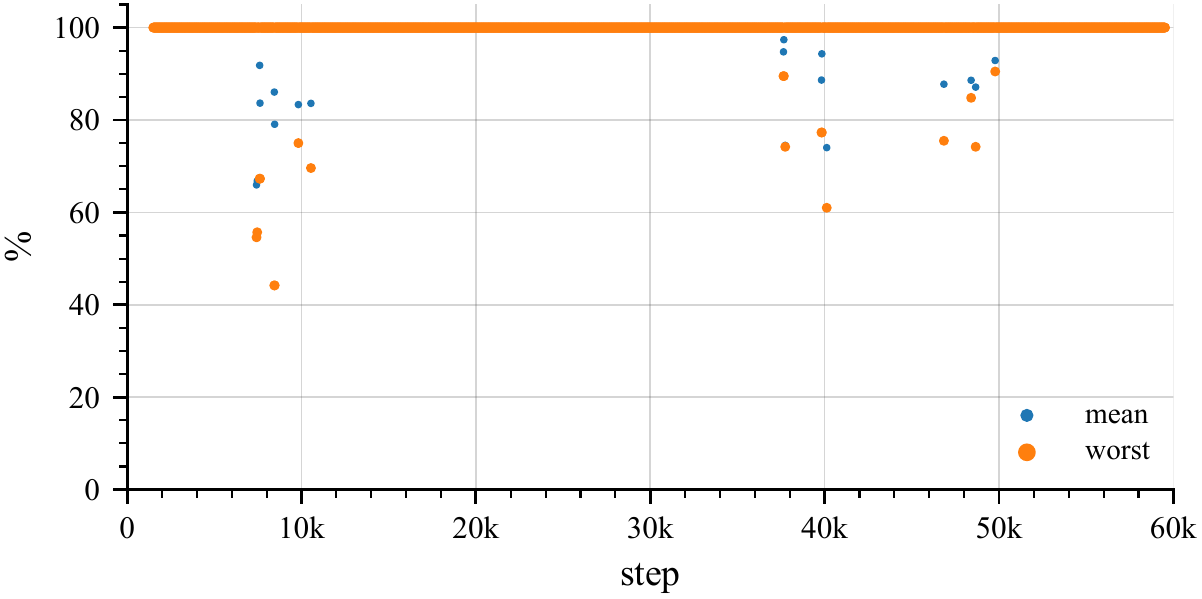}
        \caption{Averaging success over the Pluralis nodes only: worst and mean are almost always 100\%, apart from the dips around steps 10k, 40k, and 50k.}
        \label{fig:sparta_average}
    \end{subfigure}
    \caption{\textbf{All-reduce health for the body-3 stage across the run.}
    Each point is one SPARTA round, plotted at its optimizer step; each panel shows the mean and the worst value over the participating nodes in that round.
    The brief degradations around steps 10k, 40k, and 50k are most likely network issues on contributor nodes.}
    \label{fig:sparta_reduce_health}
\end{figure}

\paragraph{Observed all-reduce health.}
\Cref{fig:sparta_reduce_health} reports both metrics for the body-3 stage across the whole run.
The gather side (\Cref{fig:sparta_reducer}) is mostly healthy throughout: the mean of reduce success stays at 99--100\%, so even with flaky senders, the vast majority of parameters reach the reducers in every round.
The scatter side (\Cref{fig:sparta_average_allnodes}) is where contributor failures show up: averaging success over all body-3 nodes is much lower than the gather-side values, most commonly because a node missed the averaged slices from one of the four reducers.
Restricted to the Pluralis nodes (\Cref{fig:sparta_average}), the scatter side is also almost always 100\%.
All three panels degrade briefly around steps 10k, 40k, and 50k, most likely due to network issues on the contributor nodes.

We also measured the all-reduce health of the head stage, which is served by Pluralis nodes only.
There, both reduce success and averaging success stay at almost 100\% for every round, with no drops.

\subsubsection{Inter-Replica Divergence}

\paragraph{Measuring inter-replica divergence.}
The all-reduce health metrics in \Cref{fig:sparta_reduce_health} show how stable the SPARTA averaging was at different times during the run.
Here we measure how much the averaged parameters actually diverged across replicas, and whether the all-reduce failures had any effect on the model.

To measure inter-replica divergence, we examine parameter snapshots taken every 500 steps during training on Pluralis nodes.
For each trainable parameter we take the population variance of its value across the $R$ replicas of a stage, and average this within each of the 20 SPARTA segments to obtain that segment's \emph{inter-replica variance} for the snapshots at the given step.
Note that we only have snapshots from Pluralis nodes, and it is likely that contributor nodes would have higher divergence, as they have higher failure rates during SPARTA averaging.

\begin{figure}[tbp]
    \centering
    \begin{subfigure}{0.49\linewidth}\centering
        \includegraphics[width=\linewidth]{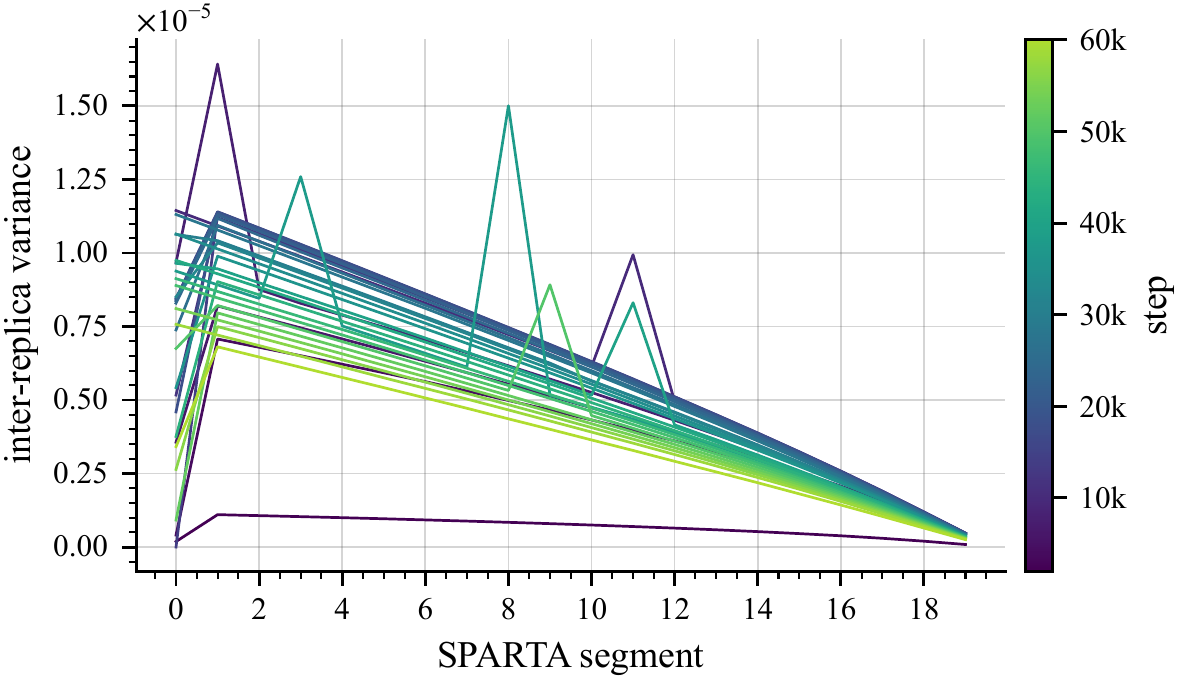}
        \caption{body-3 (Pluralis $+$ contributor nodes).
        The spikes coincide with the partial all-reduce failures.}
        \label{fig:sparta_sawtooth_body}
    \end{subfigure}\hfill
    \begin{subfigure}{0.49\linewidth}\centering
        \includegraphics[width=\linewidth]{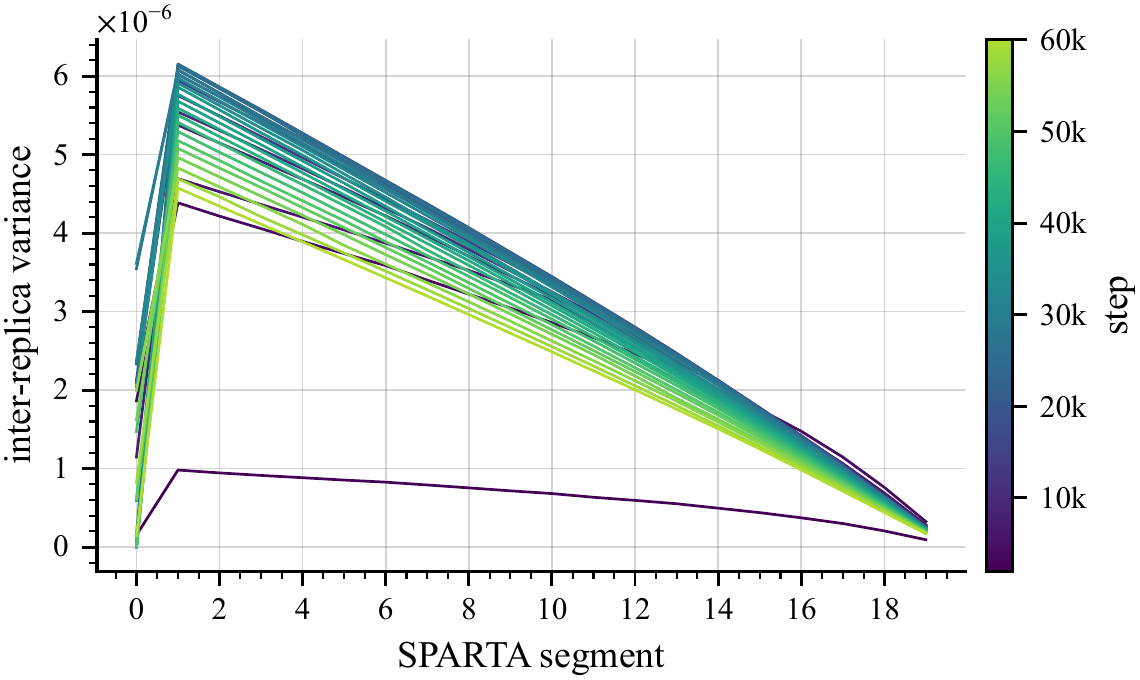}
        \caption{head (Pluralis nodes): lower variance overall and no spikes, as the all-reduce rarely fails between Pluralis nodes.}
        \label{fig:sparta_sawtooth_head}
    \end{subfigure}
    \caption{\textbf{Inter-replica variance vs.\ SPARTA segment, for (a) body-3 and (b) head.}
    Each curve is one parameter snapshot, colored by training step.
    Curves are drawn every 2000 steps so that the 500-step snapshot cadence aligns with the 400-step averaging cycle: in every curve, segment 19 was averaged 20 steps before the snapshot, segment 18 was averaged 40 steps before it, and so on.
    Variance therefore ramps up linearly with the time since a segment was last averaged.
    Segment 0 is averaged at roughly the time the snapshot is taken, so it can sit anywhere between just-averaged (near zero) and just-before-averaging (near the maximum).}
    \label{fig:sparta_sawtooth}
\end{figure}

\begin{figure}[tbp]
    \centering
    \includegraphics[width=0.8\linewidth]{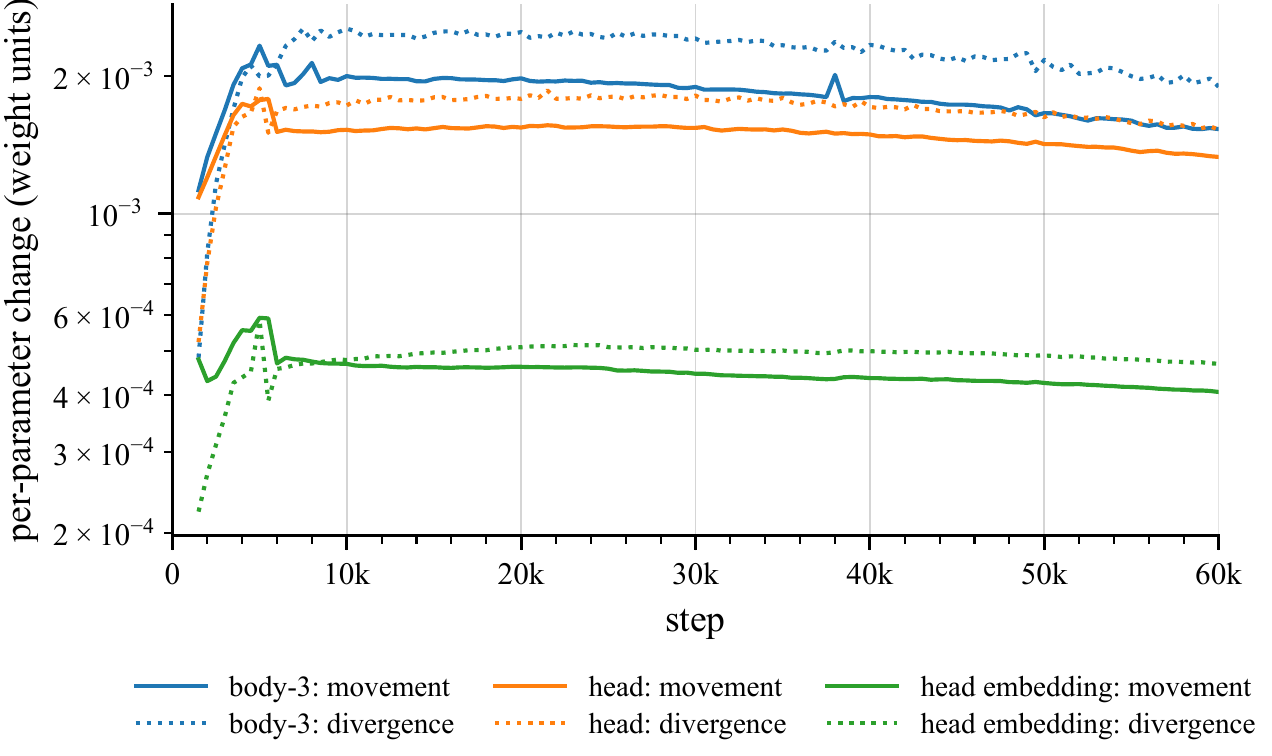}
    \caption{\textbf{Between-snapshot movement of the consensus mean vs.\ per-parameter inter-replica divergence.} Solid curves show movement and dotted curves divergence, for the body-3 and head stages, with the head token embedding shown separately.
    Movement and divergence track each other closely for every group, with divergence sitting slightly above movement throughout the run.}
    \label{fig:sparta_movement_divergence}
\end{figure}

\Cref{fig:sparta_sawtooth} plots the inter-replica variance of each SPARTA segment, for snapshots taken throughout the run.
In the body-3 stage (\Cref{fig:sparta_sawtooth_body}), variance rises linearly with the time since a segment was last averaged, with occasional spikes; we examine both effects in more detail below.
The head stage (\Cref{fig:sparta_sawtooth_head}) shows the same linear relationship with no spikes, at a noticeably lower level: the head-stage replicas drift apart less between rounds (see \Cref{fig:sparta_movement_divergence}).

\begin{figure}[tbp]
    \centering
    \begin{subfigure}{0.49\linewidth}\centering
        \includegraphics[width=\linewidth]{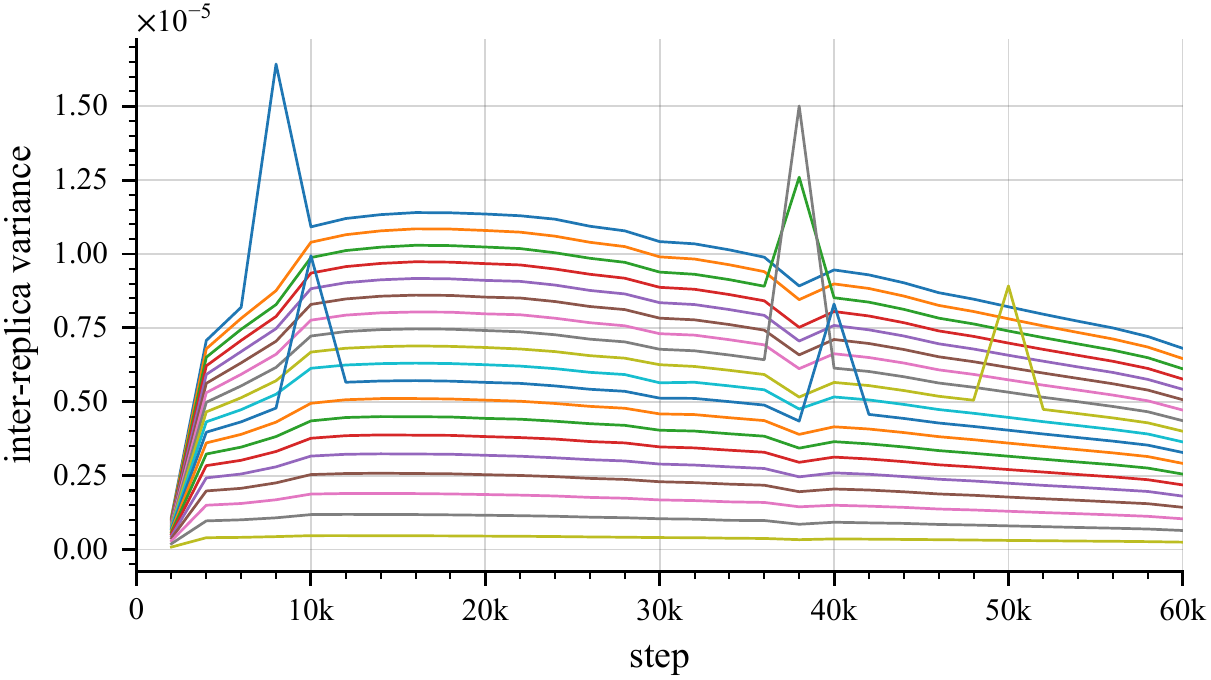}
        \caption{body-3: the upward spikes coincide with the partial all-reduce failures of \Cref{fig:sparta_reduce_health}.}
        \label{fig:sparta_vs_step_body}
    \end{subfigure}\hfill
    \begin{subfigure}{0.49\linewidth}\centering
        \includegraphics[width=\linewidth]{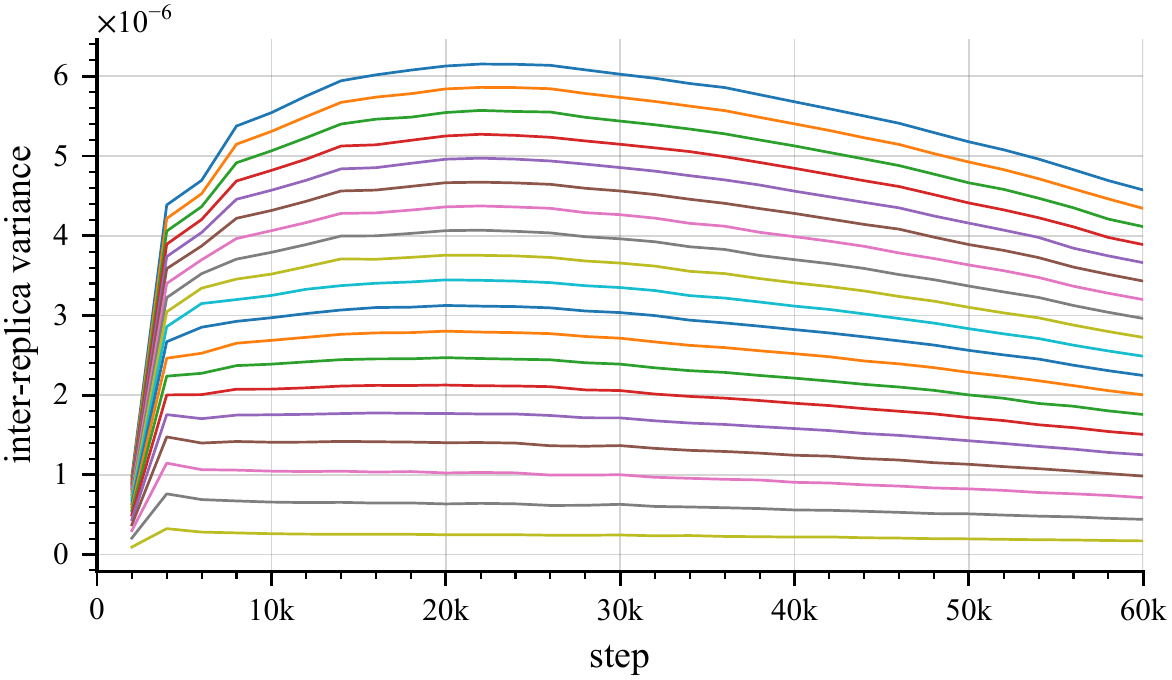}
        \caption{head: no spikes, as there are no all-reduce failures.}
        \label{fig:sparta_vs_step_head}
    \end{subfigure}
    \caption{\textbf{Inter-replica variance vs.\ optimizer step, one line per SPARTA segment.}
    Segment 19, always the most recently averaged before each snapshot, has the lowest variance; the other segments stack above it in order of averaging recency.
    Segment 0 is omitted because its variance is noisy due to snapshot timing (see \Cref{fig:sparta_sawtooth}).}
    \label{fig:sparta_vs_step}
\end{figure}

\paragraph{Divergence vs.\ step.}
\Cref{fig:sparta_vs_step} flips the axes of \Cref{fig:sparta_sawtooth}, plotting variance against step for each segment instead of variance against segment for each step.
This view shows the same linear relationship between a segment's averaging recency and its inter-replica variance, showing a clear separation between the segment curves.

The overall level of divergence also drifts slowly over the run: variance is significantly lower during the warm-up phase (around step 2000), peaks in the first third of the run, and then declines gradually, broadly consistent with the learning-rate schedule.

\Cref{fig:sparta_vs_step_body} shows spikes in variance around steps 10k, 40k, and 50k, which correspond exactly to the all-reduce failures seen in \Cref{fig:sparta_reduce_health}.
As noted, these divergence plots include only Pluralis nodes; the contributor nodes are likely to show more severe divergence spikes, as they are more affected by all-reduce failures.
These spikes are transient and are erased by the next successful averaging of the affected segment.
Notably, they have no measurable effect on the model's training loss.

\paragraph{Movement vs.\ divergence.}
In addition to measuring the variance between snapshots for a single step, we also measure the movement of the consensus mean between snapshots.
\Cref{fig:sparta_movement_divergence} highlights several features of the SPARTA averaging process.
We see that the movement of the consensus mean is comparable to the inter-replica divergence for each stage.
In fact, the inter-replica divergence is typically slightly above the consensus movement.
This indicates that more frequent averaging (e.g., every 10 steps instead of 20) could be beneficial, as the replicas are diverging faster than the consensus is moving.

We also see that some model stages are more stable than others.
The head stage has lower divergence and movement than the body stage, with the head token embedding being significantly more stable than the other parameters in the head stage.

\subsection{Communication Efficiency}
\begin{figure}[t]
    \centering
    \begin{subfigure}{0.48\linewidth}
        \centering
        \includegraphics[width=\linewidth]{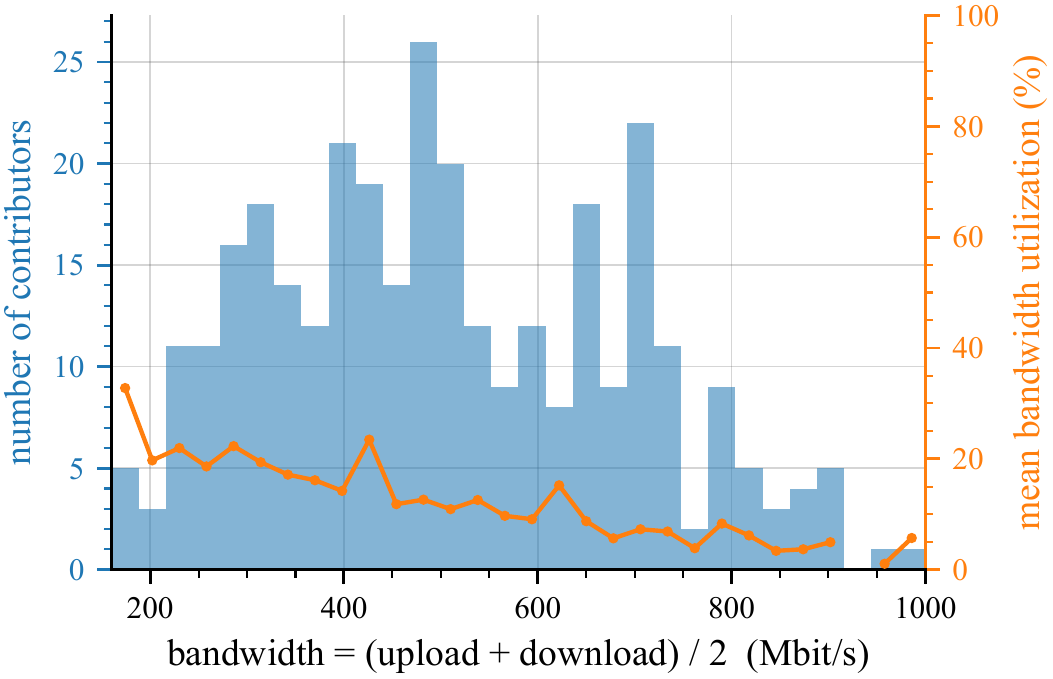}
        \caption{Contributor nodes: bandwidth distribution with per-bin utilization overlaid.}
        \label{fig:comm_bw_contributor}
    \end{subfigure}
    \hfill
    \begin{subfigure}{0.48\linewidth}
        \centering
        \includegraphics[width=\linewidth]{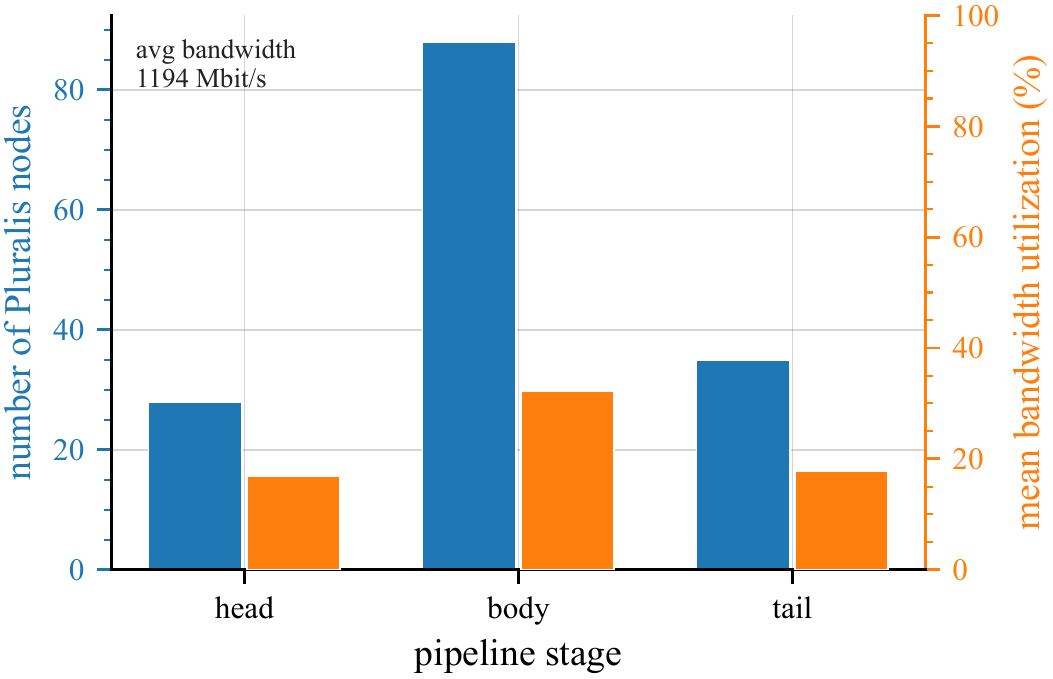}
        \caption{Pluralis nodes: node count against utilization across the stage families.}
        \label{fig:comm_bw_pluralis}
    \end{subfigure}
    \caption{\textbf{Bandwidth distribution and utilization.}}
    \label{fig:comm_bandwidth_stats}
\end{figure}

The communication efficiency of the Pluralis-8B run is highlighted by the ability of the Agora system to multiplex (\Cref{worker_section}) a node's forward and backward communication channel with the all-reduce channel, as the latter runs asynchronously.
In \Cref{fig:comm_bw_contributor} we show the bandwidth distribution for the contributor nodes, with the bandwidth utilization overlaid on each bin.
The trend is that the higher the available bandwidth, the less of it is utilized.
In \Cref{fig:comm_bw_pluralis} we show the Pluralis-hosted nodes, whose bandwidth is similar across the fleet ($\sim$1,194\,Mbit/s) because they are hosted in a datacenter; the plot shows the node count across stages with a bandwidth-utilization summary for each stage type.
Bodies use more of the available bandwidth because they send data in both directions, activations forward and gradients backward, whereas the head sends only forward and the tail only backward.

\begin{figure}[t]
    \centering
    \begin{subfigure}{0.48\linewidth}
        \centering
        \includegraphics[width=\linewidth]{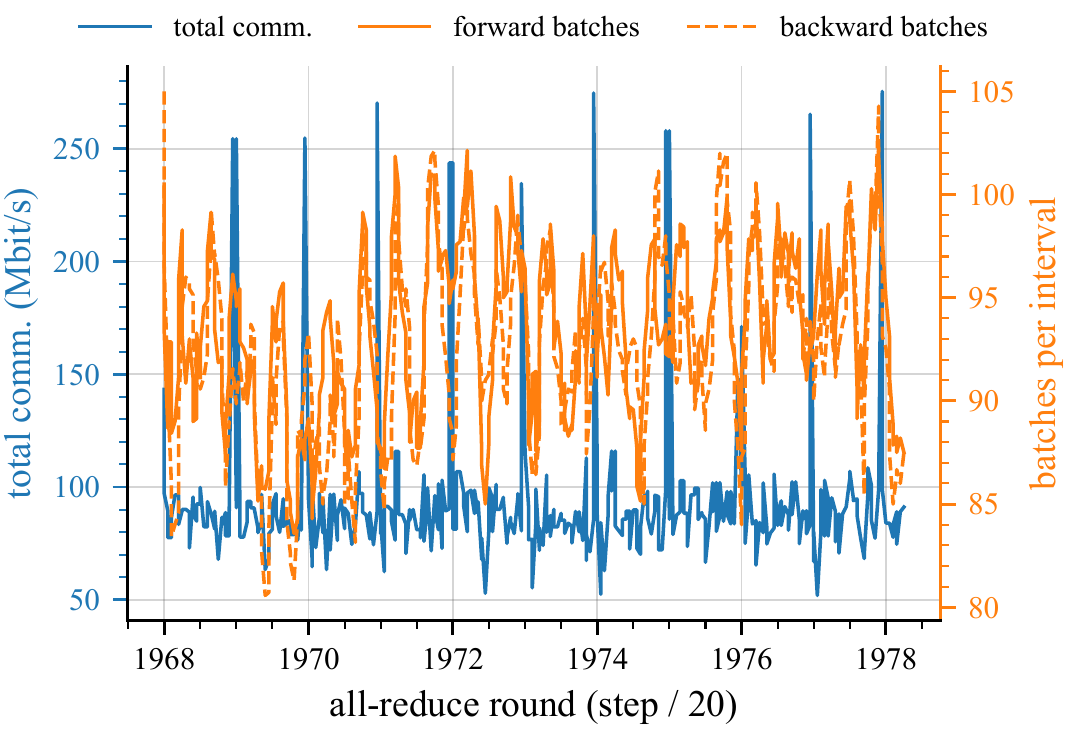}
        \caption{Communication profile of a single body-stage node with its forward and backward batch counts over 12 all-reduce rounds (240 iterations).}
        \label{fig:comm_mux_trace}
    \end{subfigure}
    \hfill
    \begin{subfigure}{0.48\linewidth}
        \centering
        \includegraphics[width=\linewidth]{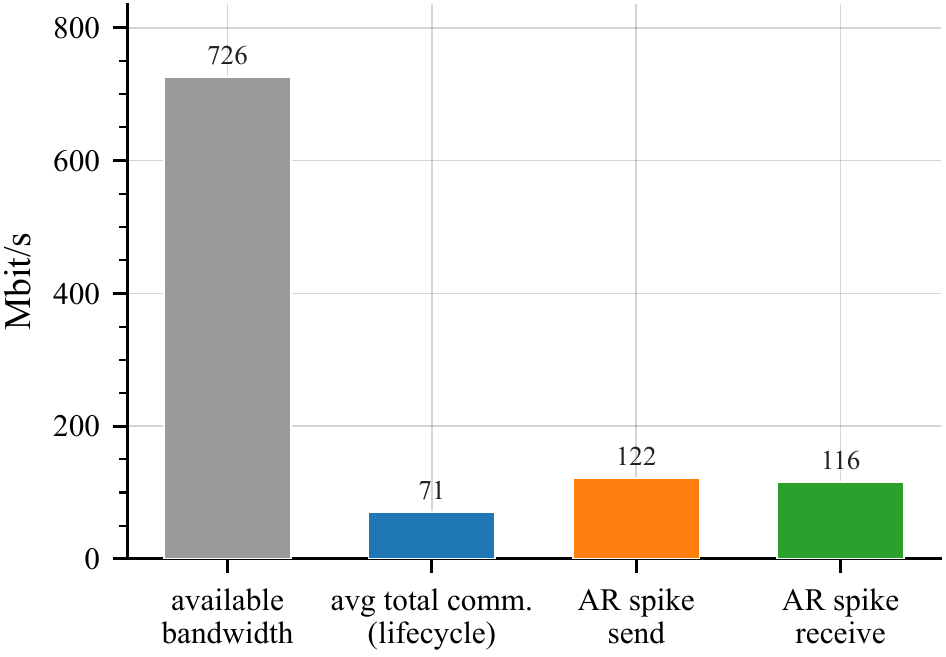}
        \caption{Summary for the same node: available bandwidth, average usage over its lifespan, and average all-reduce communication spikes.}
        \label{fig:comm_mux_summary}
    \end{subfigure}
    \caption{\textbf{Multiplexing computation with the asynchronous all-reduce.} For a single body-stage node.}
    \label{fig:comm_multiplexing}
\end{figure}

In \Cref{fig:comm_mux_trace} we show a single node in a body stage to illustrate how forward and backward batches are processed while the asynchronous all-reduce runs.
Communication spikes occur at every all-reduce round, while forward and backward processing continues uninterrupted.
In \Cref{fig:comm_mux_summary} we show a summary for the same node over its entire lifespan, with its average communication usage plotted against the averaged all-reduce spikes.
The node's total available bandwidth is much higher than its average usage; the all-reduce spikes use the available bandwidth, but only in short bursts.
This ties back to the low overall channel utilization seen in \Cref{fig:comm_bandwidth_stats}.

\begin{figure}[t]
    \centering
    \begin{subfigure}{0.32\linewidth}
        \centering
        \includegraphics[width=\linewidth]{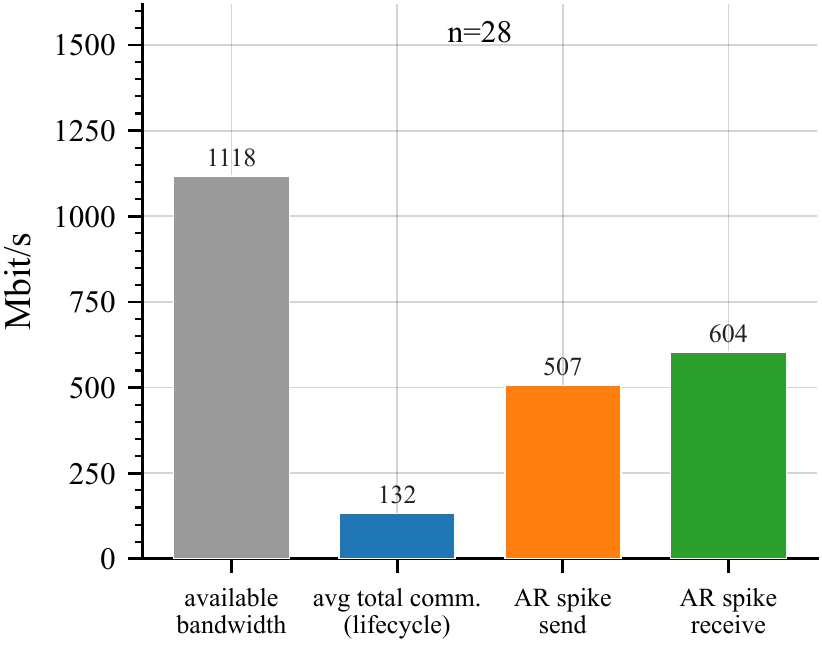}
        \caption{Head.}
        \label{fig:comm_summary_head}
    \end{subfigure}
    \hfill
    \begin{subfigure}{0.32\linewidth}
        \centering
        \includegraphics[width=\linewidth]{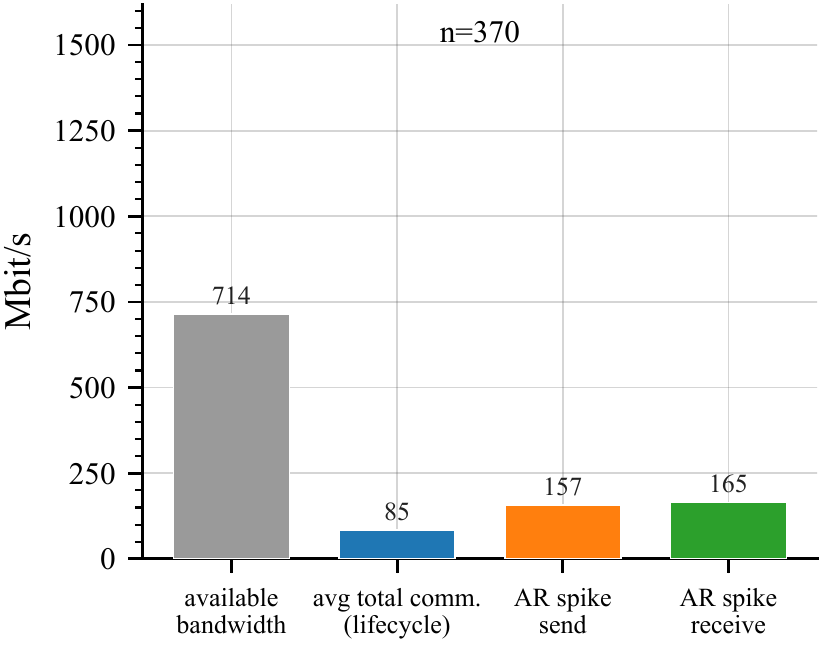}
        \caption{Bodies.}
        \label{fig:comm_summary_body}
    \end{subfigure}
    \hfill
    \begin{subfigure}{0.32\linewidth}
        \centering
        \includegraphics[width=\linewidth]{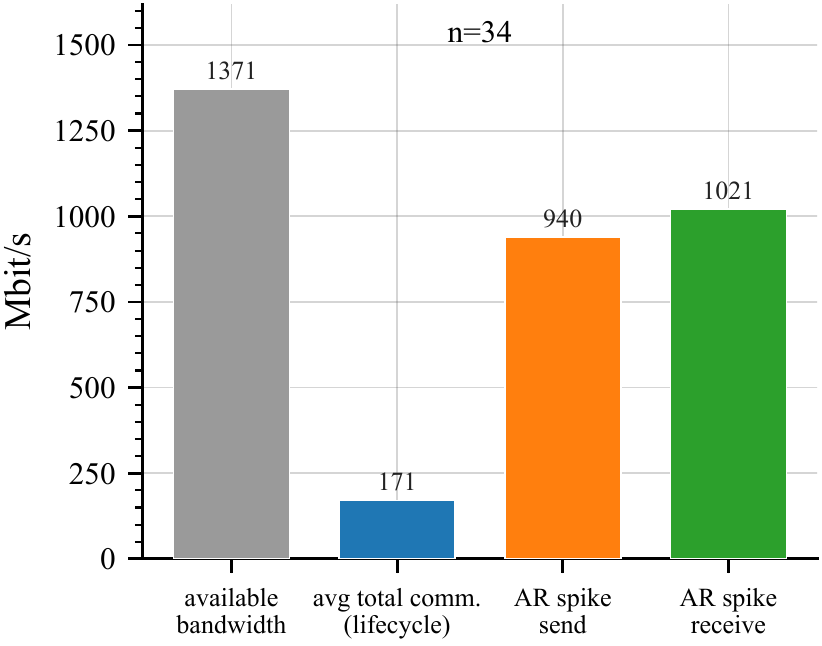}
        \caption{Tail.}
        \label{fig:comm_summary_tail}
    \end{subfigure}
    \caption{\textbf{Bandwidth usage per stage family.} Overall available bandwidth, average bandwidth used, and average upload and download bandwidth during all-reduce.}
    \label{fig:comm_fwd_ar_split}
\end{figure}

\Cref{fig:comm_fwd_ar_split} summarizes the bandwidth usage across the three stage families.
The head and tail stages were hosted only on Pluralis nodes, so their available bandwidth is datacenter grade.
The bodies were a mix of Pluralis and contributor nodes, with a much larger sample, so their measured available bandwidth is lower on average.
The all-reduce spikes differ across stage types because of the volume of weights each stage transfers, with $V_{\mathrm{data}}(\mathrm{tail}) > V_{\mathrm{data}}(\mathrm{head}) > V_{\mathrm{data}}(\mathrm{body})$.
Because the bodies transfer the least weight during all-reduce, their total average communication is the lowest.
Even so, \Cref{fig:comm_bandwidth_stats} shows their utilization is the highest: their available bandwidth is also the lowest, so a smaller absolute volume still amounts to a larger fraction.
The activations and gradients the bodies transfer are larger, but the short, high-volume all-reduce bursts on the head and tail dominate the per-stage averages, so the head and tail have larger average communication overall.

\subsection{Collective Communication and Coordination}
\label{ssec:results_collective_comms}

Agora's collective communication and coordination rest on a few primitives introduced in \Cref{worker_section}: a distributed hash table (DHT) for peer discovery and coordination, all-reduce matchmaking to form averaging groups, and the sparse state all-reduce itself.
\Cref{testing_framework} characterized their stability under controlled bandwidth constraints and fault injection; here we examine how they behaved during the Pluralis-8B run once contributors with heterogeneous, high-latency connections joined.

\subsubsection{DHT Communications}
\begin{figure}[t]
    \centering
    \begin{subfigure}{0.48\linewidth}
        \centering
        \includegraphics[width=\linewidth]{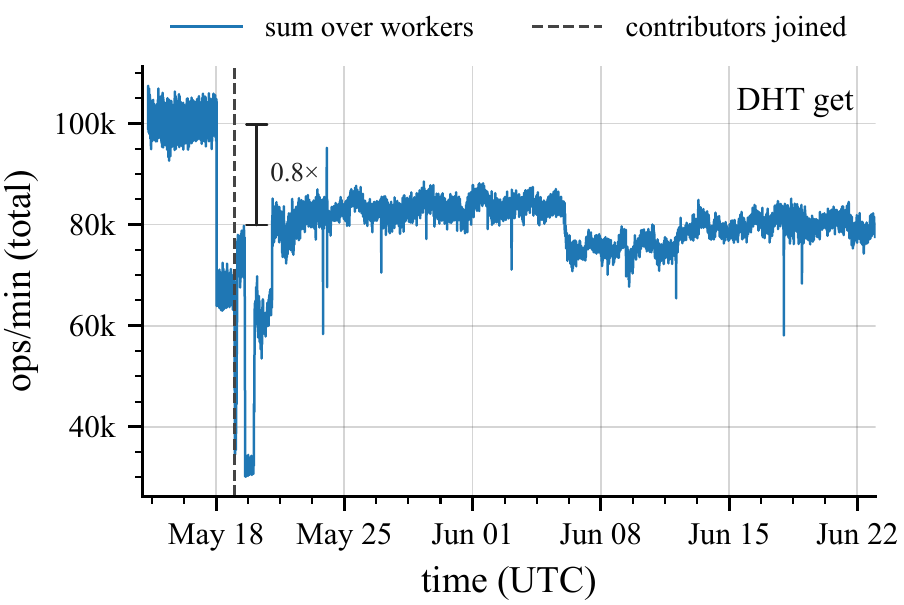}
        \caption{DHT \textsc{Get} operations.}
        \label{fig:dht_throughput_get}
    \end{subfigure}
    \hfill
    \begin{subfigure}{0.48\linewidth}
        \centering
        \includegraphics[width=\linewidth]{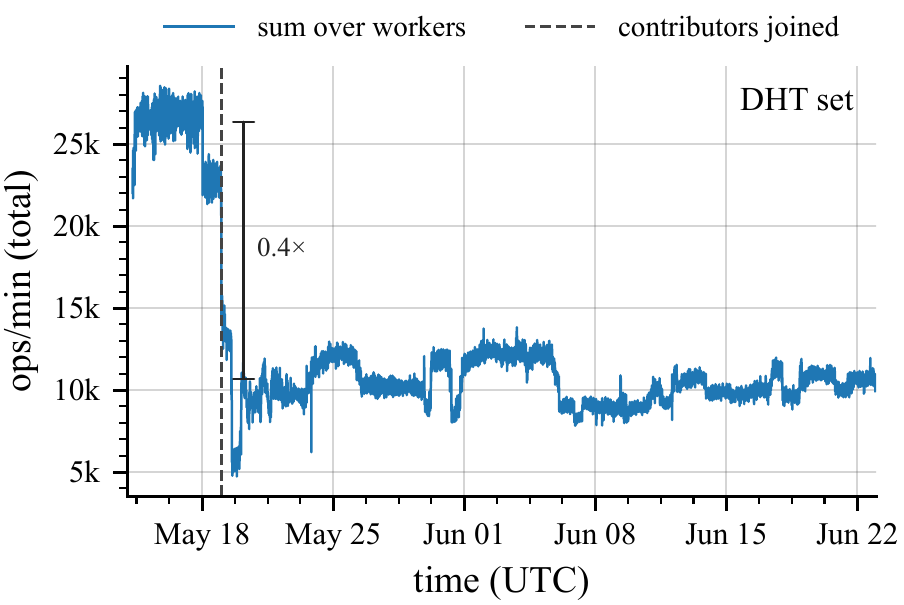}
        \caption{DHT \textsc{Set} operations.}
        \label{fig:dht_throughput_set}
    \end{subfigure}
    \caption{\textbf{DHT throughput during the Pluralis-8B training run, measured from Pluralis nodes.} Total throughput of DHT operations across all stages; the dashed line marks when contributors were allowed to join. Throughput drops noticeably after the join, despite the additional peers. Between 18 May and the join date the Health Monitor was offline, reducing the operation count over that window.}
    \label{fig:dht_throughput}
\end{figure}

\begin{figure}[t]
    \centering
    \begin{subfigure}{0.48\linewidth}
        \centering
        \includegraphics[width=\linewidth]{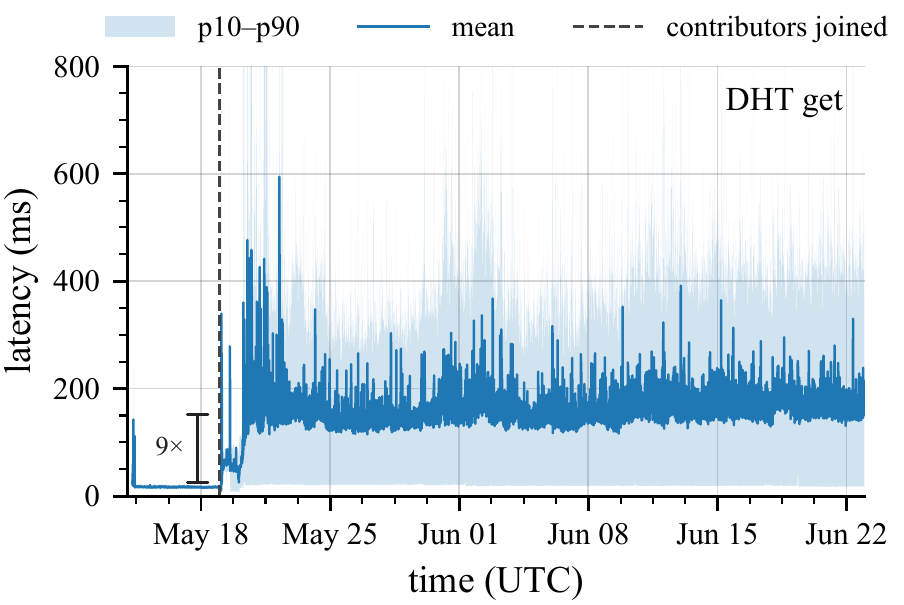}
        \caption{DHT \textsc{Get} operations.}
        \label{fig:dht_latency_get}
    \end{subfigure}
    \hfill
    \begin{subfigure}{0.48\linewidth}
        \centering
        \includegraphics[width=\linewidth]{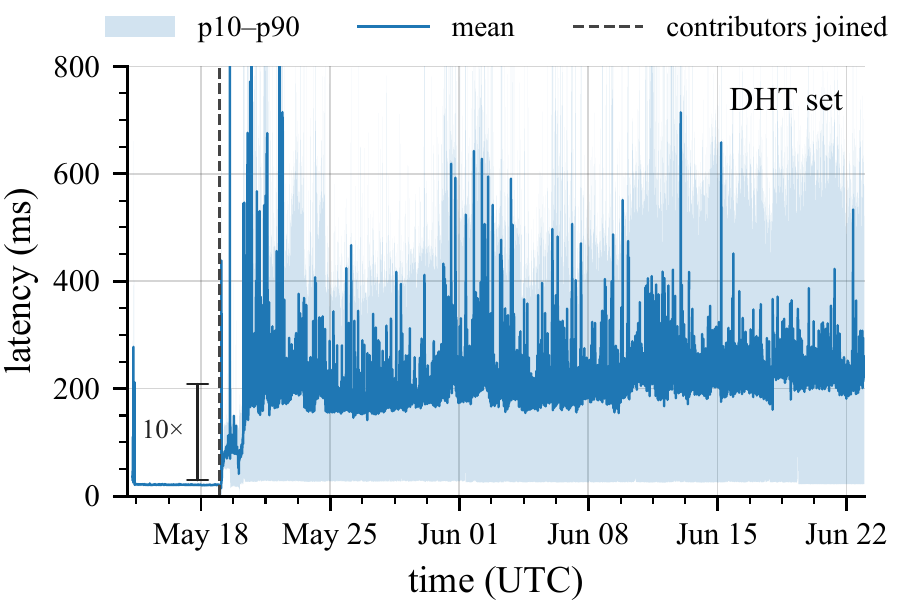}
        \caption{DHT \textsc{Set} operations.}
        \label{fig:dht_latency_set}
    \end{subfigure}
    \caption{\textbf{DHT latency during the Pluralis-8B training run, measured from Pluralis nodes.} Latency of DHT operations across all stages; the dashed line marks when contributors were allowed to join, after which latency rises by roughly an order of magnitude.}
    \label{fig:dht_latency}
\end{figure}

We first consider the DHT, which underlies peer discovery and coordination across the swarm in a decentralized manner.
Under realistic contributor conditions, critical DHT \textsc{Get} and \textsc{Set} operations can become noticeably slower, as peers may now need to route both through and to high-latency contributor nodes, affecting the system as a whole.
This is shown in \Cref{fig:dht_throughput}, where we quantify the effect that contributors have on the system's DHT operations.
We see a $\sim$20\% and $\sim$60\% decline in DHT \textsc{Get} and DHT \textsc{Set} operations per minute, respectively, despite the number of peers doubling.
This is further corroborated by \Cref{fig:dht_latency}: once contributors join, the latency of these operations on Pluralis nodes rises sharply, to roughly 9--10$\times$ its pre-join baseline.
In the worst case, a single operation takes on the order of seconds during this initial joining phase, before latency stabilizes later in the run.

This degradation is especially costly at the scale of the system, where combined DHT operations reach on the order of $10^5$ per minute.
It stems directly from the DHT's decentralized design:
\begin{enumerate*}[label=(\arabic*)]
    \item slow contributor nodes may hold important key-value pairs yet be preempted at any moment; and
    \item even fast nodes may need to route through slow contributor nodes to reach a given key.
\end{enumerate*}

\begin{figure}[t]
    \centering
    \begin{subfigure}{0.48\linewidth}
        \centering
        \includegraphics[width=\linewidth]{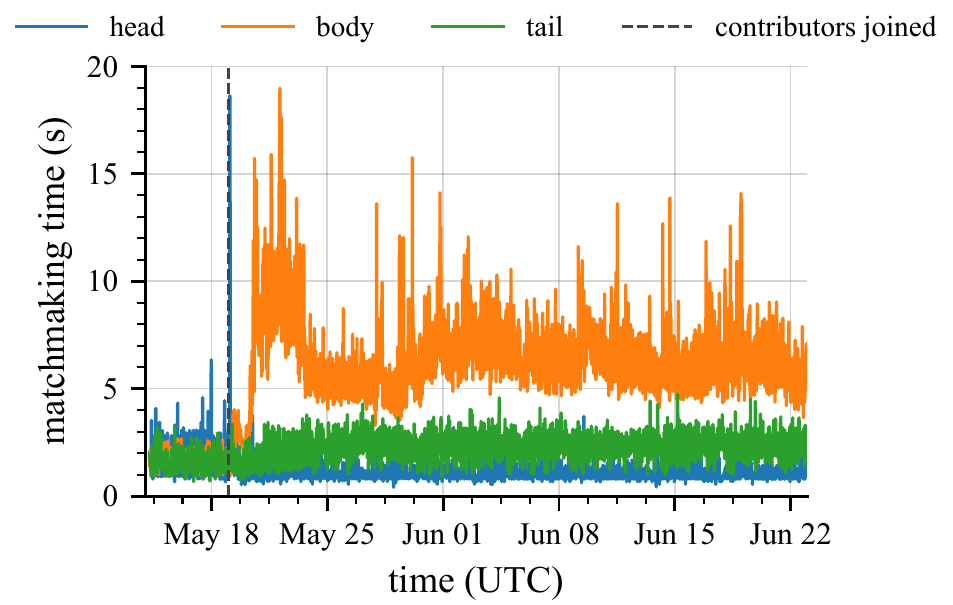}
        \caption{Matchmaking time.}
        \label{fig:mm_time}
    \end{subfigure}
    \hfill
    \begin{subfigure}{0.48\linewidth}
        \centering
        \includegraphics[width=\linewidth]{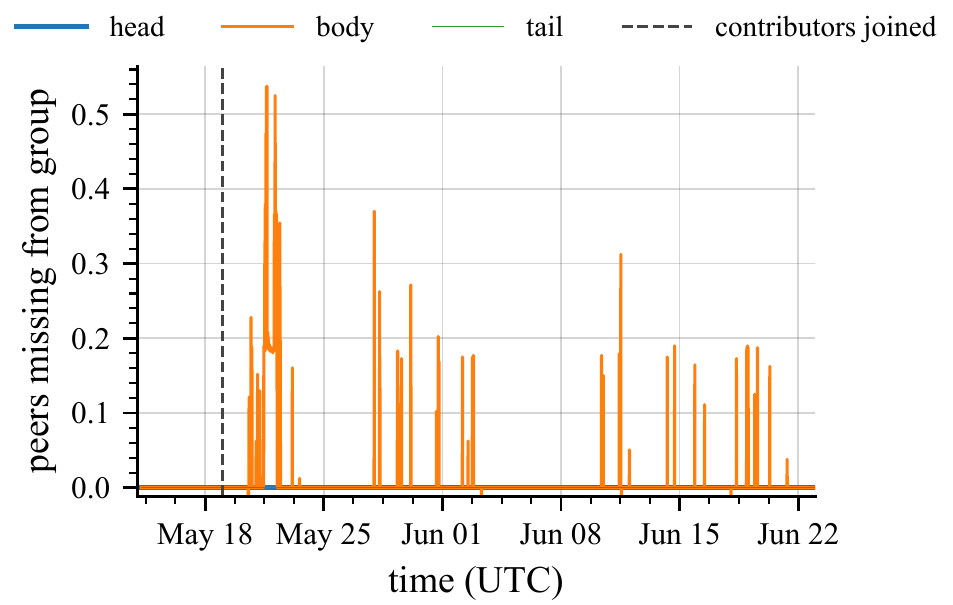}
        \caption{Group-size deficit.}
        \label{fig:groupsize}
    \end{subfigure}
    \caption{\textbf{Matchmaking statistics during the Pluralis-8B training run.} \textbf{(a)} Mean all-reduce matchmaking time per pipeline stage type (head, body, tail), averaged over all stage peers. \textbf{(b)} Group-size deficit, the difference between the expected and formed all-reduce group size, per stage type, averaged over all stage peers. Matchmaking stays stable: times remain within the 40\,s timeout on average and the group deficit is negligible, even for body stages.}
    \label{fig:mm_time_groupsize}
\end{figure}

\begin{figure}
    \centering
    \includegraphics[width=\textwidth]{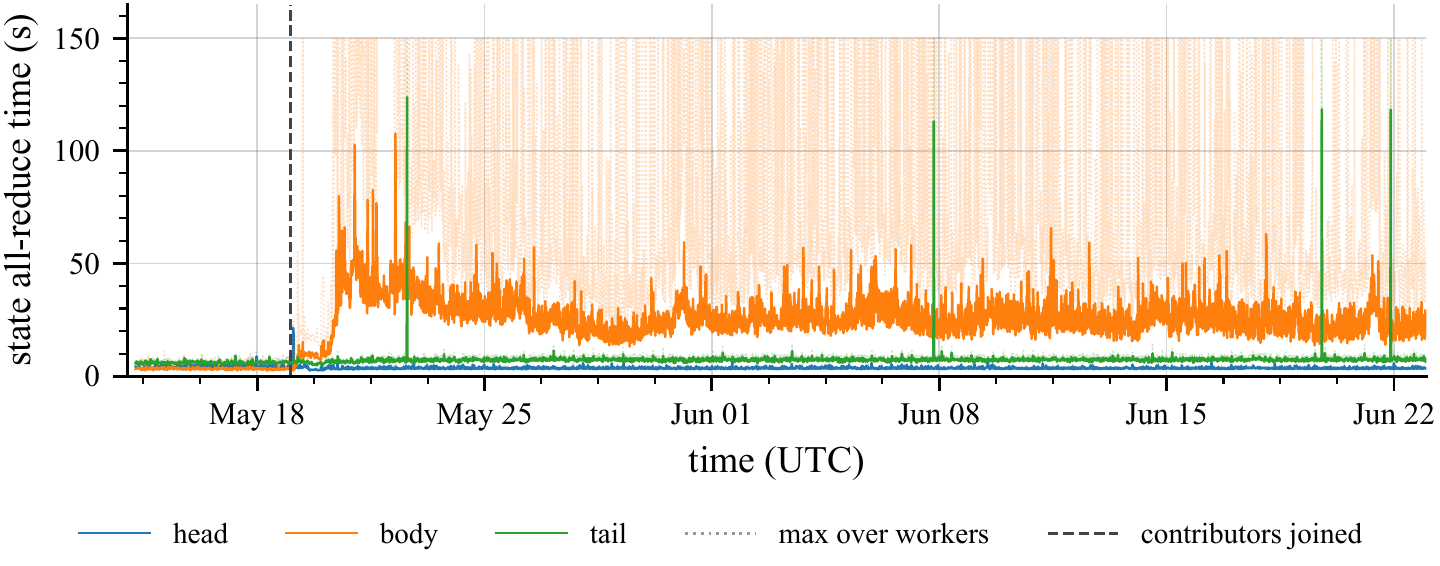}
    \caption{\textbf{State all-reduce time during the Pluralis-8B training run.} Mean and maximum durations of asynchronous state all-reduces per stage type (head, body, tail), over all stage peers. After contributors join, their networking conditions dominate body-stage all-reduce time (up to 8$\times$ higher), and the maximum reflects frequent contributor timeouts.}
    \label{fig:state_ar_time}
\end{figure}

\subsubsection{Matchmaking and All-Reduce Grouping}
One key implication of slow DHT communication concerns the coordination of the system's all-reduce through matchmaking (\Cref{worker_section}).
Matchmaking allows a bounded window for every expected peer's declaration to become visible in the DHT, so a single peer whose write propagates slowly forces all others to wait as well before proceeding.
Any peer that fails to become visible within this window is excluded from the round's all-reduce, leaving its parameters unaveraged, which may cause loss divergence. \textbf{We reiterate that the drawbacks of using a DHT-coordinated matchmaking mechanism are intentional trade-offs that we accept in pursuit of decentralization as per \Cref{training_roles}.}

However, for the Pluralis-8B run, we found the matchmaking mechanism to be extraordinarily stable, as shown in \Cref{fig:mm_time}.
We set the matchmaking timeout to 40 seconds, but in practice the average matchmaking time stays below 20 seconds, even for body stages, where contributors induce a noticeable increase in matchmaking time.
This increase is larger than the one produced by growing the tail stage from 8 to 16 workers, indicating that contributors raise matchmaking time more than enlarging a stage's group does.
Timeouts do still occur, but their effect is limited: an excluded peer only shrinks the formed all-reduce group below its expected size.
\Cref{fig:groupsize} shows this shortfall is small, with the head and tail stages never missing a peer and the body stages missing only a few.
Interestingly, we also see cases where the group size reported by some peers is larger than the expected group size, a result of coordination and synchronization issues. Indeed, this is expected in a Kademlia DHT-coordinated system: continuous contributor churn and multi-hop propagation latency mean membership views may briefly differ within long matchmaking windows, changing the expected group size during matchmaking.
\begin{figure}[tbp]
    \centering
    \includegraphics[width=\textwidth]{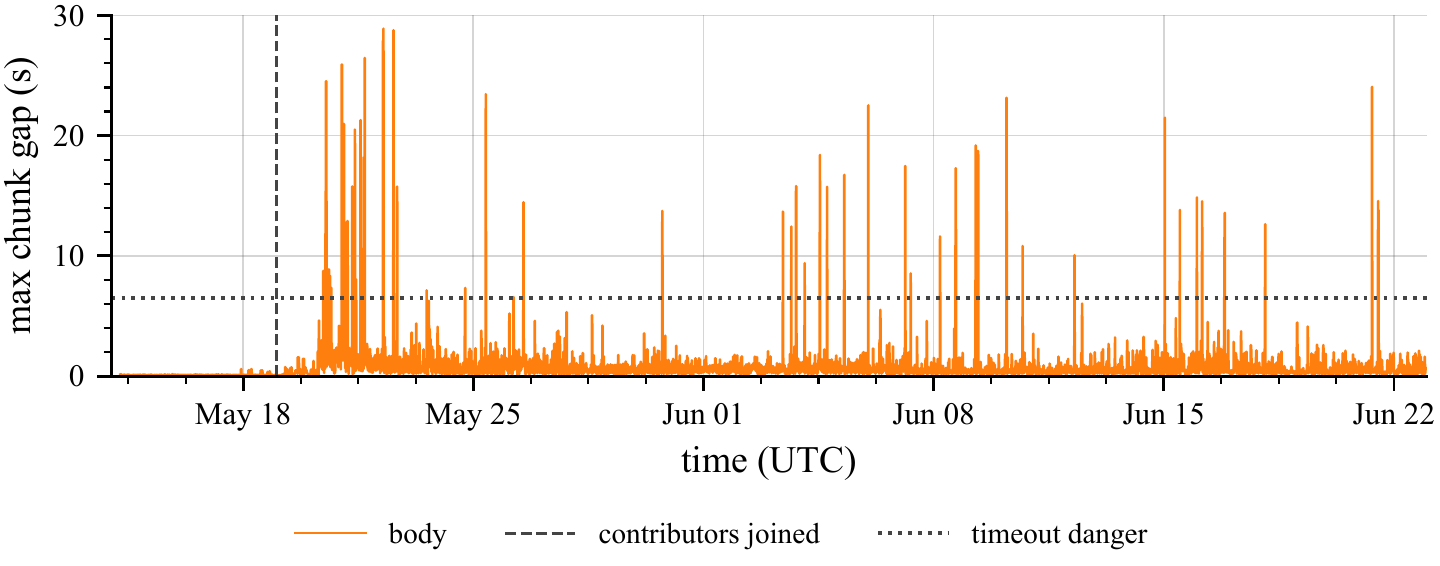}
    \caption{\textbf{Maximum time between all-reduce chunks.} Maximum time between chunks received by reducers, taken over senders, chunks, and reducers. The ``timeout danger'' line (6.5 seconds) marks the per-chunk delay beyond which an all-reduce risks timing out. It is rarely crossed, so slow contributor upload alone does not explain the observed timeouts, which likely stem from transient stream failures, peer churn, network jitter, and delays in receiving the all-reduced parameters.}
    \label{fig:state_ar_chunk_gap_max}
\end{figure}

\subsubsection{State All-Reduce}
The most important collective communication primitive in Agora is all-reduce, specifically Pluralis-8B's asynchronous sparse state all-reduce (\Cref{ssec:async_sparta}).
Under ideal networking conditions, its duration is set by the parameter payload each stage must exchange, which is largest for the tail, then the head, and smallest for the bodies.
Once contributors join, however, their high-latency, low-bandwidth links dominate instead, raising body-stage all-reduce time by up to 8$\times$ (\Cref{fig:state_ar_time}).
Contributor timeouts also become frequent in the body stages, whereas they remain only transient in the Pluralis-only stages, despite a generous 150-second all-reduce timeout.
We chose 150 seconds as 3 steps at the run's $\sim$50-second step time (\Cref{fig:fig_live_tps}), the span over which the asynchronous all-reduce finalizes.
Yet given a contributor's payload of $\sim$1,450\,Mb and the bandwidth observed in \Cref{fig:comm_fwd_ar_split}, the all-reduce should typically complete within a single step.

\Cref{fig:state_ar_chunk_gap_max} shows the maximum time between received chunks, taken across all contributors and reducers, and indicates that the timeouts are not simply due to slow contributor upload.
With a $\sim$1,450\,Mb payload, 16\,Mb chunks, and 4 reducers as in Pluralis-8B, each reducer receives only 23 chunks, so a 150-second timeout tolerates a $\sim$6.5-second gap between chunks.
This gap is rarely exceeded even though timeouts are frequent (\Cref{fig:state_ar_time}), so factors such as transient stream failures, peer churn, network jitter, and averaged-parameter download likely contribute more than slow contributor connections alone.

\subsection{Resource Heterogeneity Across Contributors}
\label{ssec:results_heterog}

Pluralis-8B was trained by a collection of Pluralis-owned nodes and a heterogeneous pool of contributor nodes that differ in GPU model, network latency, and bandwidth.
To quantify how much work each node performs, we define a per-contributor \emph{batch-processing throughput}.
Each worker reports, once per 60\,s interval, the number of microbatches its forward call completed, omitting the backward count, which is generally identical because the backward pass is pinned.
We averaged this value over the node's active lifetime to obtain a single throughput per contributor.
We relate this throughput to each contributor's hardware and network attributes.

\begin{figure}[tbp]
  \centering
  \begin{subfigure}[t]{0.32\linewidth}
    \centering
    \includegraphics[width=\linewidth]{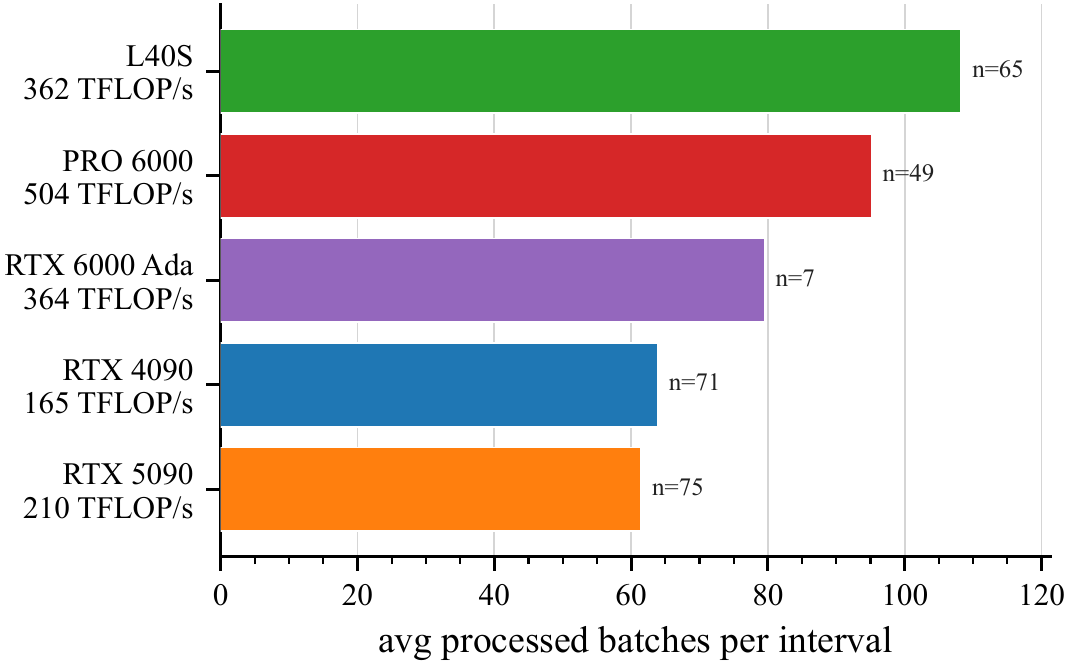}
    \caption{Grouped by GPU family.}
    \label{fig:hetero-device}
  \end{subfigure}
  \hfill
  \begin{subfigure}[t]{0.32\linewidth}
    \centering
    \includegraphics[width=\linewidth]{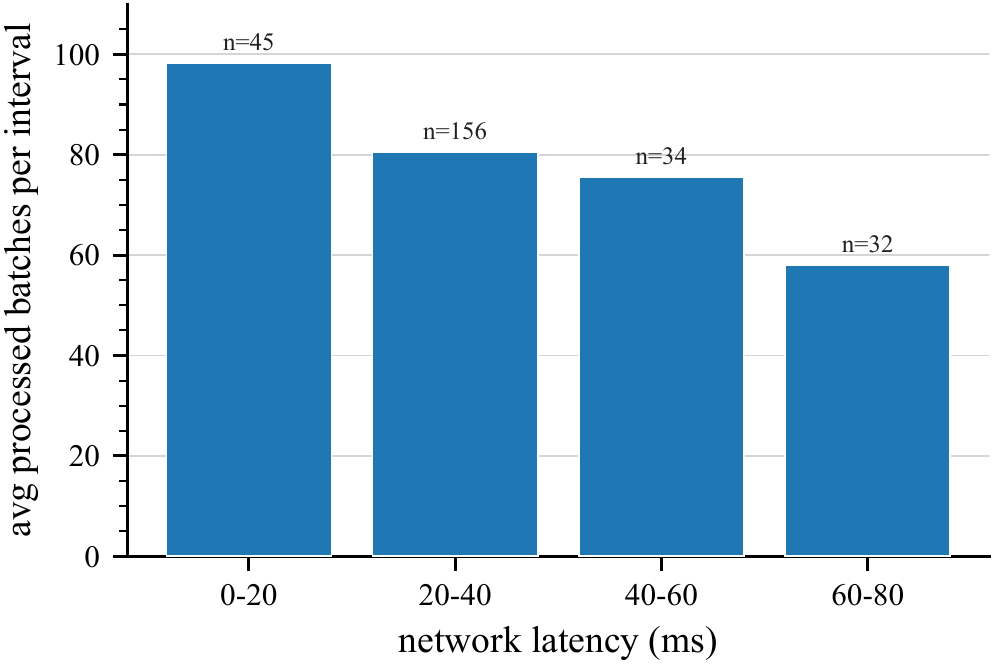}
    \caption{Binned by network latency.}
    \label{fig:hetero-latency}
  \end{subfigure}
  \hfill
  \begin{subfigure}[t]{0.32\linewidth}
    \centering
    \includegraphics[width=\linewidth]{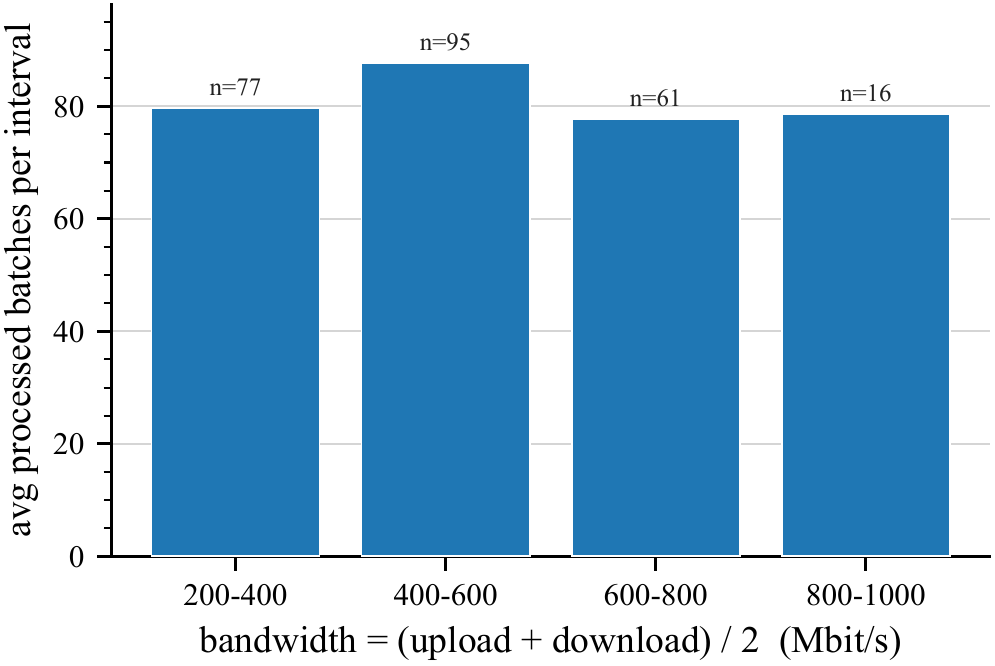}
    \caption{Binned by available bandwidth.}
    \label{fig:hetero-bandwidth}
  \end{subfigure}
  \caption{\textbf{Mean batch-processing throughput per contributor.}
  \textbf{(a)} Grouped by GPU family (the three RTX~PRO~6000~Blackwell variants are merged); bar height is the average over nodes of each node's lifetime-mean processed batches per 60\,s, with contributor count $n$ and the BF16 peak (TFLOP/s) annotated under each card. The ranking is only weakly correlated with peak FLOP/s.
  \textbf{(b)} Binned by network latency (ms); throughput declines monotonically with latency (98$\rightarrow$58 batches per interval, $\approx$40\%). Bar labels report the number of contributors $n$ in each band.
  \textbf{(c)} Binned by available bandwidth (mean of reported upload and download speed, Mbit/s); throughput is essentially flat ($\approx$80 batches per interval). Bar labels report $n$.}
  \label{fig:hetero}
\end{figure}

\paragraph{Throughput does not track raw GPU compute.}
\Cref{fig:hetero-device} groups contributors by GPU family and reports the mean throughput within each family, annotating the dense-BF16 tensor-core peak of each card.
The measured ranking is only weakly correlated with peak FLOP/s.
The L40S (362\,TFLOP/s) leads at 108 batches per interval, ahead of the much more capable RTX~PRO~6000 (504\,TFLOP/s, 95).
In other words, the fastest accelerators are not the ones doing the most work, which indicates that contributor throughput is limited by factors other than on-device compute.

\paragraph{Latency degrades throughput.}
\Cref{fig:hetero-latency} bins contributors by measured network latency, and throughput falls monotonically across the bands.
As discussed in \Cref{trainer_section}, every trainer measures the performance of each worker from the moment it dispatches a batch until the result is returned, a measurement that includes the network latency.
Because every microbatch incurs inter-stage activation and gradient exchange, nodes farther from the trainers in round-trip time spend a larger fraction of each interval blocked on communication, and therefore complete fewer batches.

\paragraph{Bandwidth is not the bottleneck.}
\Cref{fig:hetero-bandwidth} bins the same contributors by available bandwidth, computed as the mean of the reported upload and download speeds, and throughput is flat across the range.
Given the minimum requirement of 200\,Mbit/s, additional bandwidth yields no additional throughput.
The per-batch payloads are small enough that latency, not bandwidth, governs how quickly a node can keep up.

\begin{figure}[tbp]
  \centering
  \includegraphics[width=0.72\linewidth]{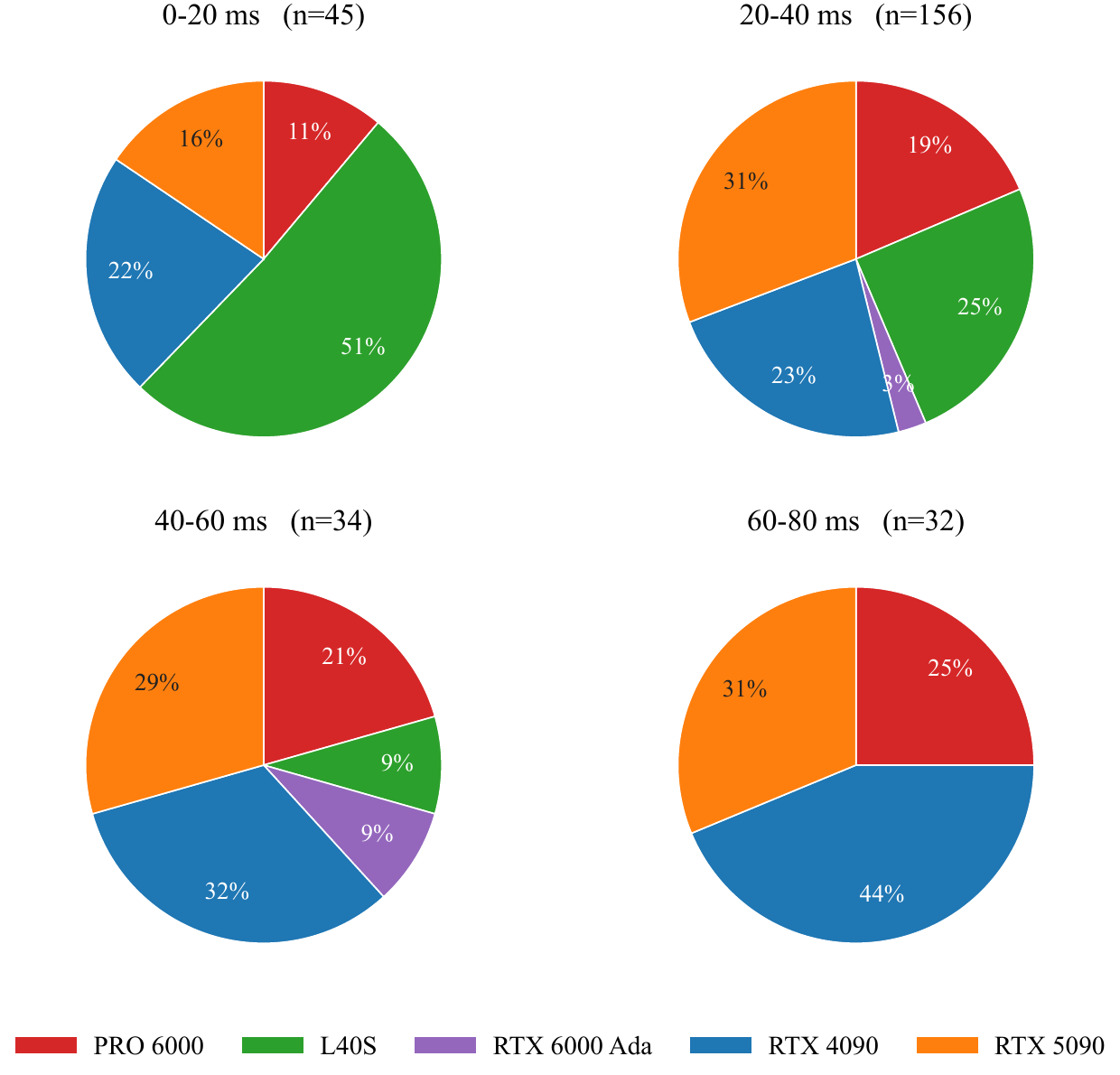}
  \caption{\textbf{GPU-family composition of contributors within each latency band.} Same population as \Cref{fig:hetero-latency}; $n$ per band shown in each title. Low-latency nodes are predominantly L40S cards, while high-latency nodes are almost entirely consumer GPUs, so latency and GPU class are confounded.}
  \label{fig:hetero-gpudist}
\end{figure}

\paragraph{Latency and hardware class are confounded.}
The latency and GPU-type signals are not independent.
\Cref{fig:hetero-gpudist} shows the composition of GPU families within each latency band.
The low-latency population (0--20\,ms) is dominated by L40S cards (51\%), which are the highest-throughput nodes; as latency grows, the mix shifts decisively toward consumer GPUs, until the 60--80\,ms band contains no L40S.
High-latency contributors are therefore predominantly consumer-GPU machines, which are also the lower-throughput cards, so the throughput decline with latency reflects the interplay between network latency and hardware class and should not be read as a pure latency effect.

Taken together, these results show that contributor throughput in our setting is governed by a combination of GPU card and network latency, rather than by raw GPU compute or available bandwidth.

\section{Ablation Studies on Convergence Robustness}
\label{sec:ablations}
In this section we present a series of ablation studies to isolate the importance of various design decisions in Agora.

\paragraph{Setup.}
For our experiments in this section, we use a 1B reparameterized SSN trained with AdamW and AsyncSPARTA.
This is the same training configuration as Pluralis-8B, but tailored to the 1B parameter scale.
Reparameterization confines the output projections and trainable embedding of its head and body stages to a rank-40 subspace, a 51$\times$ reduction in the inter-stage signal.
\Cref{tab:ablation} summarizes the model and training configuration used for our ablation studies.
Note that while we set the total number of steps to 100,000, we only train for about 10,000 steps in most of our experiments.

\begin{table}[t]
\centering
\small
\begin{tabular}{lr}
\toprule
\textbf{Parameter} & \textbf{Value} \\
\midrule
\multicolumn{2}{@{}l}{\textit{Architecture}} \\
Hidden dimension $d$                 & 2,048 \\
Number of layers (total)             & 16 \\
Layers per stage (head / body / tail) & 3 / 4 / 5 \\
Number of pipeline stages            & 4 (1 head, 2 body, 1 tail) \\
Attention heads / KV heads (GQA)     & 32 / 8 \\
Head dimension                       & 64 \\
FFN dim multiplier                   & 1.3 (inner width 7,168) \\
Vocabulary size                      & 128,256 (Llama-3 tokenizer) \\
RoPE base $\theta$                   & 500,000 \\
Max sequence length                  & 2,048 \\
Normalization                        & RMSNorm, $\varepsilon=10^{-5}$, block scale frozen \\
QK-norm + reorder                  & Enabled \\
Subspace rank $r$                    & 40 ($\approx$51$\times$ compression) \\
Trainable / dense-equivalent params  & $\sim$0.94B / $\sim$1.40B \\
\midrule
\multicolumn{2}{@{}l}{\textit{Optimization}} \\
Optimizer                            & AdamW \\
Peak learning rate                   & $4\times10^{-4}$ \\
$(\beta_1, \beta_2)$ / weight decay  & $(0.9, 0.95)$ / 0.1 \\
LR schedule                          & Linear, 1\% floor \\
Warmup / total steps                 & 4,000 / 100,000 \\
Global batch size                    & 2,048 sequences ($\approx$4M tokens) \\
Training corpus                      & FineWeb-Edu \\
Data-parallel averaging              & AsyncSPARTA \\
Sparse share $p$ / cadence $N$       & 5\% / every 20 steps \\
Consensus update                     & Delta rule ($\lambda = 1$) \\
\bottomrule
\end{tabular}
\caption{\textbf{Model and training configuration for the 1B ablation runs.} The dense-equivalent count is the size of the uncompressed model that the rank-40 factorization represents.}
\label{tab:ablation}
\end{table}

\paragraph{Distributed network configuration.}
We use a four-stage pipeline and vary the number of workers assigned to each stage.
Our base configuration places 16 workers on every stage, and we vary this count as contributors join and leave during the heterogeneity studies.
We use L40S GPUs for all workers.

\subsection{Reparameterized SSNs}
We first ask whether reparameterizing the subspace networks of \Cref{ssec:reparam_ssn} affects convergence.
The reparameterized and original SSNs confine the inter-stage signal to the same subspace and differ only in how the constraint is imposed.
The former performs this by folding the factorization $\mW = \mZ\mU_k^{\top}$ into the weights, the latter by regularizing a full-width matrix toward the subspace outside the computational graph at every step. 

\begin{figure}[t]\centering
\includegraphics[width=0.6\linewidth]{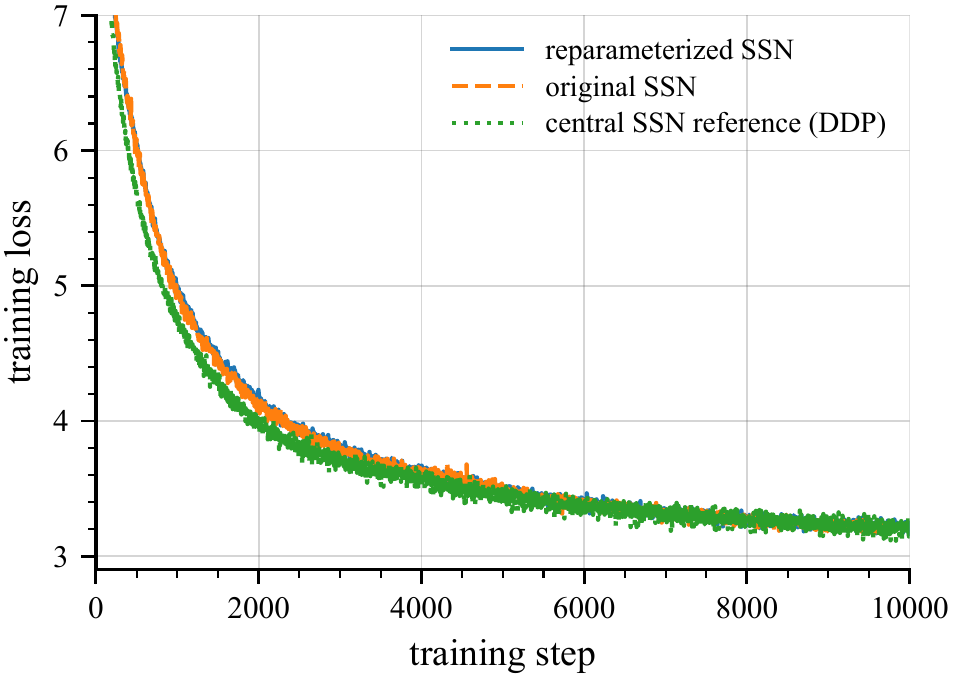}
\caption{\textbf{Training loss of the 1B model under reparameterized vs.\ original SSN.} Both variants use the same rank-40 compression and are compared against a central reference. All training runs use AsyncSPARTA for their data-parallel communication. The curves coincide, so reparameterization leaves convergence unchanged.}\label{fig:ablation-reparam-loss}\end{figure}

Holding the rank-40 compression of \Cref{tab:ablation} fixed for a fair comparison, we train the same 1B model in Agora for about 10,000 steps under each, with a centralized data-parallel run as reference.
The two Agora training loss curves coincide with each other and with the central reference (\Cref{fig:ablation-reparam-loss}), confirming that reparameterization does not impact convergence.

\begin{figure}[t]
\centering
\begin{subfigure}{0.8\linewidth}
\centering
\includegraphics[width=\linewidth]{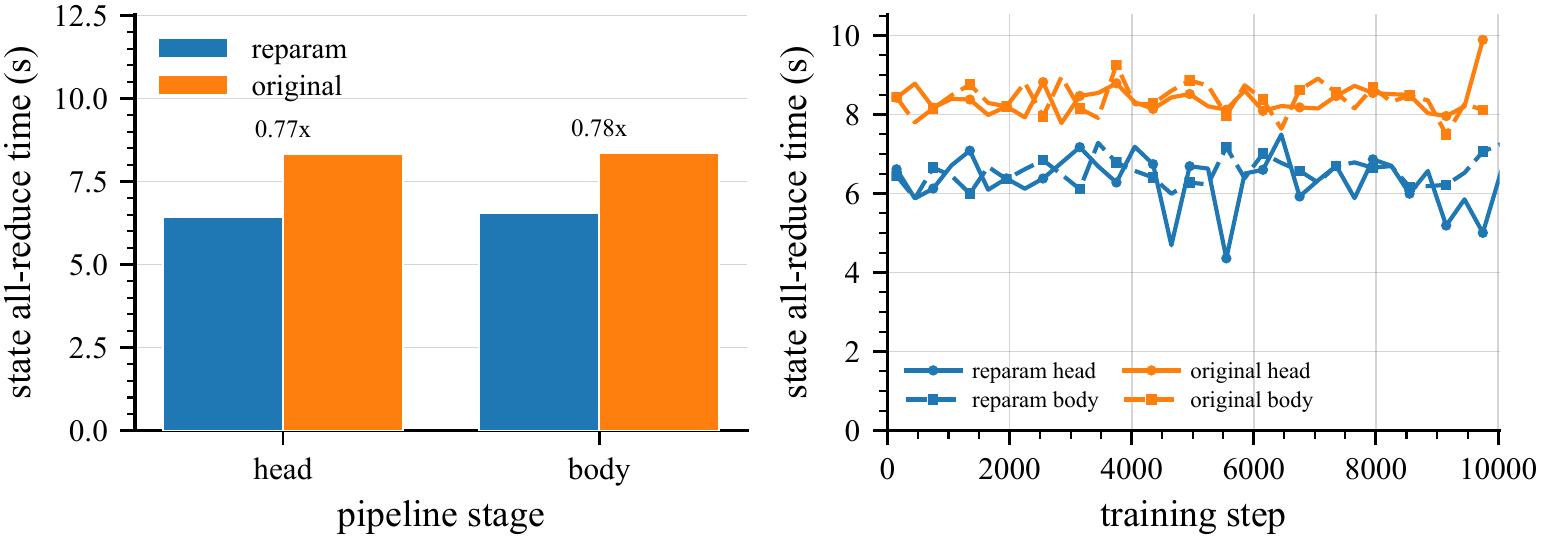}
\caption{SPARTA state all-reduce time.}\label{fig:ablation-reparam-timing-ar}
\end{subfigure}
\vfill
\begin{subfigure}{0.8\linewidth}
\centering
\includegraphics[width=\linewidth]{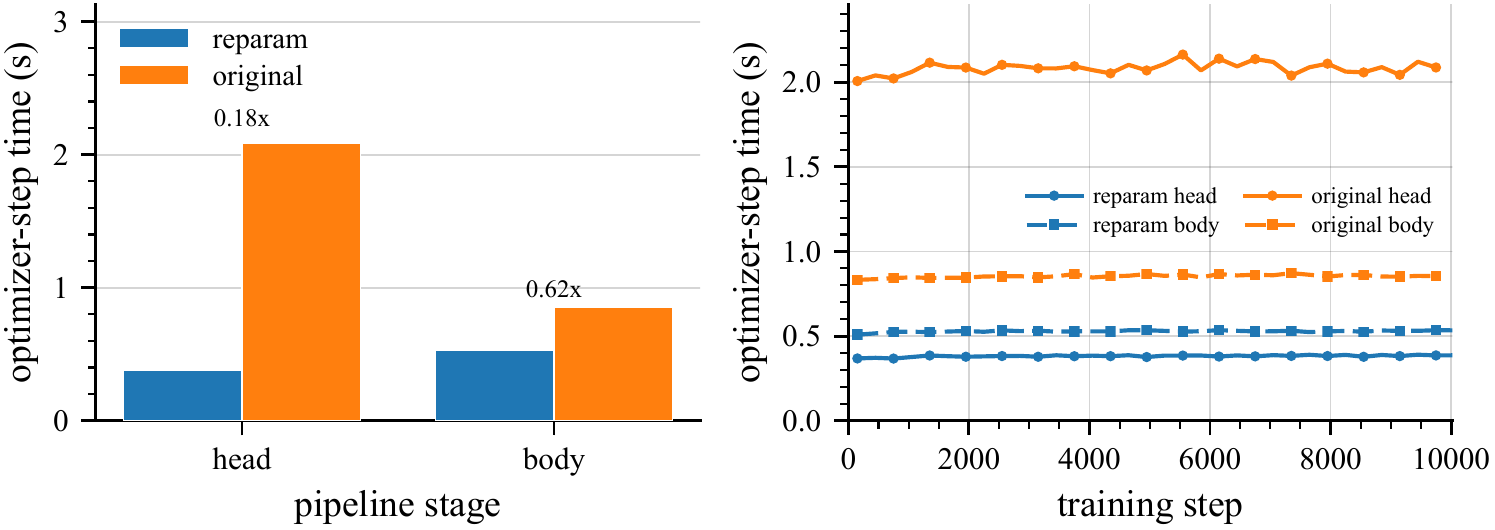}
\caption{Local optimizer-step time.}\label{fig:ablation-reparam-timing-opt}
\end{subfigure}
\caption{\textbf{Per-stage cost of reparameterization, reparameterized vs.\ original SSN.} For the head and body, we report steady-state medians with the reparameterized/original ratio and the per-step trend. The lighter head cuts its all-reduce by about a quarter, and fewer trainable parameters give a faster, cheaper optimizer step.}\label{fig:ablation-reparam-timing}\end{figure}

One key advantage of reparameterization is a lighter state for head and body stages.
Storing only the $k$ trainable columns leaves the body stages about 35\% lighter and the head, whose token embedding is also factorized, about 50\% lighter.
This leaves us with $\sim$0.94B trainable parameters against a $\sim$1.40B dense-equivalent.
The head thus becomes about as light as a body worker, which can open that stage up to contributor nodes.
Because the SPARTA state all-reduce of \Cref{eq:sparta} moves a fixed fraction of the weights, the lighter head cuts its averaging time by roughly a quarter (\Cref{fig:ablation-reparam-timing-ar}).
Moreover, the local optimizer step also occurs faster with fewer parameters to update (\Cref{fig:ablation-reparam-timing-opt}).

\begin{figure}[t]\centering
\includegraphics[width=0.7\linewidth]{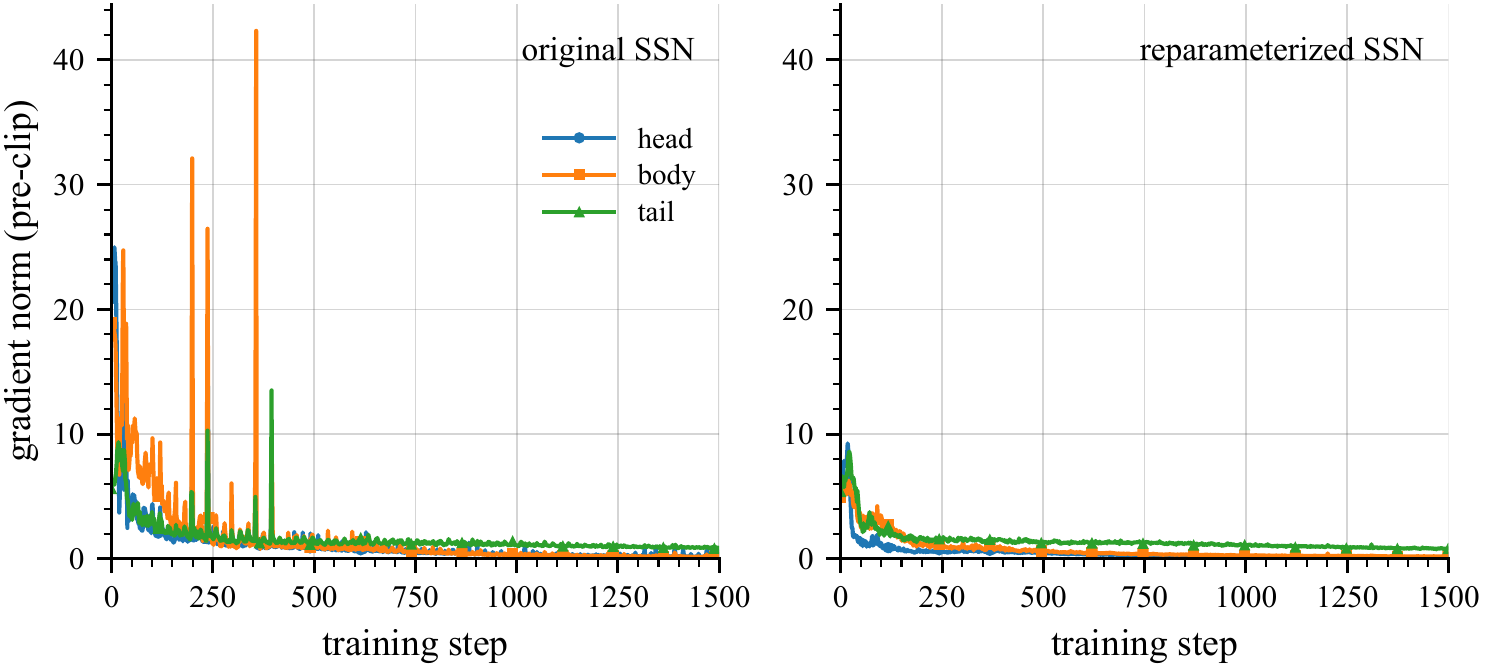}
\caption{\textbf{Gradient magnitude over training.} The original SSN spikes early while the reparameterized formulation stays stable because the compression lives inside the computational graph.}\label{fig:ablation-reparam-grad}\end{figure}

Another advantage of reparameterization is that it removes the out-of-computational-graph regularizer entirely.
In particular, since the low-rank compressors are already folded into the projection matrices, the forward and backward passes are fully aware of this operation.
In contrast, with regularization, the weights are kept full-rank and we project them back into the subspace after the optimizer step is performed.
This allows reparameterized SSN gradients to stay well behaved, whereas the original SSN's gradient magnitude is spiky from the beginning of training (\Cref{fig:ablation-reparam-grad}).

\subsection{AsyncSPARTA and Backward Stochasticity}
We saw in~\Cref{trainer_section} that Agora routes each microbatch through the pool of available devices by a stochastic, latency-aware choice made independently for the forward and backward passes.
As we discussed in \Cref{ssec:async_sparta}, under synchronous averaging this is harmless since every replica of a stage holds identical weights and a backward lands on the same weights wherever it is routed.
AsyncSPARTA breaks this guarantee because between averaging rounds the replicas drift apart, and a backward sent to a different worker rebuilds the gradient from that worker's weights rather than the ones that produced the forward activations.
We remove this inconsistency by pinning the backward pass to the workers that ran the forward, recording the sampled route, and requesting the same workers to run the backward pass.
In the case of worker unavailability, we fall back to stochastic routing.

\begin{figure}[t]\centering
\includegraphics[width=0.6\linewidth]{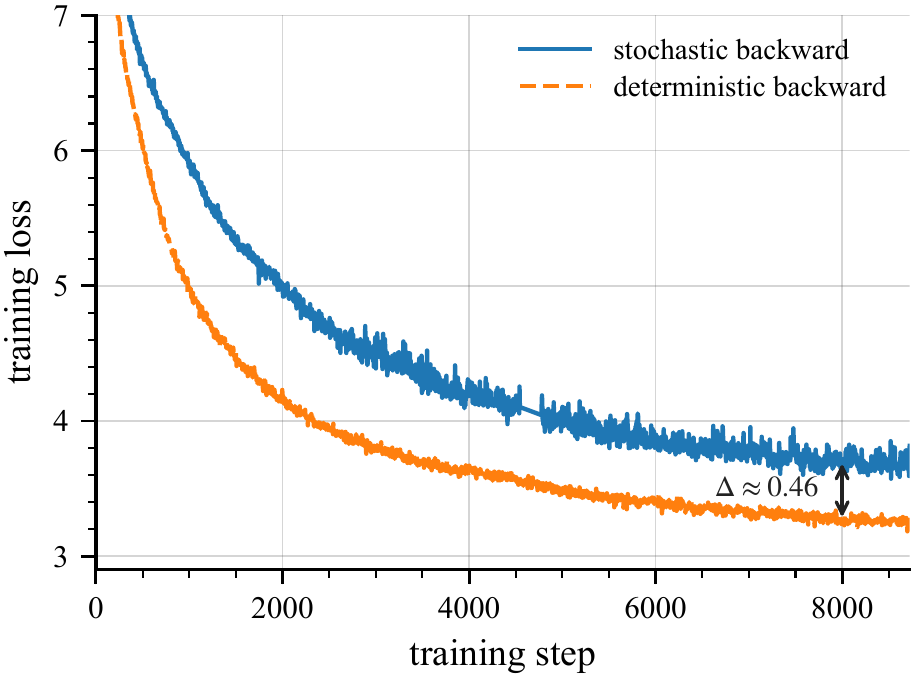}
\caption{\textbf{Training loss under stochastic vs.\ deterministic backward.} The 1B model runs at AsyncSPARTA cadence $N = 20$, with four pipeline stages over 8 data-parallel pipes. Deterministic routing significantly improves convergence.}\label{fig:ablation-backward}\end{figure}

To isolate this effect we train the 1B model under stochastic and deterministic backward routing, identical in all other settings, on four pipeline stages over 8 data-parallel pipes at an AsyncSPARTA cadence of $N = 20$ steps.
As we see in \Cref{fig:ablation-backward}, deterministic backward trains visibly better.
The curves start separating by about step 1,000 and end roughly 0.4 nats apart at step 8,000 as the per-step bias compounds.
This highlights the importance of pinning the backward to the same worker that ran the forward in order to avoid gradient drift and slow convergence.

\subsection{Batch-Size Heterogeneity}
Next, we stress test Agora under heterogeneous local batch sizes.
Since trainers route microbatches by throughput and latency, a faster worker processes more per round.
Thus, the relative batch size across a stage is set by the network rather than prescribed.
As such, we induce a target batch size spread across workers by tuning each worker's bandwidth and latency.
The run uses the four-stage 1B model of \Cref{tab:ablation} with up to 16 workers per stage, each body stage holding 8 permanent and up to 8 redundant.
There is no optimizer-state averaging, and we use the two-phase synchronization mode of \Cref{sec:worker_state_load_sync_phase}, in which a joining contributor first absorbs the stage state by averaging and then processes batches without contributing model updates before becoming fully active.

\begin{table}[t]
\centering
\small
\begin{tabular}{lccc}
\toprule
\textbf{Contributor tier} & \textbf{Upload (Mbit/s)} & \textbf{Download (Mbit/s)} & \textbf{Latency (ms)} \\
\midrule
\multicolumn{4}{@{}l}{\textit{Slow-join tiers}} \\
Fast             & 270         & 270         & 0  \\
Medium           & 210         & 210         & 10 \\
Slow             & 550         & 550         & 35 \\
\midrule
\multicolumn{4}{@{}l}{\textit{Stress-join tiers}} \\
Bare             & 210         & 210         & 80 \\
Asymmetric       & 210         & 600         & 80 \\
Jittery          & 205--600 & 205--600 & 80 \\
\bottomrule
\end{tabular}
\caption{\textbf{Per-tier network configuration used to induce batch-size heterogeneity across the joining workers.} The slow-join tiers set the steady-state relative batch sizes, while the stress tiers sit at the 80\,ms latency boundary that drives the all-reduce failures.}
\label{tab:ablation-het-tiers}
\end{table}

\begin{figure}[t]\centering
\includegraphics[width=0.6\linewidth]{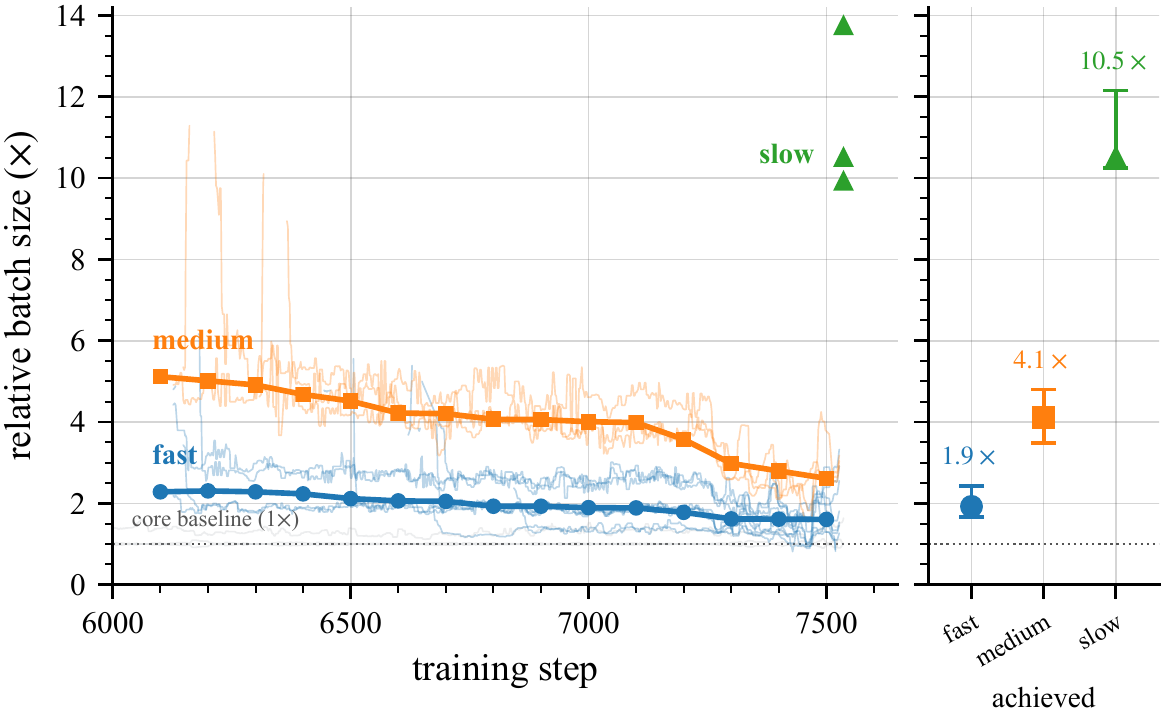}
\caption{\textbf{Achieved heterogeneity ratios across the join tiers.} The fast, medium, and slow tiers follow \Cref{tab:ablation-het-tiers}. Tuning each tier's bandwidth and latency drives a target relative batch size, the fast tier reaching about 2--2.6 to 1, the medium about 3.6--4.5 to 1, and the slow about 10 to 1.}\label{fig:ablation-het-ratios}\end{figure}

The ablation run passes through several phases of churn, with about 87 joins and 87 drops in total.
We take each body stage from 16 workers to 8 near step 4.5k, then open a slow-join schedule near step 5k across the three slow-join tiers of \Cref{tab:ablation-het-tiers}.
We tune the latency and bandwidth of the links to yield steady-state relative batch sizes of about 2:1, 4:1, and 10:1 for the fast, medium, and slow tiers (relative to the worker with the maximum number of batches processed in each stage) as shown in \Cref{fig:ablation-het-ratios}.
A heavier stress phase near step 6.5k adds workers at the stress tiers of \Cref{tab:ablation-het-tiers}, where the 80\,ms round trip pushes them past the state all-reduce timeout.
Seven of the 36 stressed workers, about 19\%, self-terminate after repeated all-reduce failures due to the heavy join pattern.
The remaining stressed workers settle near a batch-size ratio of 8:1 to 12:1. Near step 8.5k, we remove all joining contributors, returning each body stage to the same 8 permanent workers retained after the drop at step 4.5k.

\begin{figure}[t]\centering
\includegraphics[width=0.6\linewidth]{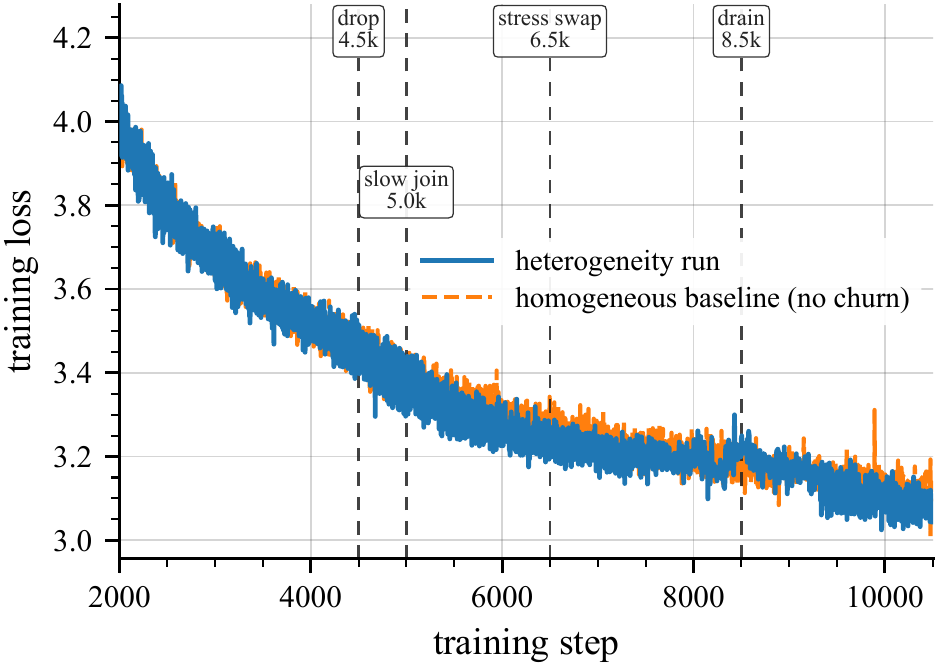}
\caption{\textbf{Training loss across the heterogeneous run.} Markers denote the drop (4.5k), slow join (5k), stress swap (6.5k), and drain (8.5k). The loss generally tracks the homogeneous no-churn baseline. During the stress phase, the sharp increase in all-reduce latency and removal of 7 out of 36 stressed workers coincide with a temporary slowing of convergence; the loss resumes its downward trajectory as all-reduce latency recovers.}\label{fig:ablation-het-loss}\end{figure}

\begin{figure}[t]\centering
\includegraphics[width=0.55\linewidth]{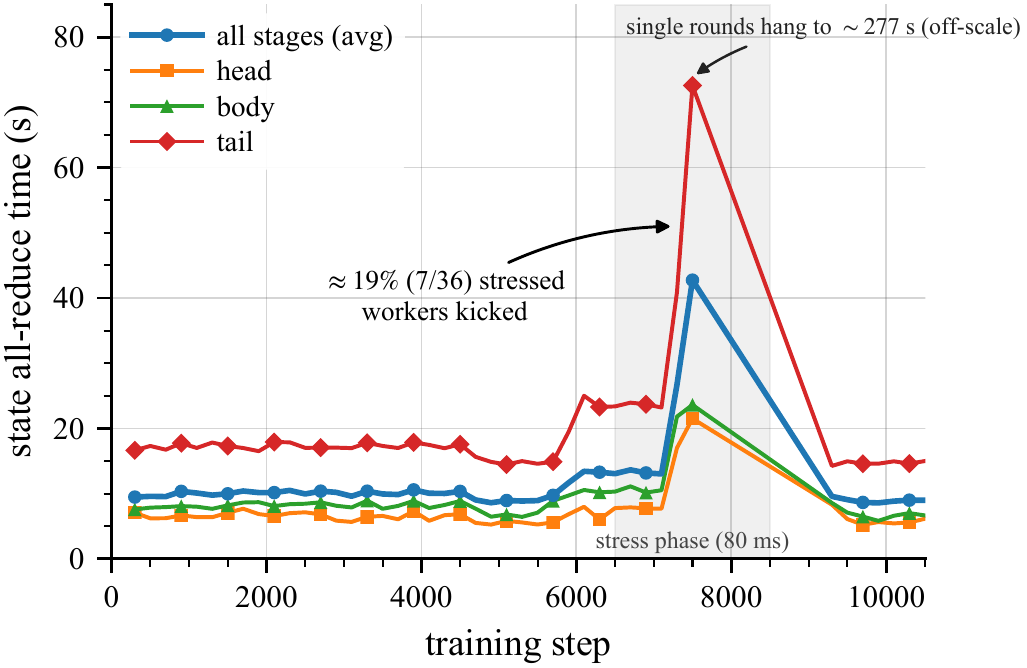}
\caption{\textbf{State all-reduce time per stage.} The stress-phase kick-out fraction is annotated. The averaging time grows with the worker count and under the 80\,ms stress exceeds the timeout for about 19\% of stressed workers, which self-terminate, yet the loss in \Cref{fig:ablation-het-loss} overall stays on track.}\label{fig:ablation-het-ar}\end{figure}

As shown in~\Cref{fig:ablation-het-loss}, the heterogeneous run generally tracks the homogeneous no-churn baseline through the drop and slow-join phases. The stress swap at step 6.5k produces the clearest transient effect. During this phase, state all-reduce latency rises sharply, individual rounds take as long as approximately 277 seconds, and 7 out of 36 stressed workers are removed (\Cref{fig:ablation-het-ar}). Over the same interval, the loss continues to decrease, but its trajectory becomes temporarily shallower. As the stressed workers are pruned and all-reduce latency returns toward its pre-stress regime, the loss resumes a clearer downward trajectory and remains comparable to the homogeneous baseline.

Thus, the stress phase temporarily reduces the convergence rate rather than destabilizing training. This behavior is consistent with our synchronization mode, in which a contributor first absorbs the current state before participating in all-reduce, limiting the influence of an unstable worker on the shared model. Overall, AsyncSPARTA tolerates substantial batch-size heterogeneity and worker churn, and training recovers after the temporary loss of the stressed workers.

\subsection{All-Reduce Failures} \label{sec:ar-fail-convergence}
The heterogeneity study of the previous subsection lets all-reduce failures arise on their own, as a side effect of slow workers crossing the state all-reduce timeout.
Here we instead study the failure rate directly as the variable and hold the relative batch size fixed at 1:1, so that we can ask how many all-reduce failures AsyncSPARTA can absorb before convergence suffers.

Forking from our clean baseline near step 4,400, we run the 1B model of \Cref{tab:ablation} with 16 workers per stage and an AsyncSPARTA cadence of $N = 20$.
We then inject failures into the sparse averaging across the four cases of \Cref{tab:ar-failure-cases}, which span the two data-movement phases of an all-reduce (gather or return) and two failure modes:
\begin{itemize}[nosep]
    \item \textbf{Timeout}: a peer can hang past its deadline, after which it is banned or contributes a zero delta, and the round proceeds.
    \item \textbf{Abrupt Exit}: a peer can exit abruptly after moving a single shard.
\end{itemize}
Either mode can strike while a peer streams its parts to the reducer or while the reducer streams the average back.
We spread these across two populations.
The \emph{persistent workers} carry all four cases, with the abrupt-exit cases 3 and 4 confined to the body stages and the timeout cases 1 and 2 spanning every stage.
The \emph{joining contributors} only send updates, so they carry only the sender-side cases 1 and 3, on an escalating schedule.
The placement of these workers is given in \Cref{tab:ar-failure-placement}.

The injected-failure phases range from a clean control, through light failures at about one round in five and heavy failures every second round, to zombie contributors that fail every round without getting kicked out.
We randomize the failure offset per worker so that failures occur haphazardly, mirroring a realistic training run with compute contributors.
We summarize these various phases in \Cref{tab:ar-failure-phases}.

\begin{table}[t]
\centering
\small
\begin{tabular}{@{}cclp{0.70\linewidth}@{}}
\toprule
\textbf{Case} & \textbf{Phase} & \textbf{Failure mode} & \textbf{Impact on the averaging round} \\
\midrule
1 & Gather & Timeout & Reducer bans the peer and averages its slot over the remaining contributors. \\
2 & Return & Timeout & Senders never receive their averaged slot and time out. \\
3 & Gather & Abrupt Exit & Stream closes mid-gather and the reducer can stall on the missing parts until timeout. \\
4 & Return & Abrupt Exit & Senders receive fewer averaged parts than expected, leaving a partial average. \\
\bottomrule
\end{tabular}
\caption{\textbf{The four AR failure cases we inject for this ablation study.} They span the two data-movement phases of a sparse AR, the gather that streams each peer's parts to the reducer and the return that streams the averaged parts back. Together with the two failure modes, a hang to its timeout and an abrupt exit, we cover four different cases of AR failures. Cases 1 and 2 are graceful timeouts and cases 3 and 4 are abrupt exits.}
\label{tab:ar-failure-cases}
\end{table}

\begin{table}[t]
\centering
\small
\begin{tabular}{@{}lccccc@{}}
\toprule
\textbf{Stage} & \textbf{Clean} & \textbf{Case 1} & \textbf{Case 2} & \textbf{Case 3} & \textbf{Case 4} \\
\midrule
Head        & 8 & 4 & 4 & 0 & 0 \\
Body (each) & 4 & 2 & 2 & 2 & 2 \\
Tail        & 8 & 4 & 4 & 0 & 0 \\
\bottomrule
\end{tabular}
\caption{\textbf{Placement of the four failure cases across the persistent workers.} Each of the two body stages carries the split shown.}
\label{tab:ar-failure-placement}
\end{table}

\begin{table}[t]
\centering
\small
\begin{tabular}{@{}llp{0.25\linewidth}p{0.40\linewidth}@{}}
\toprule
\textbf{Phase} & \textbf{Injected failure} & \textbf{Rate} & \textbf{Purpose} \\
\midrule
Control     & None                  & 0\%                          & Clean baseline \\
Light       & Gather hang (case 1)  & 1 in 5 rounds                  & Mild failures \\
Heavy       & Gather hang (case 1)  & Every 2nd round                & Heavy failures with constrained bandwidth \\
Disconnect  & Gather exit (case 3)  & Once, after $\sim$100 rounds & Short-lived abrupt crash \\
Post-join   & Gather hang (case 1)  & Every 2nd, after join          & Failure begins only after a clean join \\
Zombie      & Gather hang (case 1)  & Every round                    & Persistent failure, never self-terminates \\
Aggressive  & Gather hang (case 1)  & Every round                    & Fast kick-out after two failures \\
\bottomrule
\end{tabular}
\caption{\textbf{Sequence of failure-injection phases applied to the joining contributors.} Over the run the schedule moves from a clean baseline through mild and heavy failures to persistent every-round failures. The case numbers refer to \Cref{tab:ar-failure-cases}.}
\label{tab:ar-failure-phases}
\end{table}

\begin{figure}[t]
\centering
\includegraphics[width=0.55\linewidth]{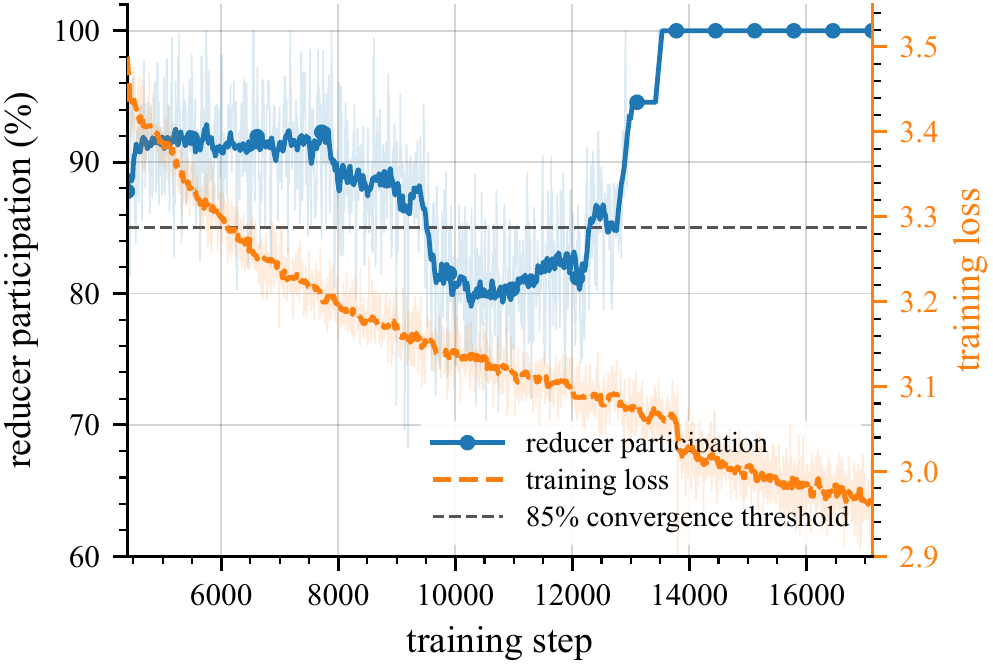}
\caption{\textbf{Convergence under injected all-reduce~(AR) failures at fixed 1:1 heterogeneity.} The run follows the AR failure schedule of \Cref{tab:ar-failure-phases}. When a large fraction of AR operations fail, convergence can stall (see steps 9k to 12k). At approximately step 14k, removing the faulty participants allows training to resume its earlier convergence rate. This observation motivates our self-termination policy, described in~\Cref{worker_section}, under which workers experiencing repeated AR failures remove themselves from the system.}
\label{fig:ablation-arfail}
\end{figure}

As shown in~\Cref{fig:ablation-arfail}, frequent AR failures slow convergence during the most aggressive failure phase. The loss continues to decrease, but its trajectory becomes noticeably shallower around 3.1. This degradation is recoverable: as the failure-injecting workers are progressively pruned, the loss resumes a clearer downward trajectory. Once the remaining faulty workers have been removed and reducer participation reaches 100\% at approximately step 14k, the loss decreases from roughly 3.04 to 2.97 by the end of the run.

The key quantity is reducer participation, defined as the fraction of expected contributors that complete an averaging round. As shown in~\Cref{fig:ablation-arfail}, the strongest slowdown in loss reduction coincides with the interval during which participation remains below roughly 85\%. The downward trajectory strengthens again as participation recovers above this level, eventually reaching 100\% after the remaining faulty workers are removed. This suggests an empirical operating threshold near 85\%: sustained participation below this level substantially impairs convergence, whereas the effect is reversed as participation recovers. This separation motivates the final self-termination policy described in~\Cref{worker_section}, where a worker terminates after two failures within any three consecutive rounds, rather than only after two consecutive failures.

\clearpage
\section{Acknowledgments}
\label{sec:contributors}
Pluralis-8B was made possible by a pool of compute contributed by 176 unique node operators. Many of them are unknown to us personally, yet, like us, they saw the importance and potential impact of this effort and chose to participate. We thank them all.

\clearpage
\bibliographystyle{plainnat}
\nobibliography*
\bibliography{main}

\end{document}